\pdfoutput=1
\documentclass[review]{elsarticle}
\usepackage{subfigure}
\usepackage{longtable}
\usepackage{algorithm, algorithmic}
\usepackage{bm}
\usepackage{booktabs}
\usepackage{multirow}
\usepackage{geometry}
\usepackage{amsfonts,amssymb}
\usepackage{float}
\usepackage{bbding}%%包含勾叉的一个bao
\usepackage{amsmath} % 添加amsmath包来支持公式编号
\usepackage[utf8]{inputenc}
\usepackage{CJKutf8}

\usepackage{lineno,hyperref}
\modulolinenumbers[5]

\begin{document}
\begin{CJK*}{UTF8}{gbsn}
	%%%%%%%%%%%%%%%%%%%%%%%%%%%%%%%%%%%%%%%%%%%%%%%%%%%%%%%%%%%%%%%%%%%%%
\begin{frontmatter}
	
	\title{A Survey of Recursive and Recurrent Neural Networks}
	
	\author[JUP]{Jian-wei Liu\corref{mycorrespondingauthor}}
	\ead{liujw@cup.edu.cn}
	\cortext[mycorrespondingauthor]{Corresponding author}
	
	\author[JUP]{Bing-rong Xu}
	\author[JUP]{Zhi-yan Song}
	
	\address[JUP]{Department of Automation, College of Artificial Intelligence, China University of Petroleum, Beijing, Beijing, China}
		\begin{abstract}
			 Recursive and recurrent neural networks are the main realization forms of sequence models based on neural networks, which have been developed rapidly in recent years. Recurrent neural networks are basically standard processing methods for machine translation, question and answer, and sequence video analysis. It is also the mainstream modeling method for handwriting automatic synthesis, speech processing and image generation. In this paper, the branches of recursive and recurrent neural networks are classified in detail according to the network structure, training objective function and learning algorithm implementation. They are roughly divided into three categories: The first category is General Recursive and Recurrent Neural Networks, including Basic Recursive and Recurrent Neural Networks, Long Short Term Memory Recursive and Recurrent Neural Networks, Convolutional Recursive and Recurrent Neural Networks, Differential Recursive and Recurrent Neural Networks, One-Layer Recursive and Recurrent Neural Networks, High-Order Recursive and Recurrent Neural Networks, Highway Networks, Multidimensional Recursive and Recurrent Neural Networks, Bidirectional Recursive and Recurrent Neural Networks; the second category is Structured Recursive and Recurrent Neural Networks, including Grid Recursive and Recurrent Neural Networks, Graph Recursive and Recurrent Neural Networks, Temporal Recursive and Recurrent Neural Networks, Lattice Recursive and Recurrent Neural Networks, Hierarchical Recursive and Recurrent Neural Networks, Tree Recursive and Recurrent Neural Networks; the third category is Other Recursive and Recurrent Neural Networks, including Array Long Short Term Memory, Nested and Stacked Recursive and Recurrent Neural Networks, Memory Recursive and Recurrent Neural Networks. Various networks cross each other and even rely on each other to form a complex network of relationships. In the context of the development and convergence of various networks, many complex sequence, speech and image problems are solved. After a detailed description of the principle and structure of the above model and model deformation, the research progress and application of each model are described, and finally the recursive and recurrent neural network models are prospected and summarized.
		\end{abstract}

	\end{frontmatter}
	%%%%%%%%%%%%%%%%%%%%%%%%%%%%%%%%%%%%%%%%%%%%%%%%%%%%%%%%%%%%%%%%%%%%%%%%%%%%
	
	\section{Introduction}
	Recursive Neural Networks (RecursiveNNs) and Recurrent Neural Networks (RecurrentNNs) are two types of artificial neural networks (ANNs) structures. ANNs was originally developed as a mathematical model to simulate the information processing process of the biological brain. Although it is now clear that ANNs have almost no similarities with real biological neurons, ANNs have been widely used as pattern classifiers. The basic structure of ANNs is a network composed of single neurons or nodes. After neurons are layered, they are connected by connecting edges with weights to represent the interaction between neurons.
	
	There are many types of ANNs, and their characteristics vary greatly. Generally, ANNs can be divided into recurrent connected ANNs and non-recurrent connected ANNs, that is, a cyclic graph structure and acyclic graph structure. Neural networks with recurrent connections are called RecurrentNNs or RecursiveNNs.
	
	\begin{figure}[H]
		\centering
		\includegraphics[width=0.8\textwidth]{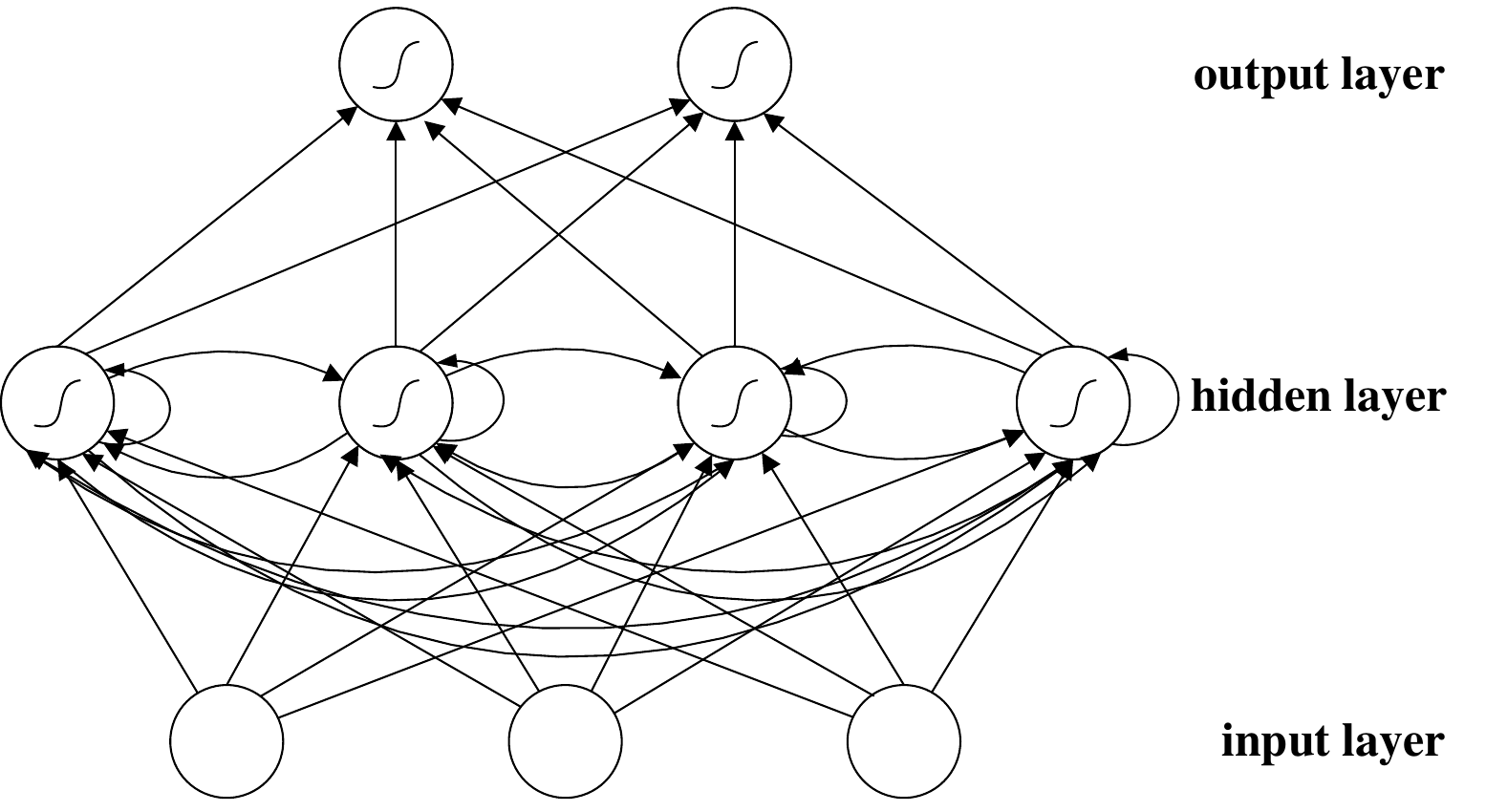}
		\caption{RNNs structure.}
		\label{fig:rnns_structure}
	\end{figure}
	
\begin{figure}[H]
		\centering
		\includegraphics[width=0.8\textwidth]{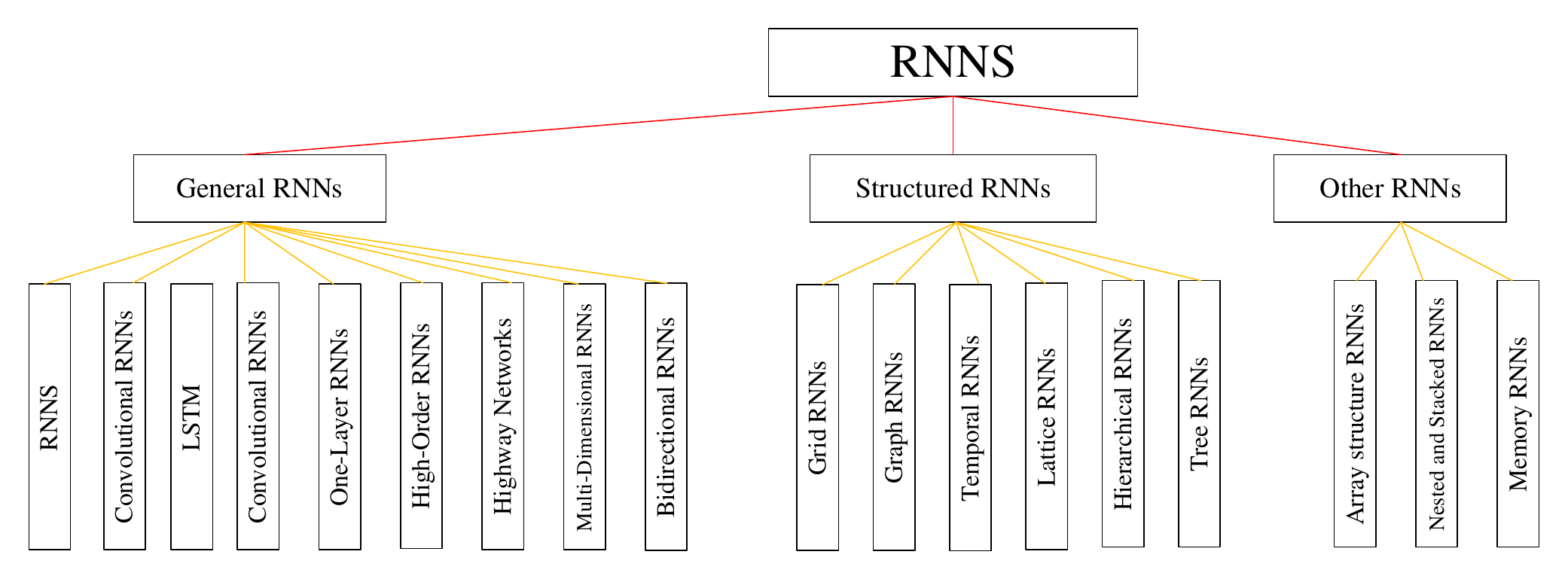}
		\caption{RNNs classification.}
		\label{fig:rnns_classification}
	\end{figure}

	In this paper, RecursiveNNs and RecurrentNNs are collectively referred to as RNNs. Because RNNs are inspired by the periodic connections of neurons in the brain, they use recurrent iteration functions to store information. RNNs have several advantages: RNNs can well capture context information and realize temporal dependency learning (because RNNs can learn what to store and what to ignore); RNNs can learn different types of data representation; RNNs can represent sequences (such as words) order well. However, RNNs also have some shortcomings, which limit their wide application in practical problems. The most serious problem is that it is difficult to model large-scale temporal dependencies. Therefore, in order to solve more complex problems, many structural variants based on RNNs have appeared. The classification of RNNs summarized in this paper is shown in Fig.~\ref{fig:rnns_classification}.
	
	\section{General RNNs}
	There are many types of General RNNs, but the most basic ones are RecursiveNNs and RecurrentNNs, so that the subsequent networks are all ordinary variants of it. It is generally applied to a series of problems such as handwriting recognition, speech recognition, natural language processing, and computer vision; Convolutional RNNs have good effects in image and audio, and are mainly used in the re-recognition or tracking process of characters or the representation of audio features learning and audio labeling tasks; LSTM is a relatively typical and mature structure among general RNNs structural variants, which can be applied to multiple phonemes, images and other issues; Differential RNNs learn different information through different orders for video processing and action recognition; One-layer RNNs are used to solve pseudo-convex optimization problems with equality and inequality constraints; High-order RNNs allow high-order interactions of neurons with strong storage capabilities, and are used in grammatical reasoning and target detection; the highway connection of the Highway Network allows information to cross different layers unimpeded, thereby alleviating the problem of gradient disappearance when building a deeper LSTM; Multi-dimensional RNNs replace the single recurrent link of RNNs with the spatio-temporal dimension in the data in dimension to solve local distortion; Bidirectional RNNs can use two separate hidden layers to process data in both directions to simultaneously utilize the previous context and the future context. The following is a detailed introduction to the above RNNs.
	
	\subsection{RecursiveNNs and RecurrentNNs}
	Next, the various network architectures of RecursiveNNs and RecurrentNNs will be described in detail.
	
	\subsubsection{RecurrentNNs}
	RecurrentNNs is an extension of traditional feedforward neural networks that can handle variable-length sequence inputs. It learns the hidden representation of the variable-length input sequence through the internal recurrent hidden variables. The output of the activation function of the hidden variable at each time step depends on the output of the recurrent hidden variable activation function at the previous time step \cite{1}.
	
	Given an input sequence  $x=\mathrm{(}{{x}_{1}},{{x}_{\mathrm{2}}}\mathrm{,}\ldots \mathrm{,}{{x}_{\text{T}}}\mathrm{)}$, the recurrent update process of RecurrentNNs hidden variables is as follows:
	
	\begin{equation}
	{{h}_{\text{t}}}=\left\{ \begin{matrix}
		\mathrm{0,}  \\
		\phi \mathrm{(}{{h}_{\text{t}-1}}\mathrm{,}{{x}_{\text{t}}}\mathrm{),}  \\
	\end{matrix} \right.\begin{matrix}
		\text{t}=\mathrm{0}  \\
		\mathrm{otherwise}  \\
	\end{matrix}
	\end{equation}
	
	where $\phi $ is a nonlinear activation function (such as the logistics sigmoid activation function pre-activated by the affine transformation). The output $y=({{y}_{\mathrm{1}}}\mathrm{,}{{y}_{\mathrm{2}}}\mathrm{,}\ldots \mathrm{,}{{y}_{\text{T}}}\mathrm{)}$ of RecurrentNNs may also be of variable length.So in general, the implementation of the hidden variable update process in .Eq.(1) is as follows:
	
	\begin{equation}
{{h}_{\text{t}}}=\text{g}\mathrm{(}\text{W}{{x}_{\text{t}}}+\text{U}{{h}_{\text{t}-1}}\mathrm{)}
	\end{equation}
	
where $g$ is a smooth bounded function (such as logistics sigmoid function or hyperbolic tangent function).$W$ is the weight matrix of the hidden variable input to this time step, and $U$ is the weight matrix of the hidden variable from the previous time step to this time step.   

Given the current state ${{h}_{t}}$, RecurrentNNs can be used to represent the joint probability distribution on the input sequence, that is, to explain the update process of RecurrentNNs from the viewpoint of a generative model: the update formula at each time step generates a conditional probability distribution, the product of the conditional probability distribution at all time steps to obtain the joint probability distribution, and RecurrentNNs can naturally represent the probability distribution on the variable-length sequence, and a special termination symbol can be introduced to detect the end position of the variable-length sequence. According to the Bayesian formula, the chain rule, the entire joint probability distribution of a random variable sequence $x=\mathrm{(}{{x}_{\mathrm{1}}}\mathrm{,}{{x}_{\text{2}}}\mathrm{,}\ldots \mathrm{,}{{x}_{\text{T}}}\mathrm{)}$ with sequence length $T$ can be expressed as:
	
	\begin{equation}
p\mathrm{(}{{x}_{\mathrm{1}}}\mathrm{,}\ldots \mathrm{,}{{x}_{\text{T}}}\mathrm{)}=\text{p}\mathrm{(}{{x}_{\text{1}}}\mathrm{)}\text{p}\mathrm{(}\left. {{x}_{2}} \right|{{x}_{\mathrm{1}}}\mathrm{)}\text{p}\mathrm{(}\left. {{x}_{\mathrm{3}}} \right|{{x}_{\mathrm{1}}}\mathrm{,}{{x}_{\mathrm{2}}}\mathrm{)}\ldots \text{p}\mathrm{(}\left. {{x}_{\text{T}}} \right|{{x}_{\mathrm{1}}}\mathrm{,}\ldots \mathrm{,}{{x}_{\text{T}-\mathrm{1}}}\mathrm{)}
	\end{equation}

The last element is a special sequence termination symbol, and each conditional probability distribution in the sequence joint probability distribution is parameterized into a neural network form to obtain:

	\begin{equation}
\text{p}\mathrm{(}\left. {{x}_{\text{t}}} \right|{{x}_{\mathrm{1}}}\mathrm{,}\ldots \mathrm{,}{{x}_{\text{t}-\mathrm{1}}}\mathrm{)}=\text{g}\mathrm{(}{{h}_{\text{t}}}\mathrm{)}
	\end{equation}

where ${{h}_{\text{t}}}$ is from Eq.(1).
	
\subsubsection{RecursiveNNs}
RecursiveNNs operate on structured input, not just for sequence structures. RecursiveNNs have been applied to grammatical analysis, sentence-level sentiment analysis, paraphrase detection, morphology oriented word vector representation, machine question answering and political ideology detection. The essence of RecursiveNNs is the process of recurrently constructing a larger subtree by multiple subtrees of the same structure.

The architecture of RecursiveNNs \cite{2} is: RecursiveNNs recursively apply the same set of weights along the length of the sequence. Given a location directed acyclic graph, it visits nodes in topological order and applies transformations recursively in the previously calculated sub representation to generate further representation. In fact, RecurrentNNs are just RecursiveNNs with a specific structure.

\begin{figure}[H]
	\centering
	\includegraphics[width=0.8\textwidth]{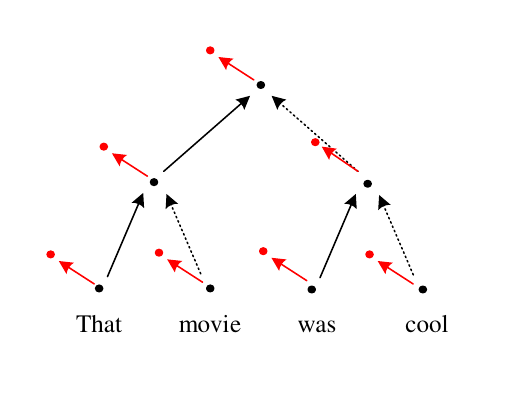}
	\caption{Recursive neural network with bottom-up propagation.}
	\label{fig3}
\end{figure}

Given a binary tree structure with fixed leaf nodes, for example, a parse tree with word vectors on the leaves (see Fig.~\ref{fig3}), RecursiveNNs calculates the representation of the internal node $\eta $ as follows:

\begin{equation}
{{x}_{\text{h}}}=\sigma \mathrm{(}{{\text{W}}_{\text{L}}}{{x}_{\text{l}\mathrm{(}\text{h}\mathrm{)}}}+{{\text{W}}_{\text{R}}}{{x}_{\text{r}\mathrm{(}\text{h}\mathrm{)}}}+\text{b}\mathrm{)}
\end{equation}

where $l(\eta )$ and $r(\eta )$ are the left and right child nodes of $\eta $, ${{\text{W}}_{\text{L}}}$ and ${{\text{W}}_{\text{R}}}$ are the weight matrices connecting the left and right child nodes to the parent node, and $b$ is the bias vector. Assuming that ${{\text{W}}_{\text{L}}}$ and ${{\text{W}}_{\text{R}}}$ are square matrices, and do not distinguish whether $l(\eta )$ and $r(\eta )$ are leaves or internal nodes, there is an interesting explanation in this case: the initial representation of the leaf node and the non-terminal intermediate representation are in the same space. In the parse tree example, RecursiveNNs combines two sub-phrases to generate a longer phrase in the same space.Therefore, there is a final output layer at root $\rho $:

\begin{equation}
y=\gamma \mathrm{(}\text{U}{{x}_{\rho }}+\text{c}\mathrm{)}
\end{equation}

where $\text{U}$ is the output weight matrix and $\text{c}$ is the bias vector of the output layer. In supervised learning tasks, supervision occurs at this layer. Therefore, in the learning process, $y$ will produce an initial error, and it will propagate back from the root to the leaf nodes.

\subsubsection{The Difference between RecursiveNNs and RecurrentNNs}
\begin{figure}[H]
	\centering
	\includegraphics[width=0.8\textwidth]{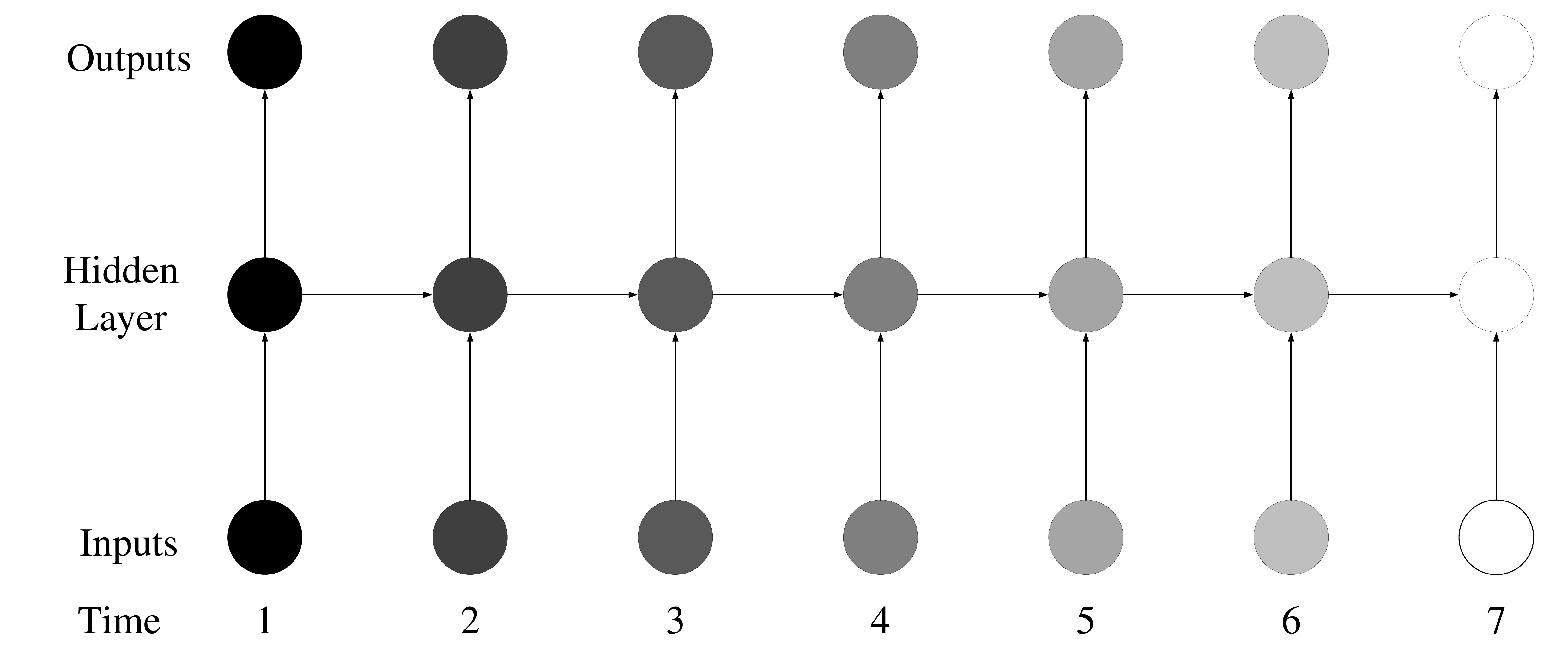}
	\caption{Unfold along time.}
	\label{fig4}
\end{figure}

RecurrentNNs are seen as unfolding along time, processing information of sequence structure. For example, Fig.~\ref{fig4} is a RecurrentNNs, which expands over time \cite{3}. At each timestep, the input of the hidden layer includes the user's input and the output of the hidden layer at the previous timestep.

RecurrentNNs is a special case of RecursiveNNs. RecursiveNNs is more like a hierarchical network. In this network, the input sequence is actually not related to time, but the input must be processed hierarchically in the form of a tree. When RecurrentNNs encounters a sequence with a different length from the sequence seen during training during the test, it cannot handle the position-related weights, so the weights are shared along the sequence length (the dimensionality remains the same). In RecursiveNNs, for the same reason, the weights on each node are also shared (the dimensionality remains the same).

Fig.~\ref{fig5} shows what a RecursiveNNs looks like:

\begin{figure}[H]
	\centering
	\includegraphics[width=0.8\textwidth]{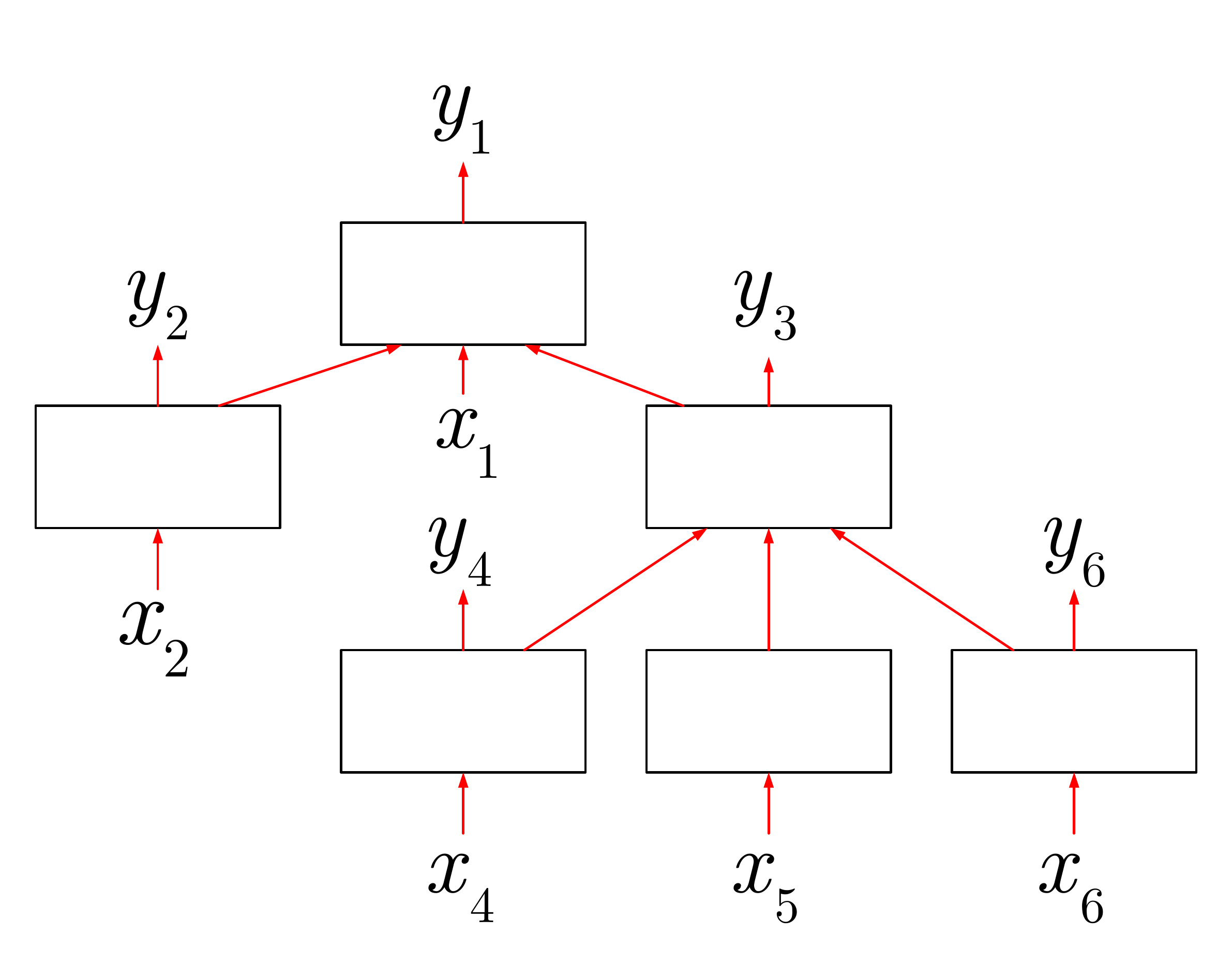}
	\caption{RecursiveNNs.}
	\label{fig5}
\end{figure}

According to Fig.~\ref{fig5}, it is easy to understand why it is called RecursiveNNs. The child node of each parent node is just a node similar to that node. Different neural networks are suitable for different scenarios. If you want to generate characters one by one, then a RecurrentNNs is very suitable. But if you want to generate a parse tree, it is better to use a RecursiveNNs because it helps to create a better hierarchical representation.
	
	\subsection{LSTM}
	Next, the various network architectures of LSTM will be described in detail.
	
	\subsubsection{Difficulties in Training RecurrentNNs}
	1)Training recurrent networks
	
	 Although RecurrentNNs is a simple and powerful model in theory, it is difficult to get good training in practice. One of the main reasons is the gradient disappearance and gradient explosion problem described in \cite{4} \cite{5}. This question is discussed below.
	 
	 At time $t$, assuming the input variable ${{x}_{t}}$ and the state variable ${{h}_{t}}$, RecurrentNNs can be expressed as:
	
	\begin{equation}
{{h}_{t}}=F({{h}_{t-1}},{{x}_{t}},\theta )\
	\end{equation}
	
The above formula can be parameterized into the following form:

\begin{equation}
{{h}_{t}}={{W}_{rec}}\sigma ({{h}_{t-1}})+{{W}_{in}}{{x}_{t}}+b
\end{equation}
	where ${{W}_{rec}}$ is the recurrent weight matrix, $b$ is the bias vector, and ${{W}_{in}}$ is the input weight matrix. Let $\theta =\left\{ {{W}_{rec}},{{W}_{in}},b \right\}$ denote the parameter set of RecurrentNNs. The initial value ${{h}_{0}}$ is given by the user, or another learning algorithm is used to learn the initial value of ${{h}_{0}}$ on the data. $\sigma $ is the sigmoid function acting on the components of the pre-activation function.
	
\begin{figure}[H]
		\centering
		\includegraphics[width=0.8\textwidth]{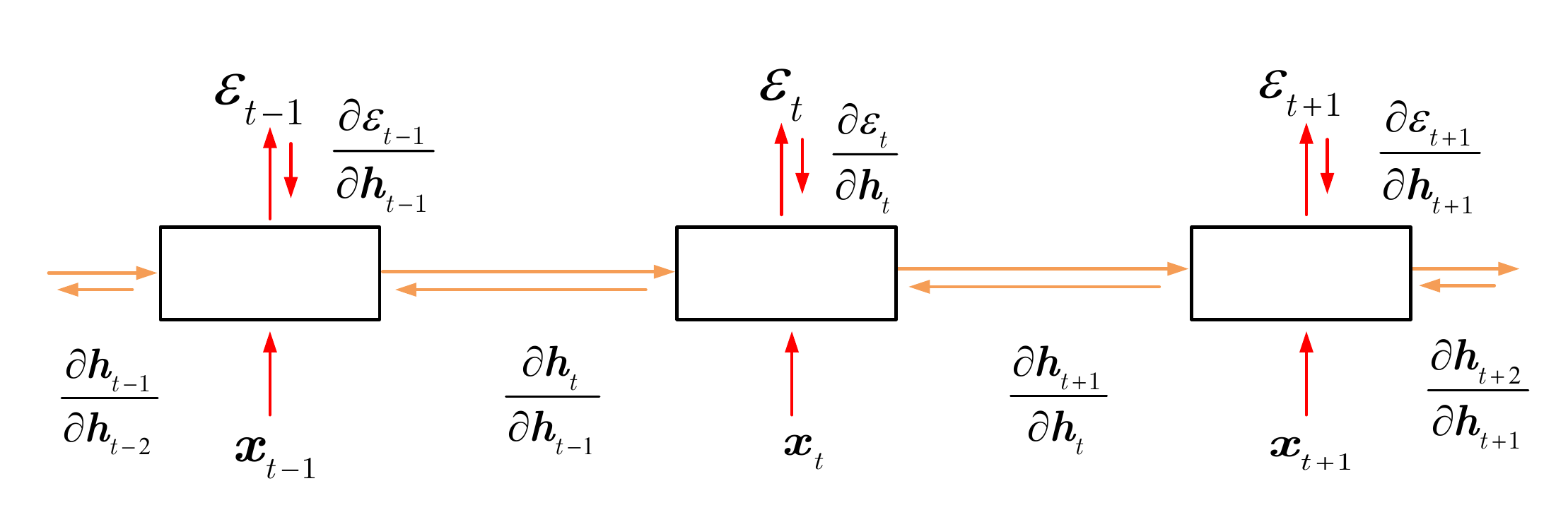}
		\caption{Unrolling recurrent neural networks.}
		\label{fig6}
	\end{figure}

	One of the methods to calculate the gradient is the Back Propagation Through Time (BPTT), in which the recurrent model is represented by a multi-layer model (with an infinite number of layers), and the back propagation process is applied to the expanded model (see Fig.~\ref{fig6}).
	
	By creating a copy of the model for each time step, the recurrent neural network is unfolded in time. Use $h_t$ to represent the hidden state of the network at time $t$, $x_t$ to represent the input of the network at time $t$, and ${{\varepsilon }_{t}}$ to represent the error of the network output at time $t$.
	
	Tomas Mikolov et al. deviated from the classic BPTT equation and rewritten the gradient to better highlight the explosion gradient problem:
	
	\begin{equation}
\frac{\partial \varepsilon }{\partial \theta }=\sum\limits_{1\le t\le T}{\frac{\partial {{\varepsilon }_{t}}}{\partial \theta }}
	\end{equation}
	
	\begin{equation}
\frac{\partial {{\varepsilon }_{t}}}{\partial \theta }=\sum\limits_{1\le k\le t}{(\frac{\partial {{\varepsilon }_{t}}}{\partial {{h}_{t}}}}\frac{\partial {{h}_{t}}}{\partial {{h}_{k}}}\frac{\partial {}^{+}{{h}_{k}}}{\partial \theta })
	\end{equation}
	
	\begin{equation}
\frac{\partial {{h}_{t}}}{\partial {{h}_{k}}}=\prod\limits_{t\ge i>k}{\frac{\partial {{h}_{i}}}{\partial {{h}_{i-1}}}}=\prod\limits_{t\ge i>k}{W_{rec}^{T}}diag({\sigma }'({{h}_{i-1}}))
	\end{equation}
	
	These equations are obtained by writing the gradient as a sum of products. $\frac{\partial {}^{+}{{h}_{k}}}{\partial \theta }$ is the immediate partial derivative of state ${{h}_{k}}$ with respect to $\theta $, and ${{h}_{k-1}}$ is a constant with respect to $\theta $. According to Eq.(8), the value of any $i$-th row of matrix $\left( \frac{{{\partial }^{+}}{{h}_{k}}}{\partial {{W}_{rec}}} \right)$ is $\sigma ({{h}_{k-1}})$. Eq.(11) also provides a Jacobian matrix $\frac{\partial {{h}_{i}}}{\partial {{h}_{i-1}}}$ for the specific parameters in Eq.(8), where $diag$ transforms the vector into a diagonal matrix, and ${\sigma }'$ is the derivative of $\sigma $. Any gradient component $\frac{\partial {{\varepsilon }_{t}}}{\partial \theta }$ is also obtained by summation (see Eq.(10)), and its term is called temporal contribution or temporal component. It can be seen that each temporal contribution $\frac{\partial {{\varepsilon }_{t}}}{\partial {{h}_{t}}}\frac{\partial {{h}_{t}}}{\partial {{h}_{k}}}\frac{\partial {}^{+}{{h}_{k}}}{\partial \theta }$ can measures the influence of $\theta $ at step $k$ on the cost at step $t>k$. The factor $\frac{\partial {{h}_{t}}}{\partial {{h}_{k}}}$ (Eq.(11)) can transmit the error transmitted from step $t$ back to step $k$ in time. Tomas Mikolov et al. further distinguish between long-term and short-term contributions. The long-term is defined as the temporal component at $k\ll t$ and the short-term refers to the remaining part.
	
	2)Exploding and vanishing gradients
	
	In the introduction of Bengio et al. \cite{6},the gradient explosion problem refers to a substantial increase in the gradient norm during training. This is caused by the explosion of long-term components, which grow exponentially faster than short-term components. The vanishing gradient problem refers to the opposite process. When the exponentially increasing long-term component quickly approaches the norm 0, the model cannot learn the dependence between a wide range of events. The following analyzes the problem of gradient disappearance and explosion from a structural perspective.
	
	Tomas Mikolov et al. extended the similar derivation of Bengio et al. and considered the case of a single hidden unit. If the linear form of the model is considered (i.e. $\sigma $ is set to the identity function in equation(8)), the power iteration method can be used to analyze the product of the Jacobian matrix and obtain the strict conditions when the gradient explodes or disappears. Define $\rho $ as the spectral radius of the recurrent weight matrix ${{W}_{rec}}$. When $\rho <1$, the gradient will not disappear or explode, and when $\rho >1$, the gradient will disappear or explode. This result is then extended to a nonlinear function $\sigma $ through singular values, where $\left| {\sigma }'(x) \right|$ is bounded, $\left\| diag({\sigma }'({{x}_{k}}) \right\|\le \gamma \in R$. First prove that when ${{\lambda }_{1}}<\frac{1}{\gamma }$, ${{\lambda }_{1}}$ is the largest singular value of ${{W}_{rec}}$, and the disappearing gradient problem will occur. By reversing this proof, the necessary condition for gradient explosion can be obtained, that is, the maximum singular value ${{\lambda }_{1}}$ is greater than $\frac{1}{\gamma }$ (otherwise the long-term component will disappear instead of exploding). Let $\gamma =1$ for the $tanh$ function and $\gamma ={}^{1}\!\!\diagup\!\!{}_{4}\;$ for the $sigmoid$ function.

\subsubsection{Long Short-Term Memory Unit}
LSTM was proposed in \cite{7} \cite{8}, and then improved and promoted in \cite{9}. On many issues, LSTM has achieved considerable success and has been widely used.Each $j$-th unit of LSTM contains a memory unit $c_{t}^{j}$ at time step $t$, then the output $h_{t}^{j}$ (or activation) of the LSTM unit is expressed as \cite{1}:

\begin{equation}
h_{t}^{j}=o_{t}^{j}tanh(c_{t}^{j})
\end{equation}

where $o_{t}^{j}$ is the output gate, which adjusts the amount of output content of the memory unit. The calculation process of the output gate is as follows:

\begin{equation}
o_{t}^{j}=\sigma {{({{W}_{o}}{{x}_{t}}+{{U}_{o}}{{h}_{t-1}}+{{V}_{o}}{{c}_{t}})}^{j}}
\end{equation}

where $\sigma $ is a logistic sigmoid function, and ${{V}_{o}}$ is a diagonal matrix.

Update the memory unit $c_{t}^{j}$ by partially forgetting the existing memory and then adding a new memory unit $\tilde{c}_{t}^{j}$:

\begin{equation}
c_{t}^{j}=f_{t}^{j}c_{t-1}^{j}+i_{t}^{j}\tilde{c}_{t}^{j}
\end{equation}

where the new memory unit is:

\begin{equation}
\tilde{c}_{t}^{j}=tanh{{({{W}_{c}}{{x}_{t}}+{{U}_{c}}{{h}_{t-1}})}^{j}}
\end{equation}

The function of the forgetting gate $f_{t}^{j}$ is to adjust the degree of forgetting of the existing memory, and the degree of adding content to the new memory unit is adjusted by the input gate $i_{t}^{j}$. The calculation process of the forget gate and input gate is as follows:

\begin{equation}
f_{t}^{j}=\sigma {{({{W}_{f}}{{x}_{t}}+{{U}_{f}}{{h}_{t-1}}+{{V}_{f}}{{c}_{t-1}})}^{j}}
\end{equation}

\begin{equation}
i_{t}^{j}=\sigma {{({{W}_{i}}{{x}_{t}}+{{U}_{i}}{{h}_{t-1}}+{{V}_{i}}{{c}_{t-1}})}^{j}}
\end{equation}

Note that ${{V}_{f}}$ and ${{V}_{i}}$ are diagonal matrices.

Unlike the traditional recurrent unit that covers content at each time step (see Eq.()), the LSTM unit can decide whether to keep the existing memory by introducing a gate. Intuitively speaking, if the LSTM unit can detect important features from the input sequence at an early stage, it can easily carry this information (the existence of features) over long distances to capture potential long-distance dependencies.

\begin{figure}[H]
	\centering
	\includegraphics[width=0.7\textwidth]{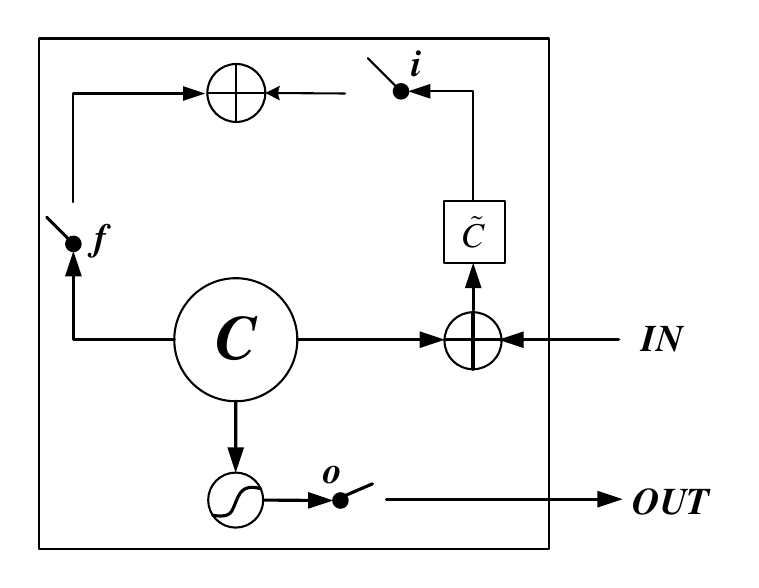}
	\caption{Long Short-Term Memory.}
	\label{fig:lstm_structure}
\end{figure}

$i$, $f$ and $o$ are input gate, forget gate and output gate respectively, $c$ and $\tilde{c}$ represent memory unit and new memory unit.

\subsubsection{Gated Recurrent Unit}
Cho et al. \cite{10} proposed a gated recurrent unit (GRU), which enables each recurrent unit to adaptively capture the dependencies of different time scales. Similar to the LSTM unit, the GRU has a gating unit to adjust the information flow inside the unit, but there is no separate memory unit.

The activation $h_{t}^{j}$ of the GRU at time step $t$ is the linear interpolation between the previously activated $h_{t-1}^{j}$ and the pre-activated $\tilde{h}_{t}^{j}$:

\begin{equation}
h_{t}^{j}=(1-z_{t}^{j})h_{t-1}^{j}+z_{t}^{j}\tilde{h}_{t}^{j}
\end{equation}

where the update gate $z_{t}^{j}$ determines the extent to which the unit updates its activation or content. The calculation method of the update gate is as follows:

\begin{equation}
z_{t}^{j}=\sigma {{({{W}_{z}}{{x}_{t}}+{{U}_{z}}{{h}_{t-1}})}^{j}}
\end{equation}

The process of taking a linear sum between the existing state and the new calculation state is similar to the LSTM unit. However, GRU does not have any mechanism to control the extent of its state exposure, but exposes the entire state every time.

The calculation method of pre-activation $\tilde{h}_{t}^{j}$ is similar to RecurrentNNs (see Eq.(2)):

\begin{equation}
\tilde{h}_{t}^{j}=tanh{{(W{{x}_{t}}+U({{r}_{t}}\odot {{h}_{t-1}}))}^{j}}
\end{equation}

where ${{r}_{t}}$ is a set of reset gates and $\odot $ is an element-wise multiplication. When $r_{t}^{j}$ is close to 0, the reset gate effectively makes the unit work like reading the first symbol of the input sequence, making it forget the previously calculated state.

The calculation method of reset gate $r_{t}^{j}$ is similar to update gate:

\begin{equation}
r_{t}^{j}=\sigma {{({{W}_{r}}{{x}_{t}}+{{U}_{r}}{{h}_{t-1}})}^{j}}
\end{equation}

\begin{figure}[H]
	\centering
	\includegraphics[width=0.7\textwidth]{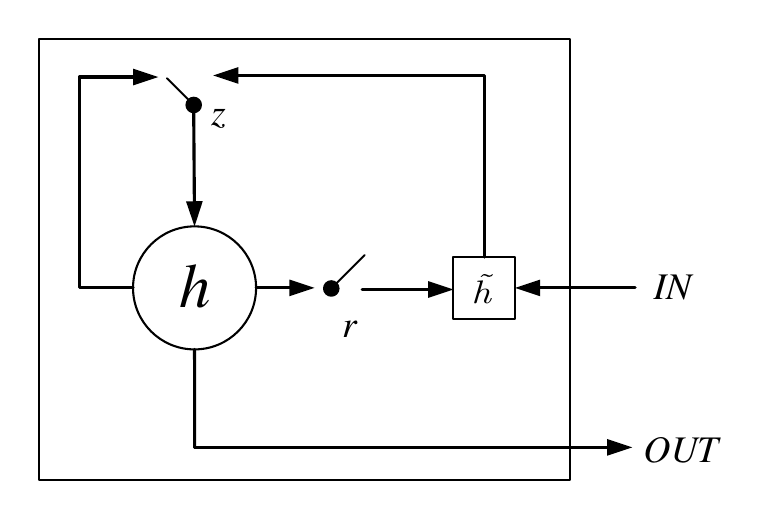}
	\caption{GRU.}
	\label{fig:gru_structure}
\end{figure}

\subsubsection{The Depth-gated LSTM}
\begin{figure}[H]
	\centering
	\includegraphics[width=0.5\textwidth]{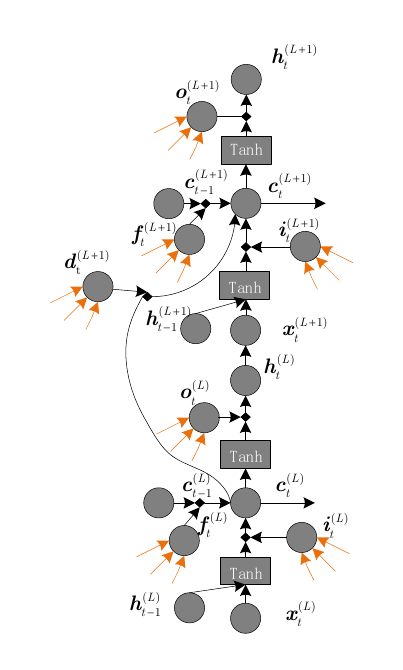}
	\caption{Schematic diagram of DGLSTM. Note the additional connections between the memory cells in the lower and upper layers in the DGLSTM.}
	\label{fig:dglstm_structure}
\end{figure}

Depth-Gated LSTM (DGLSTM) is shown in Fig.~\ref{fig:dglstm_structure} \cite{11}. It has a depth gate, which connects the memory cell $c_{t}^{(L+1)}$ in the lower layer $L+1$ and the memory cell $c_{t}^{L}$ in the upper layer $L$. The depth gate controls the flow from the lower memory unit directly to the upper memory unit. The gate function at level $L+1$ at time step $t$ is a logistic function as follows:

\begin{equation}
d_{t}^{(L+1)}=\sigma (b_{d}^{(L+1)}+W_{xd}^{(L+1)}x_{t}^{(L+1)}+w_{cd}^{(L+1)}\odot c_{t-1}^{(L+1)}+w_{ld}^{(L+1)}\odot c_{t}^{(L)})
\end{equation}

where $b_{d}^{(L+1)}$ is the bias, and $W_{xd}^{(L+1)}$ is the weight matrix associated with the depth gate and the input of this layer. The past memory unit generates correlation through the weight vector $w_{cd}^{(L+1)}$. Use the weight vector $w_{ld}^{(L+1)}$ to associate the lower memory. Note that if the lower and upper memory cells have different dimensions, $w_{ld}^{(L+1)}$ should be a matrix instead of a vector.

Using the depth gate, DGLSTM calculates the memory cell process at level $L+1$ as follows:

\begin{equation}
c_{t}^{(L+1)}=d_{t}^{(L+1)}\odot c_{t}^{(L)}+f_{t}^{(L+1)}\odot c_{t-1}^{(L+1)}+i_{t}^{(L+1)}\odot tanh(W_{xc}^{(L+1)}x_{t}^{(L+1)}+W_{hc}^{(L+1)}h_{t-1}^{(L+1)})
\end{equation}

In DGLSTM, the general LSTM structure ((12),(13),(16),(17)) is used, but DGLSTM uses superscript $L+1$ to represent the operation at layer $L+1$.

The idea of gated linear correlation can also be used to connect the first layer memory unit $c_{t}^{(1)}$ with the feature observation vector $x_{t}^{(0)}$. In this case, when $L=0$, the depth gate calculation process is as follows:

\begin{equation}
d_{t}^{(1)}=\sigma (b_{d}^{(1)}+W_{xd}^{(1)}x_{t}^{(1)}+w_{cd}^{(1)}\odot c_{t-1}^{(1)})
\end{equation}

and the memory cell is computed as:

\begin{equation}
c_{t}^{(1)}=d_{t}^{(1)}\odot (W_{xd}^{(1)}x_{t}^{(0)})+f_{t}^{(1)}\odot c_{t-1}^{(1)}+i_{t}^{(1)}\odot tanh(W_{xc}^{(1)}x_{t}^{(0)}+W_{hc}^{(1)}h_{t-1}^{(1)})
\end{equation}

\subsubsection{Structure-evolving LSTM}
Structure-evolving LSTM \cite{12} \cite{13} consists of five gates: input gate ${{g}^{u}}$, forget gate ${{g}^{f}}$, adaptive forget gate ${{\bar{g}}^{f}}$, memory gate ${{g}^{c}}$, output gate ${{g}^{o}}$ and edge gate $p$. $1$ is the indicator function. ${{W}^{e}}$ represents the weight parameter of the recurrent edge gate. ${{W}^{u}}$, ${{W}^{f}}$, ${{W}^{c}}$, ${{W}^{o}}$ are the recurrent gate weight matrices specified for the input features, and ${{U}^{u}}$, ${{U}^{f}}$, ${{U}^{c}}$, ${{U}^{o}}$ are the recurrent gate weight matrices of the hidden state of each node. ${{U}^{un}}$, ${{U}^{fn}}$, ${{U}^{cn}}$, ${{U}^{on}}$ are the weight parameters specified for the state of neighboring nodes. The structure-evolving LSTM unit assigns different forgetting gates to different neighboring nodes through the effect of the input state of the current node and its hidden state, which are defined as $\bar{g}_{ij}^{f}$,$j\in {{\mathcal{N}(i) }_{{{G}^{(t)}}}}(i)$. Adjacent nodes have different effects on the updated memory state $c_{i}^{t+1}$ and hidden state $h_{j}^{t+1}$. Use the weight matrix ${{W}^{e}}$ to weight the adaptive forgetting gate $\bar{g}_{ij}^{f}$ to calculate the merging probability ${{p}_{ij}}$ of each pair of graph nodes $<i,j>\in {{\varepsilon }^{t}}$. Intuitively, the adaptive forgetting gate is used to identify the significant correlations of different node pairs. Therefore, an adaptive forget gate is used to estimate the merging probability of each pair of graphs. The new hidden state, memory state, and edge gate (i.e., the merging probability of each pair of connected nodes) in graph ${{G}^{(t)}}$ can be calculated as follows:

\[g_{i}^{u}=\delta ({{W}^{u}}x_{i}^{t}+{{U}^{u}}h_{i}^{t-1}+{{U}^{un}}\bar{h}_{i}^{t-1}+{{b}^{u}})\]

\[\bar{g}_{ij}^{f}=\delta ({{W}^{f}}x_{i}^{t}+{{U}^{fn}}h_{j}^{t-1}+{{b}^{f}})\]

\[g_{i}^{f}=\delta ({{W}^{f}}x_{i}^{t}+{{U}^{f}}h_{i}^{t-1}+{{b}^{f}})\]

\[g_{i}^{o}=\delta ({{W}^{o}}x_{i}^{t}+{{U}^{o}}h_{i}^{t-1}+{{U}^{on}}\bar{h}_{i}^{t-1}+{{b}^{o}})\]

\[g_{i}^{c}=tanh({{W}^{c}}x_{i}^{t}+{{U}^{c}}h_{i}^{t-1}+{{U}^{cn}}\bar{h}_{i}^{t-1}+{{b}^{c}})\]

\[{{c}_{i,t}}=\frac{\sum\nolimits_{j\in {{\mathcal{N}(i) }_{g}}(i)}{(1({{q}_{j}}=1)\bar{g}_{ij}^{f}\odot c_{j}^{t}+1({{q}_{j}}=0)\bar{g}_{ij}^{f}\odot c_{j}^{t-1})}}{\left| {{\mathcal{N}(i) }_{{{G}^{(t)}}}}(i) \right|}+g_{i}^{f}\odot c_{i}^{t-1}+g_{i}^{u}\odot g_{i}^{c}\]

\[h_{i}^{t}=tanh(g_{i}^{o}\odot c_{i}^{t})\]

\begin{equation}
p_{ij}^{t}=\delta ({{W}^{e}}\bar{g}_{ij}^{f})
\end{equation}

where $\sigma$ is the logistic sigmoid function, and $\odot$ represents the point-wise product. ${{q}_{j}}$ is a visit flag, used to indicate whether the graph node ${{v}_{i}}$ has been updated, where ${{q}_{j}}$ is set to 1 if it is updated, otherwise it is set to 0. ${{\mathcal{N}(i) }_{{{G}^{(t)}}}}(i)$ represents the number of adjacent graph nodes of node $i$. As a memory system, this model writes information into the memory state and records it in sequence by each graph node, and then uses it to communicate with subsequent graph nodes and the hidden state of the previous LSTM layer. Therefore, the merge probability $\left\{ {{p}_{ij}} \right\},<i,j>\in {{\varepsilon }^{t}}$ can be effectively learned and applied in the $t+1$-th layer to generate a new high-level graph structure ${{G}^{(t+1)}}$. During training, the merging probability of graph edges is supervised by approximating the final graph structure of a specific task, such as the connection of the final semantic region of image parsing. Use the back propagation method to train all weight matrices.

The advantages of this model include: (1) Multi-level graph representation of data and LSTM network parameters can be learned simultaneously in an end-to-end manner. (2) Realize that the underlying multilayer graph structure evolves with parameter learning.

Structure-evolving LSTM is also called graph LSTM, which can be classified in graph RNNs, so it won't be repeated in Chapter 3.2.

\subsubsection{Layer Trajectory LSTM}
\begin{figure}[H]
	\centering
	\includegraphics[width=0.6\textwidth]{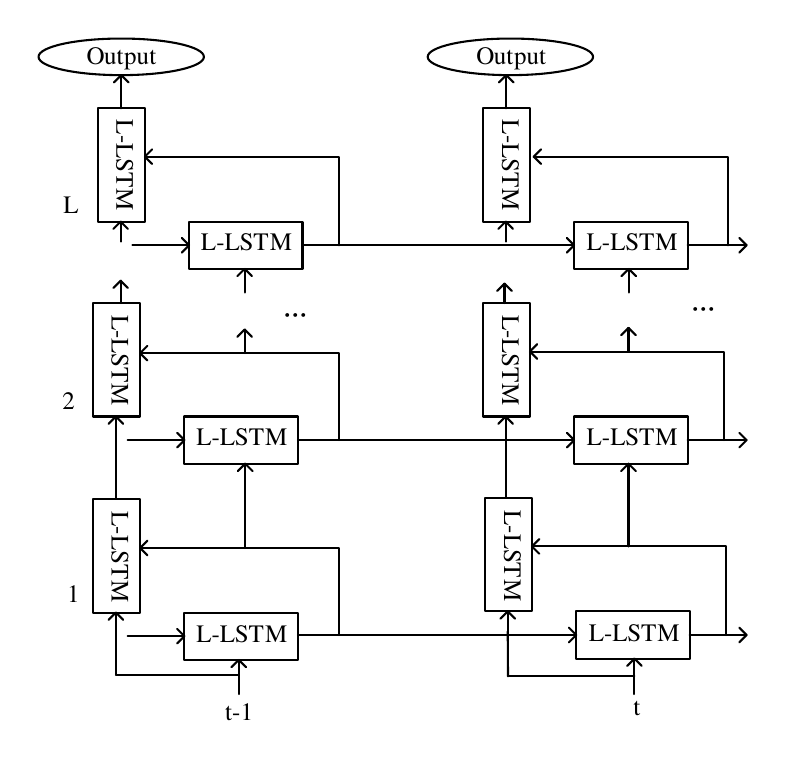}
	\caption{ltLSTM flowchart.}
	\label{fig:ltlstm_structure}
\end{figure}

Jinyu Li et al. proposed layer trajectory LSTM (ltLSTM) \cite{14} \cite{15}. Its output can satisfy both temporal modeling and senone classification. It uses all output layers of general LSTM to construct a layer LSTM (L-LSTM), as shown in Fig.~\ref{fig:ltlstm_structure}. The weights are not shared between layers because sharing doesn't reduce computational complexity. The general LSTM is used for the purpose of temporal modeling via temporal recurrence, while the L-LSTM scans multiple outputs of general LSTM layers and then uses the summarized layer trajectory information to perform the final senone classification. The L-LSTM layer provides a gated path from the output layer to the bottom layer during operation, so the problem of gradient disappearance is reduced.

L-LSTM is used to scan the general LSTM output of all layers at the current time step to obtain summary layer trajectory information for senone classification. L-LSTM has no time recurrence, it only exists between general LSTMs at different time steps.

In ltLSTM, the general LSTM formula is still as shown in section 2.2.2. As shown in Fig.~\ref{fig:ltlstm_structure}, there is no time recurrence between L-LSTMs at different time steps. Therefore, the formula of L-LSTM only has a cross-layer recurrence:

\[j_{t}^{l}=\sigma (U_{jh}^{l}h_{t}^{l}+U_{jg}^{l}g_{t}^{l-1}+q_{j}^{l}\odot c_{t}^{l-1}+d_{j}^{l})\]

\[e_{t}^{l}=\sigma (U_{eh}^{l}h_{t}^{l}+U_{eg}^{l}g_{t}^{l-1}+q_{e}^{l}\odot c_{t}^{l-1}+d_{e}^{l})\]

\[c_{t}^{l}=e_{t}^{l}\odot c_{t}^{l-1}+j_{t}^{l}\odot \phi (U_{sh}^{l}h_{t}^{l}+U_{sg}^{l}g_{t}^{l-1}+d_{s}^{l})\]

\[v_{t}^{l}=\sigma (U_{vh}^{l}h_{t}^{l}+U_{vg}^{l}g_{t}^{l-1}+q_{v}^{l}\odot c_{t}^{l}+d_{v}^{l})\]

\begin{equation}
g_{t}^{l}=v_{t}^{l}\odot \phi (c_{t}^{l})
\end{equation}

The vectors $j_{t}^{l}$, $v_{t}^{l}$, $e_{t}^{l}$, and $m_{t}^{l}$ are the input gate, output gate, forget gate and memory unit of L-LSTM, respectively. $g_{t}^{l}$ is the output of L-LSTM. The matrices $U_{.h}^{l}$ and $U_{.g}^{l}$ are the weight matrices of input $h_{t}^{l}$ and output $g_{t}^{l}$, respectively. $d_{.}^{l}$ is the bias vector. $q_{j}^{l}$, $q_{v}^{l}$, $q_{e}^{l}$ are vector parameters related to peephole connection.

The $g_{t}^{l-1}$-layer recurrence of L-LSTM in ltLSTM is the biggest difference from general LSTM. L-LSTM takes the general LSTM output $h_{t}^{l}$ of the current layer as input, which is in contrast with the $x_{t}^{l}$ in the general LSTM.

Advantages of LtLSTM: Since the forward propagation of the general LSTM at the next time step has nothing to do with the calculation of the L-LSTM layer at the current time step, the forward propagation of the general LSTM and L-LSTM can process two separate threads in parallel \cite{16}, and the network calculation time runs layer by layer and frame by frame. Another advantage of decoupling temporal and layer operations in ltLSTM is that batch processing \cite{17} can be used to evaluate L-LSTM, which can improve the running time of feedforward deep neural networks by simultaneously evaluating the network scores of multiple time frames. But this batch processing cannot be applied to general LSTM.

\subsubsection{Multiplicative LSTM}
Since LSTM and multiplicative RNN \cite{18} (mRNN) structures are complementary, Krause et al. proposed multiplicative LSTM (mLSTM) \cite{19}, which is a hybrid structure that combines the factorization of mRNN hidden variable to hidden variable conversion and LSTM gated framework. The structure of mRNN and LSTM can add connections from the intermediate state $m_t$ of mRNN to each gate unit in LSTM, resulting in the following system:

\[{{m}_{t}}=({{W}_{mx}}{{x}_{t}})\odot ({{W}_{mh}}{{h}_{t-1}})\]

\[{{\hat{h}}_{t}}={{W}_{hx}}{{x}_{t}}+{{W}_{hm}}{{m}_{t}}\]

\[{{i}_{t}}=\sigma ({{W}_{ix}}{{x}_{t}}+{{W}_{im}}{{m}_{t}})\]

\[{{o}_{t}}=\sigma ({{W}_{ox}}{{x}_{t}}+{{W}_{om}}{{m}_{t}})\]

\begin{equation}
{{f}_{t}}=\sigma ({{W}_{fx}}{{x}_{t}}+{{W}_{fm}}{{m}_{t}})
\end{equation}

The goal of this structure is to transform the flexible input dependency of mRNN and combine it with the long time lag and information control of LSTM. The gated unit of LSTM can more easily control (or bypass) the complex conversion of the factorization hidden variable weight matrix. The additional sigmoid input and forget gate in the LSTM unit exhibit a more flexible input-dependent conversion function than the regular mRNN.

The advantages of this model include: (1) mLSTM does not use nonlinear recurrent depth, so it is easier to parallelize. (2) For each possible input, there is a different recurrent transfer function, so it can better express the autoregressive density estimation.

There are also many variants of LSTM, which are interspersed in different types of RNNs, so they won't be repeated in this section 2.2.

\subsection{Convolutional RNNs}
Next, the various network architectures of the Convolutional RNNs will be described in detail.

\subsubsection{Recurrent Convolutional Network}
\begin{figure}[H]
	\centering
	\includegraphics[width=0.9\textwidth]{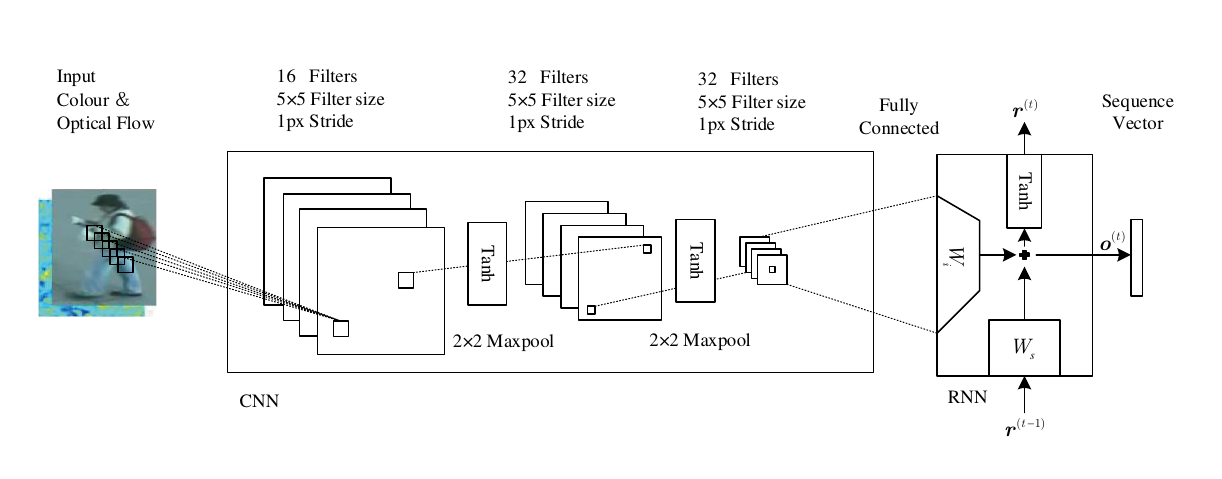}
	\caption{The structure of the proposed CNN and recurrent layer.}
	\label{fig:recurrent_conv_structure}
\end{figure}

In the recurrent convolutional network, each frame of the input video sequence is first processed by a convolutional neural network (CNN) to generate a feature vector representing the appearance of a person at a specific time step. Then, before using the temporal pooling to combine the output of all time steps, the recurrent layer is used to let the information flow between each time step. The temporal pooling can make the network use a feature vector to represent an arbitrarily long video sequence, and the recurrent layer can make the network make better use of the time information in the sequence. In order to train the feature extraction network for re-recognition, McLaughlin et al. used the Siamese network \cite{20}. Therefore, the video re-identification method based on recurrent convolutional network \cite{21} has higher accuracy. The structure of the recurrent convolutional network is detailed below.Let $s={{s}^{\mathrm{(1)}}}\mathrm{,}\ldots \mathrm{,}{{s}^{\mathrm{(}\text{T}\mathrm{)}}}$ be a video sequence of length $T$, which consists of full-body images of a person, where ${{s}^{(t)}}$ is the image at time step $t$. Each image ${{s}^{(t)}}$ generates a vector ${{f}^{\mathrm{(}\text{t}\mathrm{)}}}=\text{C}\mathrm{(}{{s}^{\mathrm{(}\text{t}\mathrm{)}}}\mathrm{)}$ through CNN, where ${{f}^{\mathrm{(}\text{t}\mathrm{)}}}$ is the vector representation of the activation map of the CNN output layer. Then the vector ${{f}^{\mathrm{(}\text{t}\mathrm{)}}}$ is passed to the recurrent layer, where it is projected to a low-dimensional feature space and combined with the information at the previous moment. One issue to note is that the parameters of CNN are shared at all time steps, which means that each input frame is processed by the same feature extraction network. Therefore, dropout is used between the CNN and the recurrent layer to reduce overfitting. Fig.~\ref{fig:recurrent_conv_structure} shows the complete details of the CNN and recurrent layer structure. Among them, ${{r}^{\mathrm{(}\text{t}\mathrm{)}}}$ is state variable of RNNs at time step $t$, and ${{o}^{\mathrm{(}\text{t}\mathrm{)}}}$ is the sequence output vector at time step $t$.

Join the recurrent connection after the CNN, as shown below:

\begin{equation}
{{o}^{\mathrm{(}\text{t}\mathrm{)}}}={{\text{W}}_{\text{i}}}{{f}^{\mathrm{(}\text{t}\mathrm{)}}}+{{\text{W}}_{s}}{{r}^{\mathrm{(}\text{t}-\mathrm{1)}}}
\end{equation}

\begin{equation}
{{r}^{\mathrm{(}\text{t}\mathrm{)}}}=\mathrm{Tanh}\mathrm{(}{{o}^{\mathrm{(}\text{t}\mathrm{)}}}\mathrm{)}
\end{equation}

The output ${{o}^{\mathrm{(}\text{t}\mathrm{)}}}\in {{\mathbb{R}}^{\text{e}\times \mathrm{1}}}$ is a linear combination of vectors at each time step, ${{f}^{\mathrm{(}\text{t}\mathrm{)}}}\in {{\mathbb{R}}^{\text{N}\times \mathrm{1}}}$ contains the information of the current input image, and ${{r}^{\mathrm{(}\text{t}-\mathrm{1)}}}\in {{\mathbb{R}}^{\text{e}\times \mathrm{1}}}$ contains the state information of the RNNs at the previous time step. Use the fully connected layers ${{\text{W}}_{\text{i}}}\in {{\mathbb{R}}^{\text{e}\times \text{N}}}$ and ${{\text{W}}_{s}}\in {{\mathbb{R}}^{\text{e}\times \text{e}}}$ to calculate the output, $e$ is the dimension of the feature embedding space, and $N$ is the dimension of the activation mapping vector output of the final layer of CNN. Note that the parameter matrix ${{\text{W}}_{\text{i}}}$ is a non-square matrix, which means that the final layer activation map of the CNN is projected into a vector in the lower-dimensional feature embedding space. The state variable ${{r}^{\mathrm{(}\text{t}\mathrm{)}}}$ of RNNs is initialized to a zero vector at the first time step ${{r}^{\mathrm{(0)}}}$, and is transferred between each time step through the Tanh nonlinear function. Add a temporal pool after the recurrent layer, so that information can be aggregated at any time step, long-term information existing in the sequence can be captured, and the feature vector representation of the sequence of any length can be learned.

\subsubsection{Spatially Supervised Recurrent Convolutional Neural Networks}
\begin{figure}[H]
	\centering
	\includegraphics[width=0.8\textwidth]{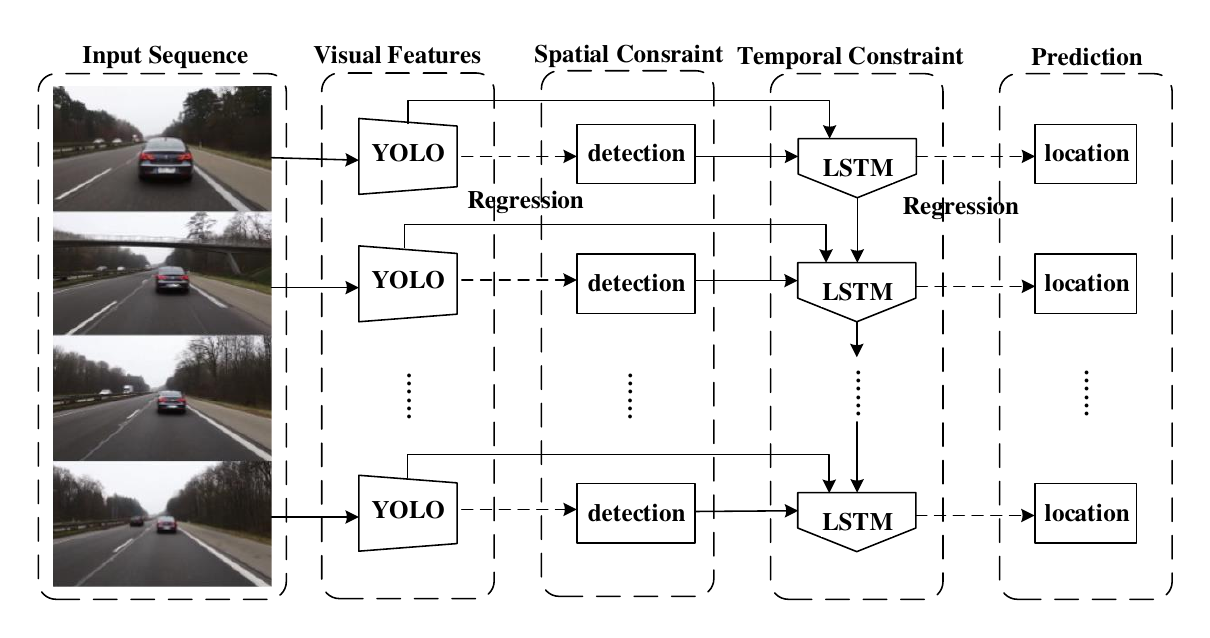}
	\caption{Simplified system overview and tracking procedures.}
	\label{fig:spatial_supervised_rcnn}
\end{figure}

The human and object tracking process based on spatially supervised recurrent convolutional neural network is shown in Fig.~\ref{fig:spatial_supervised_rcnn} \cite{22}. Guanghan Ning et al. use the YOLO \cite{23} \cite{24} (a framework for target detection problems) system to collect visual features and perform preliminary position inference; then use LSTM in the next stage. LSTM can capture a large range of spatial dependencies and is suitable for human and sequence processing of the object tracking process. This model belongs to a deep neural network. It takes the original video frame as input and returns the coordinates of the bounding box of the tracked object in each frame. This model formalizes the complete tracking conditional probability distribution as:

\begin{equation}
\text{p}\left( {{B}_{\mathrm{1}}}\mathrm{,}{{B}_{\mathrm{2}}}\mathrm{,}\ldots \mathrm{,}{{B}_{\text{T}}}\left| {{X}_{\mathrm{1}}}\mathrm{,}{{X}_{\mathrm{2}}}\mathrm{,}\ldots \mathrm{,}{{X}_{\text{T}}} \right. \right)=\prod\limits_{\text{t}=\mathrm{1}}^{\text{T}}{\text{p}\mathrm{(}\left. {{B}_{\text{t}}} \right|{{B}_{<\text{t}}}\mathrm{,}{{X}_{\le \text{t}}}\mathrm{)}}
\end{equation}

where $B_t$ and ${{X}_{t}}$ are the target and the position of the input frame at time step $t$, respectively. ${{B}_{<\text{t}}}$ is the history information of all previous positions before time step $t$, and ${{X}_{\le \text{t}}}$ is the history information of the input frame up to time step $t$. The spatially supervised recurrent convolutional neural network can effectively process spatiotemporal information and infer the location of the region. The YOLO deep convolutional neural network is extended to the spatiotemporal domain by using the RecurrentNNs.

The advantages of this model include: (1) A modular neural network is introduced, which can be trained end-to-end through a gradient-based learning method. (2) Compared with the existing trackers based on convolutional networks, the framework proposed by Guanghan Ning et al. extends the neural network analysis to the spatio-temporal domain, which can achieve effective visual object tracking. (3) The proposed network is accurate and efficient, with low complexity.

\subsubsection{Convolutional Gated Recurrent Neural Network}
\begin{figure}[H]
	\centering
	\includegraphics[width=0.8\textwidth]{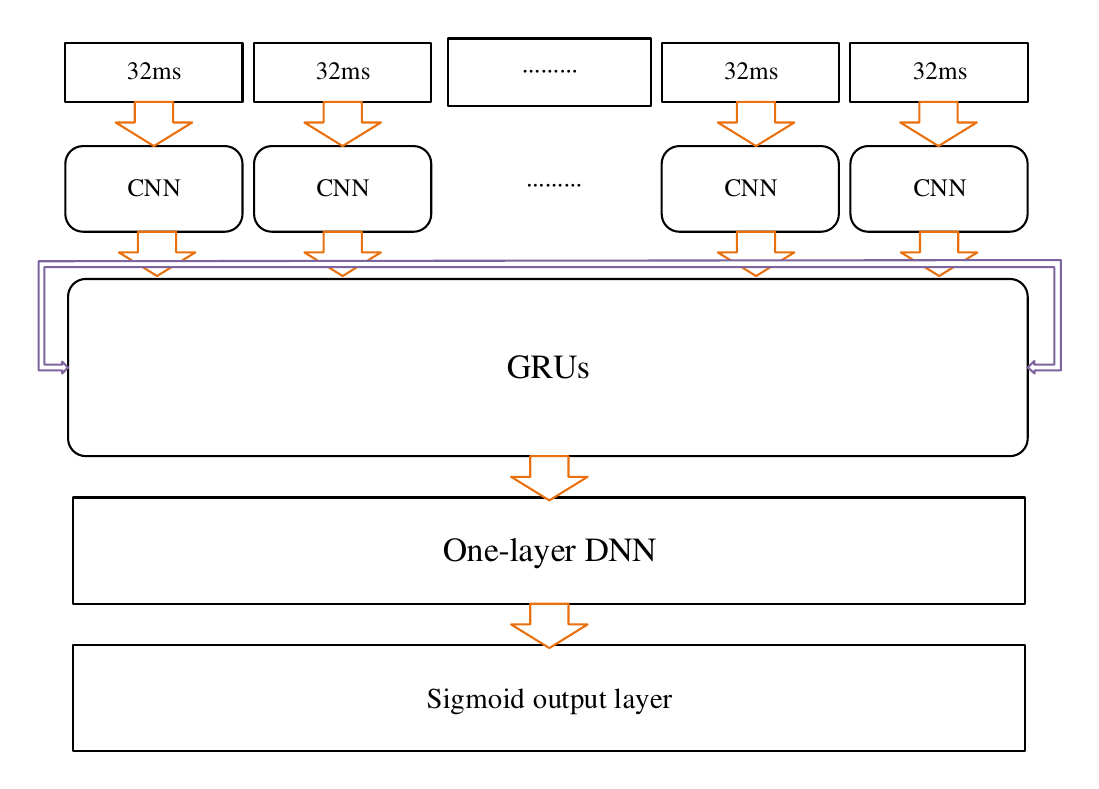}
	\caption{The framework of convolutional gated recurrent neural network for audio tagging.}
	\label{fig:conv_gru_framework}
\end{figure}

Fig.~\ref{fig:conv_gru_framework} shows the framework of a convolutional gated recurrent neural network applied to audio tagging \cite{25} \cite{26} \cite{27} \cite{28}. CNN is a feature extractor that can extract robust features against background noise, especially the original audio waveform, through a maximum pooling operation. Then, the extracted robust features are fed into Gated Recurrent Unit-RNN (GRU-RNN) to learn long-term audio mode. GRU-RNN can select relevant information from the long-term context of each audio event. In order to use the information of the future, this system has designed a bidirectional GRU-RNN. Finally, under the condition of the sigmoid output activation function, the output of GRU-RNN is mapped to the posterior probability of the target audio event through a feedforward neural layer. The framework is flexible enough to be applied to any type of feature, especially raw audio waveforms. Since the original waveform has many values, it will cause high-dimensional problems. However, the proposed CNN can learn under short windows such as the Short-Time Fourier Transform (STFT) process, and similar Fast Fourier Transform (FFT) basis vector set or mel-like filters can also be used for learning the original waveform. Finally, a single-layer feedforward deep neural network (DNN) obtains the final GRU sequence to predict the posterior probability of the tag.

Yong Xu et al. chose binary cross entropy as the loss function, which is more effective than the mean square error \cite{29}. The loss function is defined as follows:

\begin{equation}
	\mathrm{E}=-\sum\limits_{\text{n}=\mathrm{1}}^{\text{N}}{\left\| {{T}_{\text{n}}}\mathrm{log}{{{\hat{T}}}_{\text{n}}} \right.}+\mathrm{(1}-{{T}_{\text{n}}}\mathrm{)log(1}-{{\hat{T}}_{\text{n}}}\left. \mathrm{)} \right\|
\end{equation}

\begin{equation}
	{{\hat{T}}_{\text{n}}}={{\mathrm{(1}+\mathrm{exp(}-O\mathrm{))}}^{-\mathrm{1}}}
\end{equation}

where $\mathrm{E}$ is the binary cross entropy, ${{\hat{T}}_{\text{n}}}$ and ${{T}_{\text{n}}}$ represent the estimated value and reference tag vector at sample index $n$, respectively. Before using the sigmoid activation function, the linear output of DNN is defined as $O$.

\subsubsection{Long-term Recurrent Convolutional Network (LRCN)}
\begin{figure}[H]
	\centering
	\includegraphics[width=0.8\textwidth]{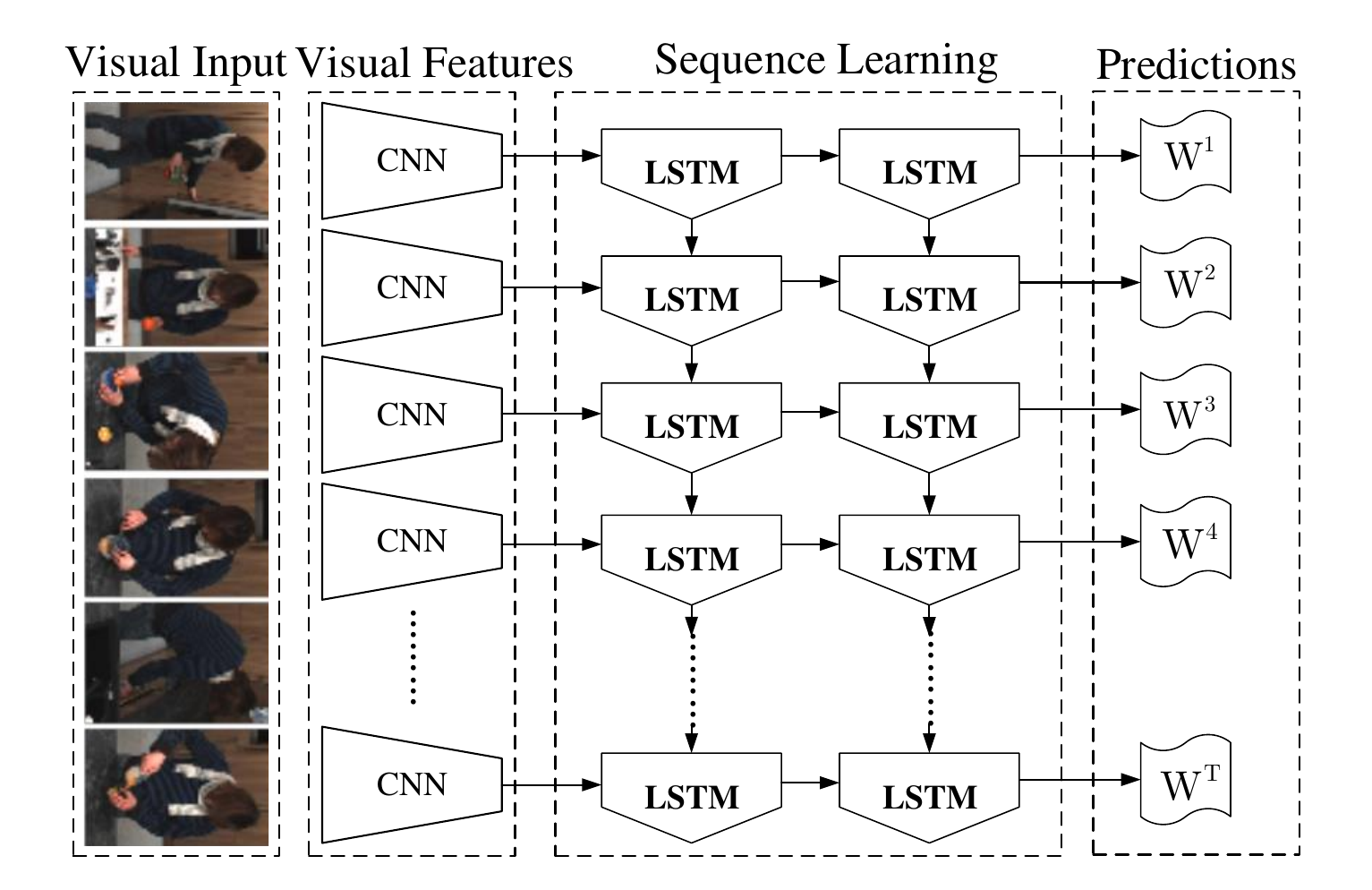}
	\caption{LRCN diagram. LRCN uses CNN (middle left) to process (possibly) variable-length visual input (left), and its output is fed to a bunch of recurrent sequence models (LSTM, middle right), which finally produces variable-length predictions (right).}
	\label{fig:lrcn_diagram}
\end{figure}

Long-term Recurrent Convolutional Network (LRCN) \cite{30} \cite{31} combines a deep hierarchical visual feature extractor (such as CNN) with a specific model that can recognize and synthesize sequential data (input or output), visual, language or other aspects of the temporal dynamic characteristics. Fig.~\ref{fig:lrcn_diagram} depicts the core of this model. The LRCN model generates a fixed-length vector representation ${{\phi }_{t}}\in {{\mathbb{R}}^{d}}$ by passing each visual input ${{v}_{t}}$ (isolated image or video frame) to a feature transformation $\phi {}_{V}({{v}_{t}})$ parameterized by $V$ (visual feature extractor parameters). After calculating the feature space representation of the visual input sequence $\langle {{\phi }_{1}},{{\phi }_{2}},\ldots ,{{\phi }_{T}}\rangle $, the sequence model can be substituted.

In the most general form, the sequence model parameterized by $W$ (sequence model parameter) maps the input ${{x}_{t}}$ and the hidden state ${{h}_{t-1}}$ at the previous time step to the output ${{z}_{t}}$ and updates the hidden state ${{h}_{t}}$. Therefore, the inference must be run sequentially (that is, run from top to bottom in the sequence learning box in Fig.~\ref{fig:lrcn_diagram}), through the sequential calculation: ${{h}_{1}}={{f}_{W}}({{x}_{1}},{{h}_{0}})={{f}_{W}}({{x}_{1}},0)$, ${{h}_{2}}={{f}_{W}}({{x}_{2}},{{h}_{1}})$ and so on, until ${{h}_{T}}$.

The last step of predicting the distribution $P({{y}_{t}})$ at time step $t$ is to perform softmax (normalized exponential function) processing on the output ${{z}_{t}}$ of the sequence model, and generate the possible output distribution at each time step in the (in this example, finite and discrete) space $C$:

\begin{equation}
P({{y}_{t}}=c)=\frac{exp({{W}_{zc}}{{z}_{t,c}}+{{b}_{c}})}{\sum\limits_{{c}'\in C}{exp({{W}_{zc}}{{z}_{t,{c}'}}+{{b}_{c}})}}
\end{equation}

The advantages of this model include: (1) The model can be flexibly applied to various visual tasks involving sequential input and output. (2) Suitable for large-scale visual learning, with end-to-end trainability. (3) It has advantages when the target concept is complex or/and training data is limited. (4) Time dynamics and convolutional perception representations can be learned at the same time.

\subsubsection{Multimodal CNN-LSTM Structure}
\begin{figure}[H]
	\centering
	\includegraphics[width=0.9\textwidth]{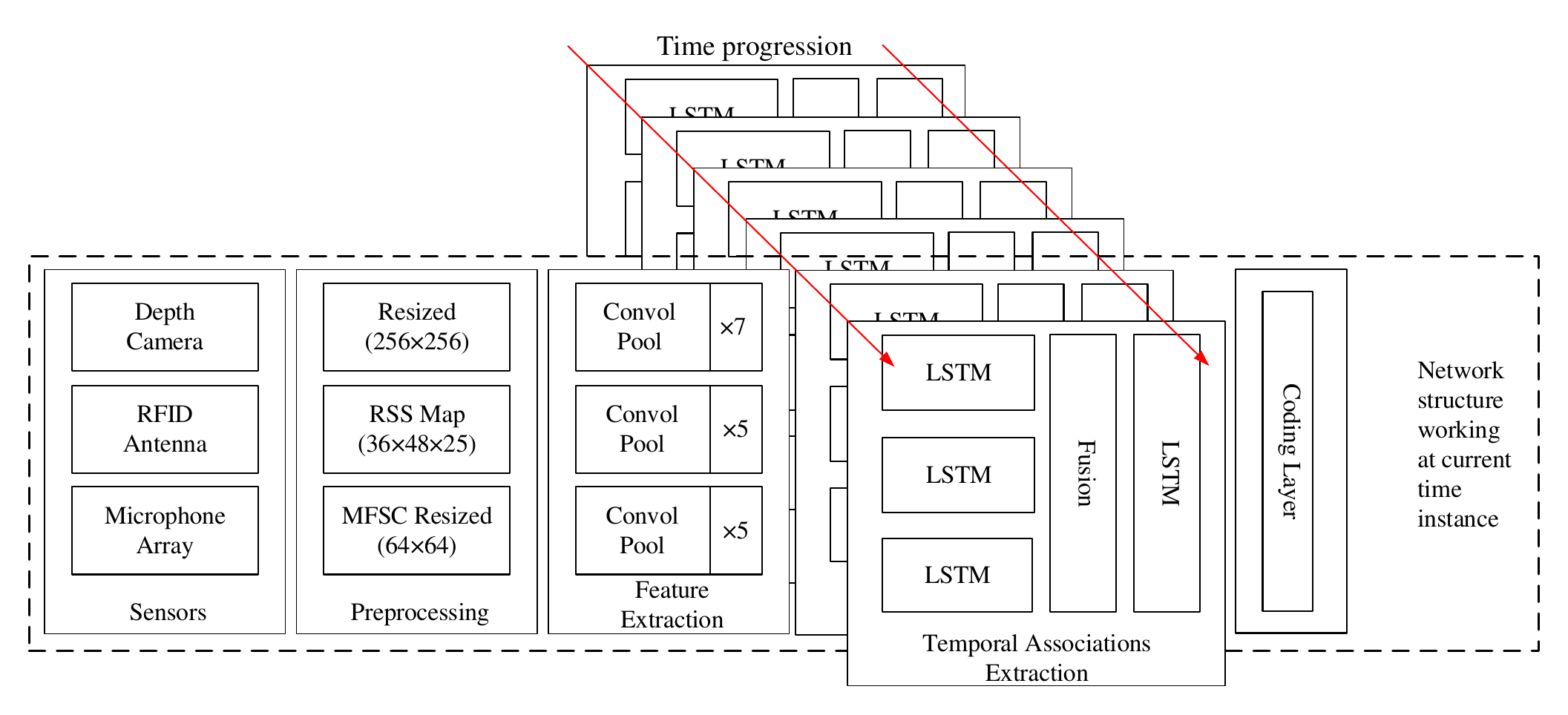}
	\caption{The structure of the concurrent activity recognition system running at a given time instance. The stacked part represents the evolution of the LSTM memory over time.}
	\label{fig:multimodal_cnn_lstm}
\end{figure}

Xinyu Li et al. \cite{32} introduce a system based on the multimodal CNN-LSTM structure, as shown in Fig.~\ref{fig:multimodal_cnn_lstm}. The system can identify concurrent activities \cite{33} \cite{34} from real data captured by multiple sensors of different types. The recognition is done in two steps. First, extract spatial and temporal features from multimodal data, and input each data type into a CNN that extracts spatial features, and then an LSTM that extracts temporal information from sensor data. The extracted features are fused in the second step. Second, use a single classifier to realize concurrent activity recognition. This system is the first to use a single model to solve concurrent activity recognition of multi-sensory data. The model is scalable, easy to train and easy to deploy. This activity recognition system consists of four main modules connected in series: sensory data preprocessing, spatial feature extraction, temporal correlation and fusion, and coding layer. The multimodal CNN-LSTM structure is responsible for feature extraction and temporal association extraction in the system.

\subsubsection{Diffusion Convolutional Recurrent Neural Network}
\begin{figure}[H]
	\centering
	\includegraphics[width=0.9\textwidth]{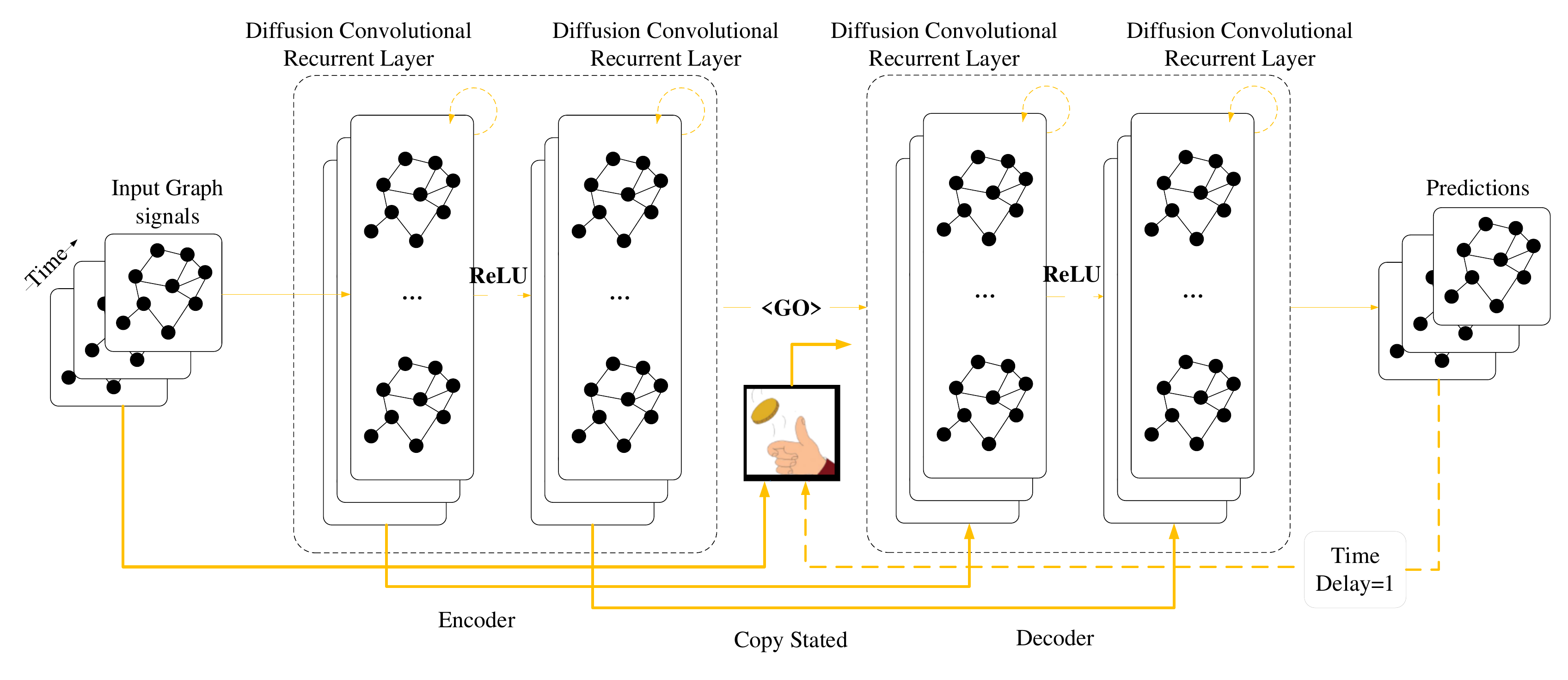}
	\caption{The system structure of the DCRNN for temporal and spatial traffic prediction. The historical time sequence is fed into the encoder, and its final state is used to initialize the decoder. The decoder makes predictions based on previous ground truth or model output.}
	\label{fig:dcrnn_architecture}
\end{figure}

Yaguang Li et al. introduced Diffusion Convolutional Recurrent Neural Network (DCRNN) \cite{35}, which is a deep learning framework for traffic prediction \cite{36} \cite{37}, which combines the spatial and temporal dependencies of traffic flow. Specifically, DCRNN uses a bidirectional random walk on the graph to capture spatial dependence, and uses an encoder-decoder architecture with scheduled sampling to capture temporal dependence. Through space and time modeling, Yaguang Li et al. established DCRNN. The model architecture of DCRNN is shown in Fig.~\ref{fig:dcrnn_architecture}. The entire network trains the entire network by maximizing the likelihood of generating the target future time series using backpropagation. DCRNN can capture the spatio-temporal dependence between time series and can be applied to various spatio-temporal prediction problems.

\subsection{Differential RNNs}
Next, the various network architectures of Differential RNNs will be described in detail.

\subsubsection{Differential RNN}
For action recognition tasks \cite{38} \cite{39}, not all video frames contain significant patterns to distinguish different types of actions. Many spatio-temporal models, such as 3D-SIFT \cite{40} and HoGHoF \cite{41}, have been used to locate and encode significant spatiotemporal points. They detect and encode the spatio-temporal points in the video frame that are related to the significant action of the object, revealing important dynamic characteristics of the action.

Veeriah et al. proposed a new LSTM model \cite{42}, which automatically learns the dynamic characteristics of actions by detecting and integrating significant spatio-temporal sequences. Traditional LSTM may not be able to capture these remarkable dynamic patterns, because the gated unit does not explicitly consider the influence of the dynamic structure in the input sequence, which makes it difficult for the model to learn the evolution of behavioral states. To solve this problem, Veeriah et al. proposed a differential RNN (dRNN) model to learn and integrate the dynamic characteristics of actions.

The dRNN model proposed by Veeriah et al. is based on the following observation: the internal state of each memory unit contains accumulated information about the spatio-temporal structure, that is, it is a long short-term representation of the input sequence. Therefore, the derivative of states (DoS) $\frac{d{{s}_{t}}}{dt}$ quantifies the information change at each time step $t$. In other words, a large number of DoS is an indicator of a significant spatio-temporal structure, which contains information dynamics caused by sudden changes in the action state. In this case, the gated unit should allow more information to enter the memory unit to update its internal state, otherwise, when the DoS amplitude is very small, the input information should be gated out of the memory unit, so that the internal state will not be affected by the current input. Therefore, DoS should be used as a factor to control the flow of information in and out of the internal state of the memory unit.

\begin{figure}[H]
	\centering
	\includegraphics[width=0.8\textwidth]{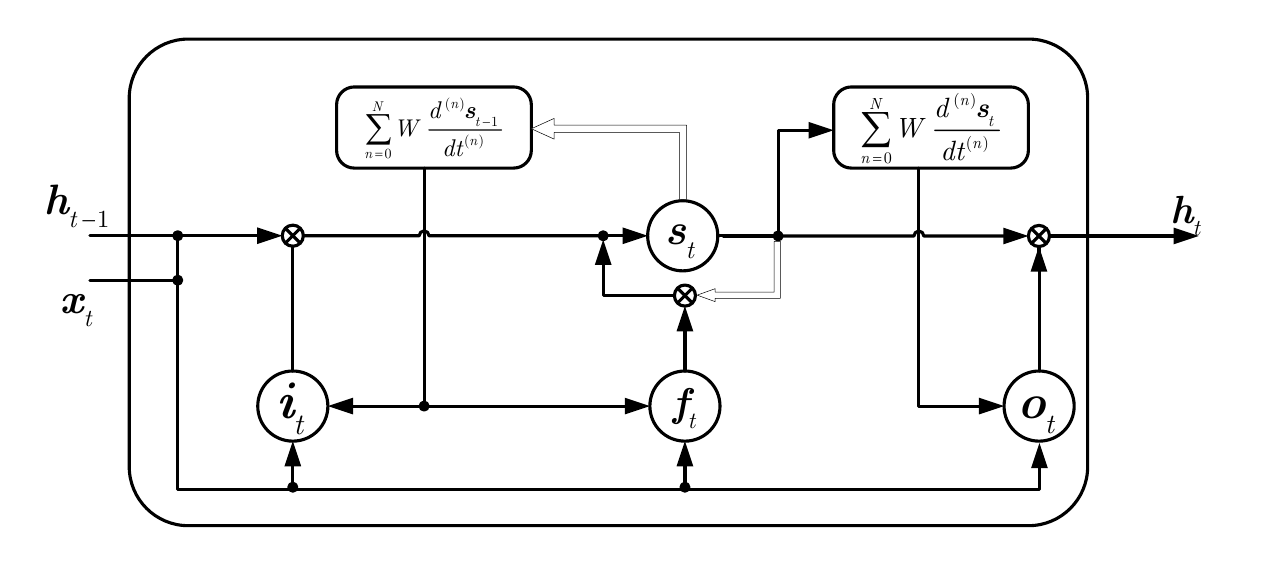}
	\caption{Architectures of dRNN at time $t$.}
	\label{fig:drnn_architecture}
\end{figure}

The dRNN has the following update equation:

\[{{i}_{t}}=\sigma (\sum\limits_{n=0}^{N}{W_{id}^{(n)}\frac{d{}^{(n)}{{s}_{t-1}}}{d{{t}^{(n)}}}}+{{W}_{ih}}{{h}_{t-1}}+{{W}_{ix}}{{x}_{t}}+{{b}_{i}})\]

\[{{f}_{t}}=\sigma (\sum\limits_{n=0}^{N}{W_{fd}^{(n)}\frac{d{}^{(n)}{{s}_{t-1}}}{d{{t}^{(n)}}}}+{{W}_{fh}}{{h}_{t-1}}+{{W}_{fx}}{{x}_{t}}+{{b}_{f}})\]

\[{{s}_{t}}={{f}_{t}}\otimes {{s}_{t-1}}+{{i}_{t}}\otimes tanh({{W}_{sh}}{{h}_{t-1}}+{{W}_{sx}}{{x}_{t}}+{{b}_{s}})\]

\[{{o}_{t}}=\sigma (\sum\limits_{n=0}^{N}{W_{od}^{(n)}\frac{d{}^{(n)}{{s}_{t}}}{d{{t}^{(n)}}}}+{{W}_{oh}}{{h}_{t-1}}+{{W}_{ox}}{{x}_{t}}+{{b}_{o}})\]

\begin{equation}
{{h}_{t}}={{o}_{t}}\otimes tanh({{s}_{t}})
\end{equation}

where $\frac{d{}^{(n)}{{s}_{t-1}}}{d{{t}^{(n)}}}$ is the n-order DoS, and $W_{*d}^{(n)}$ are the corresponding mapping matrices. $\otimes $ stands for element-wise product. In the memory cell, the input gate ${{i}_{t}}$ and the forget gate ${{f}_{t}}$ are controlled by DoS $\frac{d{}^{(n)}{{s}_{t-1}}}{d{{t}^{(n)}}}$ at time step $t-1$, and the output gate ${{o}_{t}}$ is controlled by the DoS $\frac{d{}^{(n)}{{s}_{t}}}{d{{t}^{(n)}}}$ at time step $t$.

The advantages of this model include: (1) High-order state derivatives can be used to learn complex time series. (2) It is more suitable to describe the short-term and long-term dynamics of behavior.

\subsubsection{Deep Differential RNN}
Naifan Zhuang et al. propose deep differential RNNs (${{d}^{2}}$) \cite{43}, inspired by Alex Graves \cite{44} to adjust the LSTM gate through an individual order of DoS. In the case of increasing DoS orders, multiple layers of LSTM units are stacked. Specifically, the first layer of ${{d}^{2}}$ uses zero-order DoS, which is similar to traditional LSTM units; the second layer uses LSTM units with first-order DoS; the third layer uses LSTM units with second-order DoS, and so on. Because it integrates the DoS idea from dRNN and the idea of deep stacked layers from deep RNN, the model is called deep differentiable RNN (${{d}^{2}}$). Each layer of ${{d}^{2}}$ learns the change of information gain through an individual order of DoS. As the unit layer of ${{d}^{2}}$ becomes deeper, this model can learn higher-order and more complex dynamic patterns.

\begin{figure}[H]
	\centering
	\includegraphics[width=0.8\textwidth]{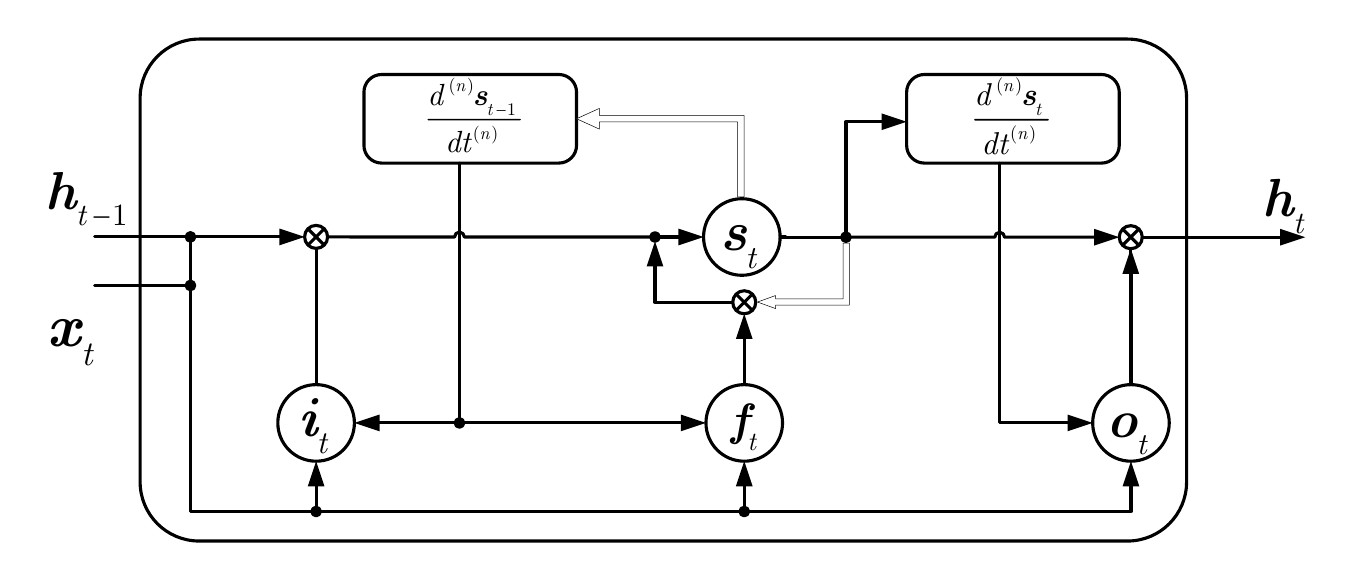}
	\caption{$n$-order (layer $n$) ${{d}^{2}}$ unit at time step $t$.}
	\label{fig:d2rnn_unit}
\end{figure}

Fig.~\ref{fig:d2rnn_unit} shows the LSTM unit in the proposed ${{d}^{2}}$ model layer $(n+1)$. The hollow line represents the information flow of ${{s}_{t-1}}$. The following recurrent equation can be used to control the LSTM gate in the ${{d}^{2}}$ layer $(n+1)$.

1) Input gate:

\begin{equation}
{{i}_{t}}=\sigma (W_{id}^{(n)}\frac{d{}^{(n)}{{s}_{t-1}}}{d{{t}^{(n)}}}+{{W}_{ih}}{{h}_{t-1}}+{{W}_{ix}}{{x}_{t}}+{{b}_{i}})
\end{equation}

2) Forget gate:

\begin{equation}
{{f}_{t}}=\sigma (W_{fd}^{(n)}\frac{d{}^{(n)}{{s}_{t-1}}}{d{{t}^{(n)}}}+{{W}_{ih}}{{h}_{t-1}}+{{W}_{fx}}{{x}_{t}}+{{b}_{f}})
\end{equation}

3) Output gate:

\begin{equation}
{{o}_{t}}=\sigma (W_{fd}^{(n)}\frac{d{}^{(n)}{{s}_{t}}}{d{{t}^{(n)}}}+{{W}_{ih}}{{h}_{t-1}}+{{W}_{ox}}{{x}_{t}}+{{b}_{o}})
\end{equation}

The advantages of this model include: (1) It can simulate more complex dynamics than general LSTM. (2) It can detect significant spatio-temporal structure. (3) Reduce the problem of information distortion.

\subsection{One-Layer RNN}
Sitian Qin et al. proposed a one-layer recurrent neural network to solve the pseudoconvex optimization problem \cite{45}.Consider the following nonsmooth pseudoconvex optimization problem \cite{46} \cite{47} \cite{48}:

\[minimize\quad f(x)\]

\[subject\ to\quad g(x)\le 0\]

\begin{equation}
Ax=b
\end{equation}

where $x={{({{x}_{1}},{{x}_{2}},\ldots ,{{x}_{n}})}^{T}}\in {{\mathbb{R}}^{n}}$ is the vector form of decision variable, $f:{{\mathbb{R}}^{n}}\to \mathbb{R}$ is set to be local Lipschitz continuous, but it may be non-smooth, $g(x)={{({{g}_{1}}(x),{{g}_{2}}(x),\ldots ,{{g}_{p}}(x))}^{T}}:$${{\mathbb{R}}^{n}}\to {{\mathbb{R}}^{p}}$is a $p$-dimensional vector-valued function and ${{g}_{i}}$ is convex function but may be non-smooth $(i=1,2,\ldots ,p)$, $A\in {{\mathbb{R}}^{m\times n}}$ is a full row-rank matrix (i.e.,rank $(A)=m\le n$), and $b={{({{b}_{1}},{{b}_{2}},\ldots ,{{b}_{m}})}^{T}}\in {{\mathbb{R}}^{m}}$. Without loss of generality, Sitian Qin et al. assume that the nonlinear pseudoconvex optimization problem (Eq.(39)) has at least an optimal solution. For convenience, suppose that

\begin{equation}
	\Omega =\left\{ x\in {{\mathbb{R}}^{n}}:Ax=b \right\}
\end{equation}

\begin{equation}
S=\left\{ x\in {{\mathbb{R}}^{n}}:g(x)\le 0 \right\}
\end{equation}

\begin{equation}
X=\Omega \bigcap S=\left\{ x\in {{\mathbb{R}}^{n}}:g(x)\le 0,Ax=b \right\}
\end{equation}

The following describes the model. First, assume that the following Assumption 1 holds.

Assumption 1:here exists $\hat{x}\in {{\mathbb{R}}^{n}}$, $r>0$, such that $\hat{x}\in int(S)\bigcap \Omega $, $S\subseteq B(\hat{x},r)=\left\{ x\in {{\mathbb{R}}^{n}}:\left\| x-\hat{x} \right\|<r \right\}$.

Based on assumption 1, the inequality constraint set $S$ defined as in Eq.(40)-(42)is bounded, that is, there is $R>0$ such that

\begin{equation}
S\subset B(0,R)=\left\{ x\in {{\mathbb{R}}^{n}}:\left\| x \right\|<R \right\}
\end{equation}

Let

\begin{equation}
{{G}_{i}}(x)={{g}_{i}}(x)+max\left\{ \left\| x \right\|,R \right\}-R,i=1,\ldots ,p
\end{equation}

\begin{equation}
G(x)={{({{G}_{1}}(x),\ldots ,{{G}_{p}}(x))}^{T}}
\end{equation}

Similar to the proofs in M. Forti et al. \cite{49}, Q. Liu et al. \cite{50} and W. Bian et al. \cite{51}, Sitian Qin et al. define a penalty function for the constraint set $S$ as follows:

\begin{equation}
D(x)=\sum\limits_{i=1}^{p}{max\left\{ 0,{{G}_{i}}(x) \right\}}
\end{equation}

Lemma 1:If assumption 1 holds, then:

1) The penalty function $D(x)$ is coercive, that is

\begin{equation}
\underset{\left\| x \right\|\to +\infty }{\mathop{lim}}\,D(x)=+\infty 
\end{equation}

2) $\left\{ x\in {{\mathbb{R}}^{n}}:g(x)\le 0 \right\}=\left\{ x\in {{\mathbb{R}}^{n}}:G(x)\le 0 \right\}$

Lemma 2:Based on the assumption 1, the penalty function $D(x)$ is convex. In addition:

\begin{equation}
\partial D(x)=\left\{ \begin{aligned}
	& \sum\limits_{i\in {{I}^{0}}(x)}{\left[ 0,1 \right]\partial {{G}_{i}}(x)\quad \quad \quad \ \ x\in bd(S)} \\ 
	& \left\{ 0 \right\}\quad \quad \quad \quad \quad \quad \ \ \ x\in int(S) \\ 
	& \sum\limits_{i\in {{I}^{+}}(x)}{\partial {{G}_{i}}(x)+\sum\limits_{i\in {{I}^{0}}(x)}{\left[ 0,1 \right]\partial {{G}_{i}}(x)\ \,x\notin S}} \\ 
\end{aligned} \right.
\end{equation}

The meaning of $x\in bd(S)$ is that $x$ is on the outer edge of convex set $S$, and the meaning of $x\in int(S)$ is that $x$ is inside convex set $S$. Where

\begin{equation}
	I^0(x) = \left\{ i \in \{1, 2, \ldots, m\} : G_i(x) = 0 \right\}
\end{equation}

\begin{equation}
{{I}^{+}}(x)=\left\{ i\in \left\{ 1,2,\ldots ,m \right\}:{{G}_{i}}(x)>0 \right\}
\end{equation}

In this section, assume that the objective function $f$ in Eq.(39) satisfies the following assumptions.

Assumption 2:

1) The objective function $f$ in Eq.(39) is pseudoconvex and regular on $X=\Omega \bigcap S$, while the local Lipschitz is continuous on $B(0,R)\bigcap \Omega $, where $R$ comes from Eq.(43).

2) $f$ is bounded below on the equality constraint set $\Omega $, i.e., $\underset{x\in \Omega }{\mathop{inf}}\,f(x)>-\infty $.

Remark 1: After simple row transformation of the matrix, the matrix $A$ in Eq.(39) can be translated into a full row-rank matrix. Therefore, assume that $A\in {{\mathbb{R}}^{m\times n}}$ is a full row-rank matrix (i.e.,rank$(A)=m\le n$). In this case, $A{{A}^{T}}$ is invertible.

Let

\begin{equation}
P={{A}^{T}}{{(A{{A}^{T}})}^{-1}}A
\end{equation}

Based on the above results, Sitian Qin et al. propose a one-layer neural network for solving nonsmooth pseudoconvex optimization problem, which has the following dynamic characteristics:

\begin{equation}
{{\epsilon }_{0}}\frac{d}{dt}x(t)\in -(I-P)\left[ \partial f(x(t))+\varepsilon (t)\partial D(x(t)) \right]-{{A}^{T}}h(Ax(t)-b)
\end{equation}

where ${{\epsilon }_{0}}$ is a positive scaling constant, $I$ is the identity matrix, $h(x)={{(\tilde{h}({{x}_{1}}),\tilde{h}({{x}_{2}}),\ldots ,\tilde{h}({{x}_{m}}))}^{T}}$ and its component is defined as

\begin{equation}
\tilde{h}({{x}_{i}})=\left\{ \begin{aligned}
	& 1\quad \quad \quad if\ {{x}_{i}}>0 \\ 
	& \left[ -1,1 \right]\quad \ \ \ if\ {{x}_{i}}=0 \\ 
	& -1\quad \quad \ \ \,if\ {{x}_{i}}<0 \\ 
\end{aligned} \right.
\end{equation}

and $\varepsilon :\left[ 0,+\left. \infty  \right)\to \left( 0,\left. +\infty  \right) \right. \right.$ satisfies the following conditions:

1) $\varepsilon (t)$ is a non-decreasing differentiable function;

2) $\underset{t\to +\infty }{\mathop{lim}}\,\varepsilon (t)=+\infty $;

3) ${{\varepsilon }^{-1}}(t)$ exists and differentiable;

4) $\int_{0}^{+\infty }{{{\varepsilon }^{-1}}(t)dt=+\infty }$

Remark 2:The function $\varepsilon (t)$ defined above is a Tikhonov regularization item, which serves as a penalty parameter.

The advantages of this model include: (1) The introduced penalty parameters do not need to determine specific parameters first, and have better convergence. (2) It is more convenient for design and implementation.

\subsection{High-Order RNN}
There can be two-way connections between the units of RecurrentNNs. This distinguishes them from feed-forward neural networks, in which the output of one unit is only connected to units in the next layer. In simple cases, the historical state of each unit or neuron is determined by a differential equation of the following form:

\begin{equation}
{{\dot{x}}_{i}}=-{{a}_{i}}{{x}_{i}}+{{b}_{i}}\sum\limits_{j}{{{w}_{ij}}{{y}_{j}}}
\end{equation}

where ${{x}_{i}}$ is the state of the $i$-th neuron, ${{a}_{i}}$, ${{b}_{i}}$ are constants,${{w}_{ij}}$ is the synaptic weight connecting the $j$-th input to the $i$-th neuron, and ${{y}_{j}}$ is the $j$-th input connected to the above neuron. Each ${{y}_{j}}$ is either an external input or a neuron state passed through a sigmoidal function, that is, ${{y}_{j}}=S({{x}_{j}})$, where $S(\cdot )$ is sigmoidal nonlinearity.

In a recurrent second-order neural network, the total input of a neuron is not only a linear combination of the input components ${{y}_{j}}$, but also a linear combination of their product ${{y}_{j}}{{y}_{k}}$. In addition, high-level interaction research including ternary ${{y}_{j}}{{y}_{k}}{{y}_{l}}$ and quaternary representations can also be carried out. This type of neural network forms a recurrent high-order neural network (RHONN) \cite{52} \cite{53}.

Consider a RHONN consisting of $n$ neurons and $m$ inputs. The state of each neuron is controlled by the following differential equation:

\begin{equation}
{{\dot{x}}_{i}}=-{{a}_{i}}{{x}_{i}}+{{b}_{i}}\left[ \sum\limits_{k=1}^{L}{{{w}_{ik}}\prod\limits_{j\in {{I}_{k}}}{y_{j}^{{{d}_{j}}(k)}}} \right]
\end{equation}

where $\left\{ {{I}_{1}},{{I}_{2}},\ldots ,{{I}_{L}} \right\}$ is the set of $L$ disordered subsets of $\left\{ 1,2,\ldots ,m+n \right\}$, ${{a}_{i}}$,${{b}_{i}}$ are real coefficients, ${{w}_{ik}}$ is the (adjustable) synaptic weight of the neural network, and ${{d}_{j}}(k)$ is a non-negative integer. The state of the $i$-th neuron is again represented as ${{x}_{i}}$, $y={{\left[ {{y}_{1}},{{y}_{2}},\ldots ,{{y}_{m+n}} \right]}^{T}}$ is the vector composed of the input of each neuron, defined as follows:

\begin{equation}
y=\left[ \begin{aligned}
	& \ {{y}_{1}} \\ 
	& \ \,\vdots  \\ 
	& \ {{y}_{n}} \\ 
	& {{y}_{n+1}} \\ 
	& \ \,\vdots  \\ 
	& {{y}_{m+n}} \\ 
\end{aligned} \right]=\left[ \begin{aligned}
	& S({{x}_{1}}) \\ 
	& \ \ \vdots  \\ 
	& S({{x}_{n}}) \\ 
	& \ \,{{u}_{1}} \\ 
	& \ \ \vdots  \\ 
	& \ \,{{u}_{m}} \\ 
\end{aligned} \right]
\end{equation}

where $u={{\left[ {{u}_{1}},{{u}_{2}},\ldots ,{{u}_{m}} \right]}^{T}}$ is the external input vector of the network. The function $S(\cdot )$ is a monotonically increasing sigmoidal function with the following form:

\begin{equation}
S(x)=\alpha \frac{1}{1+{{e}^{-\beta x}}}-\gamma 
\end{equation}

where $\alpha$ and $\beta$ are positive real numbers, and $\gamma$ is a real number. In special cases, let $\alpha =\beta =1$ and $\gamma =0$ to get the logistic function, and let $\alpha =\beta =2$ and $\gamma=1$ to get the hyperbolic tangent function; these are the most commonly used sigmoidal activation functions in neural network applications.

On this basis, the $L$-dimensional vector $z$ is introduced, which is defined as follows:

\begin{equation}
	z = 
	\begin{aligned}
		&\left[ \begin{array}{c}
			z_1 \\ 
			z_2 \\ 
			\vdots \\ 
			z_L \\ 
		\end{array} \right]
		= 
		\left[ \begin{array}{c}
			\prod\limits_{j \in I_1} y_j^{d_j(1)} \\ 
			\prod\limits_{j \in I_2} y_j^{d_j(2)} \\ 
			\vdots \\ 
			\prod\limits_{j \in I_L} y_j^{d_j(L)} \\ 
		\end{array} \right]
	\end{aligned}
	\label{eq:high_order_vector}
\end{equation}

Therefore, the RHONN model is simplified to:

\begin{equation}
{{\dot{x}}_{i}}=-{{a}_{i}}{{x}_{i}}+{{b}_{i}}\sum\limits_{k=1}^{L}{{{w}_{ik}}{{z}_{k}}}
\end{equation}

In addition, if the adjustable parameter vector is defined as ${{\theta }_{i}}:={{b}_{i}}{{\left[ {{w}_{i1}},{{w}_{i2}},\ldots ,{{w}_{iL}} \right]}^{T}}$, the above formula becomes:

\begin{equation}
{{\dot{x}}_{i}}=-{{a}_{i}}{{x}_{i}}+\theta _{i}^{T}z
\end{equation}

The vector $\left\{ {{\theta }_{i}}:i=1,\ldots ,n \right\}$ represents the adjustable weight of the network, and the coefficient $\left\{ {{a}_{i}}:i=1,\ldots ,n \right\}$ is a part of the underlying network architecture, which is fixed during training. In order to ensure that the bounded-input bounded-output (BIBO) of each neuron ${{x}_{i}}$ is stable, it is assumed that each ${{a}_{i}}$ is positive. Kosmatopoulos et al. found that in the special case of the continuous-time Hopfield model \cite{54}, ${{a}_{i}}={1}/{{{R}_{i}}{{C}_{i}}}\;$, where ${{R}_{i}}>0$ and ${{C}_{i}}>0$ are the resistance and capacitance of the $i$-th of the network, respectively.

The dynamic behavior of the entire network is expressed as a vector:

\begin{equation}
\dot{x}=Ax+{{\Theta }^{T}}z
\end{equation}

where $x={{\left[ {{x}_{1}}\ldots {{x}_{n}} \right]}^{T}}$, $\Theta =\left[ {{\theta }_{1}}\ldots {{\theta }_{n}} \right]\in {{\mathbb{R}}^{L\times n}}$, $A:=diag\left\{ -{{a}_{1}},-{{a}_{2}},\ldots ,-{{a}_{n}} \right\}$ is a diagonal matrix of $n\times n$. Because each ${{a}_{i}}$ is positive, $A$ is a stable matrix. Although not written in the formula, the vector $z$ is a function of the network state $x$ and the external input $u$.

The advantages of this model include: (1) If enough high-order connections are allowed then this network is capable of approximating arbitrary dynamical systems. (2) It is a learning algorithm that can ensure the stability of the entire system and is effective.

\subsection{Highway Networks}
Next, the various network architectures of Highway Networks will be described in detail.

\subsubsection{Highway Networks}
Rupesh Kumar Srivastava et al. \cite{55} took inspiration from the LSTM architecture and further expanded the Mixture of Units (MoU) architecture by adding additional units (implemented using nonlinear transformation $C$) \cite{56} \cite{57}. In addition to these units within the layer, they can also learn to control the flow of information across layers:

\begin{equation}
{{y}_{l}}=H({{x}_{l}})\cdot T({{x}_{l}})+{{x}_{l}}\cdot C({{x}_{l}})
\end{equation}

Since there are now two types of gating units, the gating conversion is named as $T$ and $C$. $T$ realizes the transform gate, which learns the contribution of the regular conversion $H$ of the input $x$ to the total output $y$ of the layer. $C$ implements a carry gate, which learns how many inputs are brought to the output. These gates are similar to the input and forget gates in the LSTM architecture, and a constant-weight skip connection (weight 1) from input to output is similar to the recurrent self-connection used by the LSTM unit. Networks with this architecture are called highway networks because they allow information to flow unimpeded across layers.

The conversion realized by the highway layer is much more flexible than the multilayer perceptron. Pay special attention to the specific output of $T$ and $C$:

\begin{equation}
{{y}_{l}}=\left\{ \begin{aligned}
	& x{}_{l},\quad \quad if\ T(x{}_{l})=0,C(x{}_{l})=1 \\ 
	& H(x{}_{l}),\quad \ if\ T(x{}_{l})=1,C(x{}_{l})=0\  \\ 
\end{aligned} \right.
\end{equation}

Similarly, for the derivative of the layer outputs with respect to its inputs,

\begin{equation}
\frac{d{{y}_{l}}}{dx{}_{l}}=\left\{ \begin{aligned}
	& I,\quad \quad \ if\ T(x{}_{l})=0,C(x{}_{l})=1 \\ 
	& {H}'{{(x)}_{l}},\quad \ if\ T(x{}_{l})=1,C(x{}_{l})=0\  \\ 
\end{aligned} \right.
\end{equation}

Therefore, according to the output of the transition and carry gate, the highway layer can change its behavior between the identity layer (a layer that simply passes its input) and their unit combination. This design allows for credit assignment at a greater depth, because during back propagation, the gradient can flow in the reverse direction without reducing the path through which the carry gate output is non-zero.

The advantages of this model include: (1) It can solve the difficult problem of deep network training. (2) Effectively alleviate the gradient problem.

\subsubsection{Recurrent Highway Networks}
Georg Zilly et al. proposed the Recurrent Highway Network (RHN) layer \cite{58}, in which one or more highway layers are in a recurrent state transition (equal to the desired recurrence depth). Formally, let ${{W}_{H,T,C}}\in {{\mathbb{R}}^{n\times m}}$ and ${{R}_{{{H}_{l}},{{T}_{l}},{{C}_{l}}}}\in {{\mathbb{R}}^{n\times n}}$ denote the weight matrix of the nonlinear transformation of $H$ and the weight matrix of the gates $T$ and $C$ at layer $l\in \left\{ 1,\ldots ,L \right\}$. The bias is denoted as ${{b}_{{{H}_{l}},{{T}_{l}},{{C}_{l}}}}\in {{\mathbb{R}}^{n}}$, ${{s}_{l}}$ denotes the intermediate output of layer $l$, and $s_{0}^{\left[ t \right]}={{y}^{\left[ t-1 \right]}}$. The following describes a RHN layer with a recurrence depth of $L$:

\begin{equation}
s_{l}^{\left[ t \right]}=h_{l}^{\left[ t \right]}\cdot t_{l}^{\left[ t \right]}+s_{l-1}^{\left[ t \right]}\cdot c_{l}^{\left[ t \right]}
\end{equation}

\begin{equation}
h_{l}^{\left[ t \right]}=tanh({{W}_{H}}{{x}^{\left[ t \right]}}{{\mathbb{I}}_{\left\{ l=1 \right\}}}+{{R}_{{{H}_{l}}}}s_{l-1}^{\left[ t \right]}+{{b}_{{{H}_{l}}}})
\end{equation}

\begin{equation}
t_{l}^{\left[ t \right]}=\sigma ({{W}_{T}}{{x}^{\left[ t \right]}}{{\mathbb{I}}_{\left\{ l=1 \right\}}}+{{R}_{{{T}_{l}}}}s_{l-1}^{\left[ t \right]}+{{b}_{{{T}_{l}}}})
\end{equation}

\begin{equation}
c_{l}^{\left[ t \right]}=\sigma ({{W}_{C}}{{x}^{\left[ t \right]}}{{\mathbb{I}}_{\left\{ l=1 \right\}}}+{{R}_{{{C}_{l}}}}s_{l-1}^{\left[ t \right]}+{{b}_{{{C}_{l}}}})
\end{equation}

and ${{\mathbb{I}}_{\left\{ {} \right\}}}$ is the indicator function.

The schematic diagram of RHN calculation is shown in Fig.(19). The output of the RHN layer is the output of the $L$-th layer of the highway layer, that is, ${{y}^{\left[ t \right]}}=s_{L}^{\left[ t \right]}$.

Note that in the recurrent transition, ${{x}^{\left[ t \right]}}$ is only directly converted by the first highway layer $(l=1)$. For this layer, $s_{l-1}^{\left[ t \right]}$ is the output of the RHN layer at the previous time step. The subsequent highway layer only processes the output of the previous layer. The vertical dashed line in Fig.(19) separates the multiple highway layers in the recurrent transition.

\begin{figure}[H]
	\centering
	\includegraphics[width=0.8\textwidth]{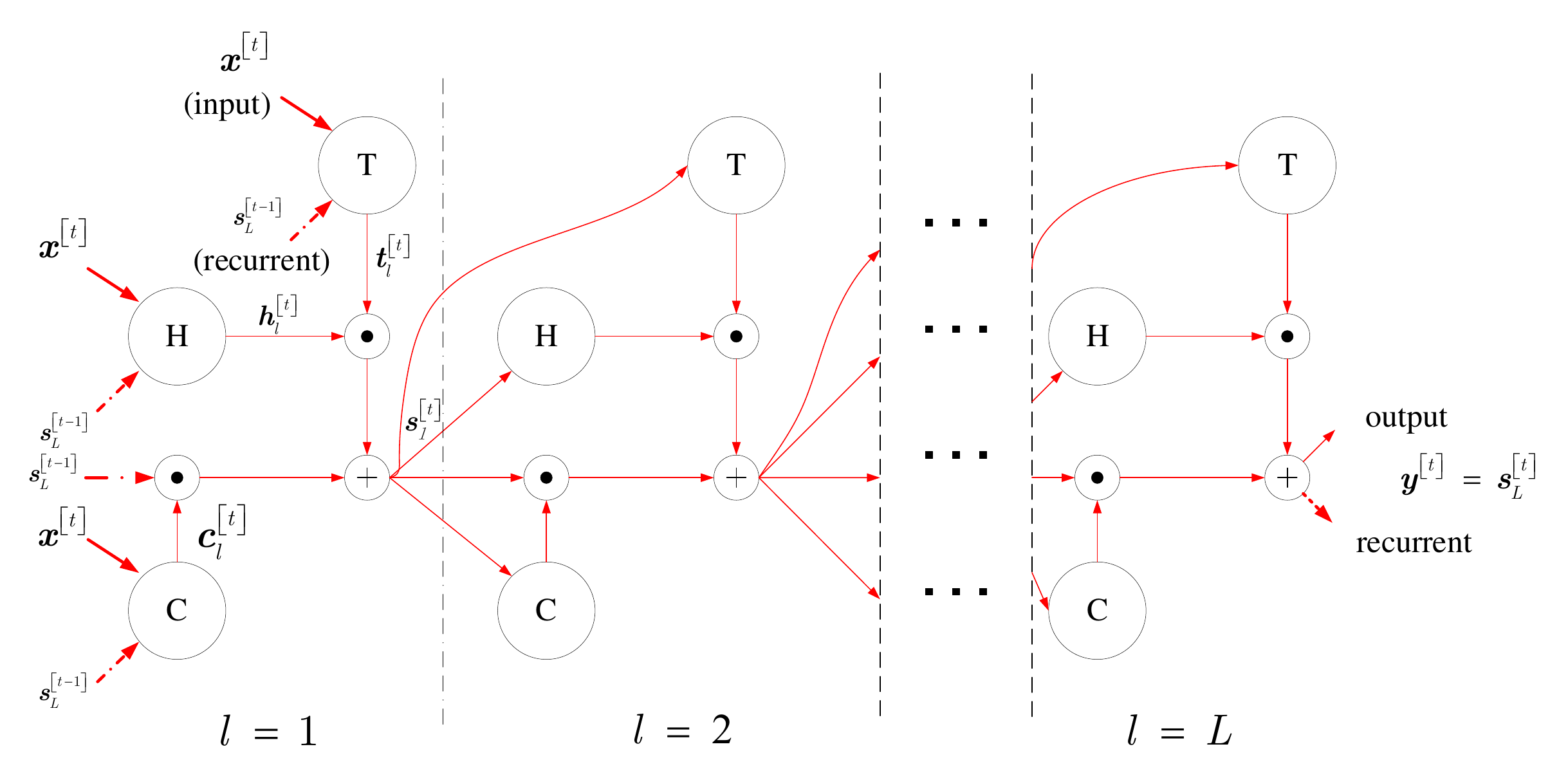}
	\caption{The figure shows the calculation of the RHN layer in the recurrent loop. Vertical dashed lines delimit stacked highway layers. The horizontal dashed line means to extend the recurrent depth by stacking more layers.}
	\label{fig:rhn_structure}
\end{figure}

Conceptually, the RHN layer with $L=1$ is basically a basic variant of the LSTM layer. Like other variants \cite{59} \cite{60} \cite{61}, it retains the basic essential components of LSTM—the multiplier gating unit, which can control information flow through self-connection. The RHN layer will naturally expand to $L>1$, thereby expanding the LSTM to simulate more complex state transitions.

The simpler RHN layer formula can be analyzed similar to the standard RNN based on the Geršgorin circle theorem (GCT). Omit the input and biases, the temporal Jacobian $A={\partial {{y}^{\left[ t \right]}}}/{\partial {{y}^{\left[ t-1 \right]}}}\;$ of the RHN layer recurrence depth of 1 (${{y}^{[t]}}={{h}^{[t]}}\cdot {{t}^{[t]}}+{{y}^{[t-1]}}\cdot {{c}^{[t]}}$) is given by the following formula:

\begin{equation}
A=diag({{c}^{\left[ t \right]}})+{H}'diag({{t}^{\left[ t \right]}})+{C}'diag({{y}^{\left[ t-1 \right]}})+{T}'diag({{h}^{\left[ t \right]}})
\end{equation}

\begin{equation}
{H}'=R_{H}^{\top }diag\left[ tan{h}'({{R}_{H}}{{y}^{\left[ t-1 \right]}}) \right]
\end{equation}

\begin{equation}
{T}'=R_{T}^{\top }diag\left[ {\sigma }'({{R}_{T}}{{y}^{\left[ t-1 \right]}}) \right]
\end{equation}

\begin{equation}
{C}'=R_{C}^{\top }diag\left[ {\sigma }'({{R}_{C}}{{y}^{\left[ t-1 \right]}}) \right]
\end{equation}

and has a spectrum of:

\begin{equation}
spec(A)\subset \bigcup\limits_{i\in \left\{ 1,\ldots ,n \right\}}{\left\{ \lambda \in \mathbb{C} \right.}\left| \left\| \lambda -c_{i}^{\left[ t \right]}-{{{{H}'}}_{ii}}t_{i}^{\left[ t \right]} \right. \right.-{{{C}'}_{ii}}y_{i}^{\left[ t-1 \right]}-{{{T}'}_{ii}}{{\left. h_{i}^{\left[ t \right]} \right\|}_{\mathbb{C}}}\le \sum\limits_{j=1,j\ne i}^{n}{\left| {{{{H}'}}_{ij}}t_{i}^{\left[ t \right]}+{{{{C}'}}_{ij}}y_{i}^{\left[ t-1 \right]}+{{{{T}'}}_{ij}}h_{i}^{\left[ t \right]} \right|}\left. {} \right\}
\end{equation}

Eq.(73) shows the influence of gating on the eigenvalues $A$. Compared with RecurrentNNs, the RHN layer has more flexibility in adjusting the center and radius of Geršgorin circle, and it is especially able to notice two restrictions. If all the carry gates are fully opened and the transform gates are fully closed, then $c={{1}_{n}}$, $t={{0}_{n}}$, ${T}'={C}'={{0}_{n\times n}}$ (because $\sigma$ is saturated).

This leads to $c={{1}_{n}},\quad t={{0}_{n}}\Rightarrow {{\lambda }_{i}}=1\ \forall i\in \left\{ 1,\ldots ,n \right\}$.

Since the radius of the Geršgorin circle is reduced to 0, and each diagonal element is set to ${{c}_{i}}=1$, all eigenvalues are set to 1. In another extreme case, if $c={{0}_{n}}$ and $t={{1}_{n}}$, then the eigenvalue is the eigenvalue of ${H}'$. When the gate changes between 0 and 1, each characteristic value of $A$ can be dynamically adjusted to any combination of the above restriction behaviors.

The advantages of this model are: (1) It is a powerful new model designed to take advantage of increased depth in the recurrent transition while retaining the ease of training of LSTMs. (2) Solve sequential processing tasks.

\subsubsection{Highway LSTM RNNs}

\begin{figure}[H]
	\centering
	\includegraphics[width=1.0\textwidth]{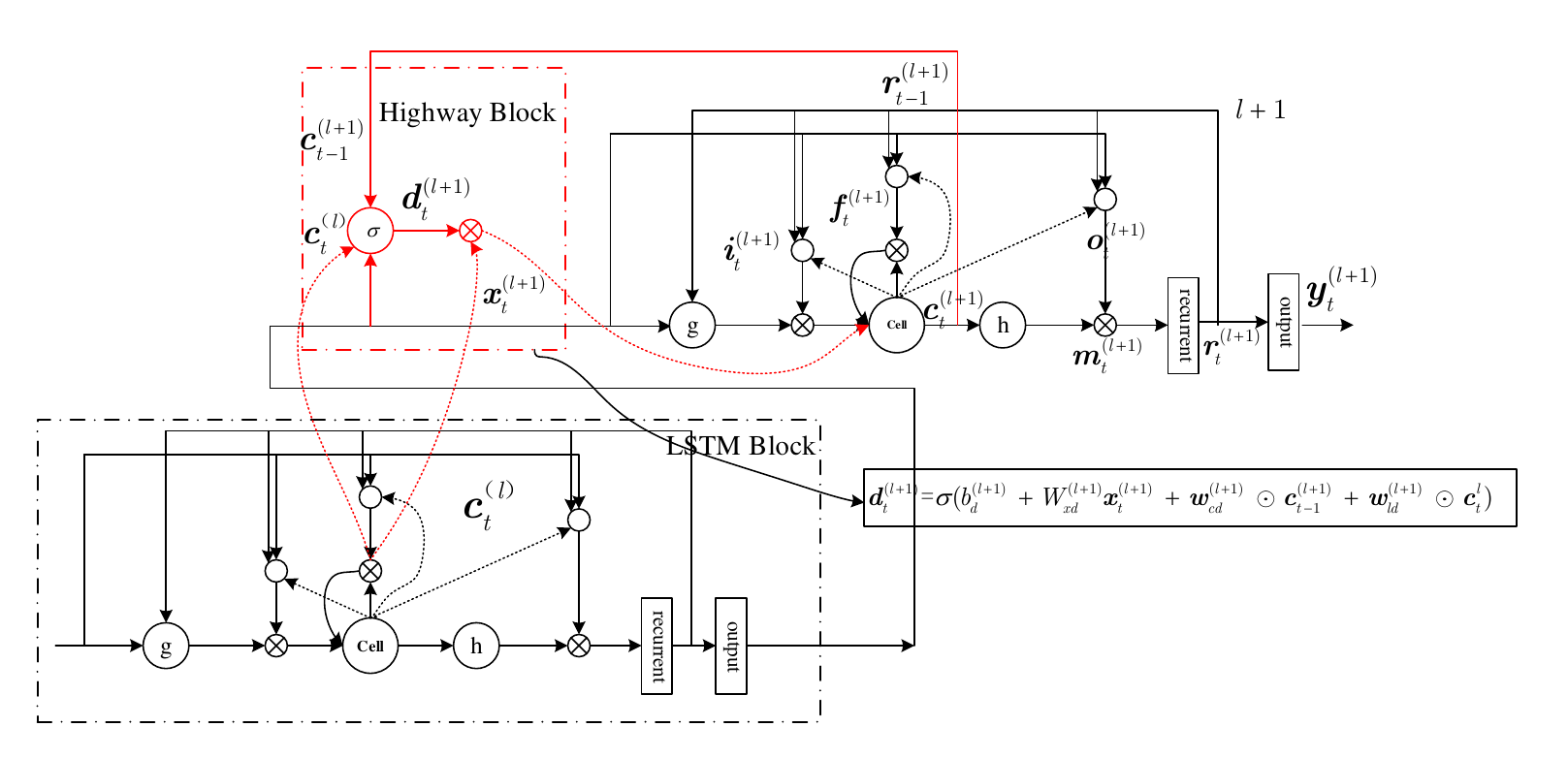}
	\caption{Highway LSTM RNNs}
	\label{fig:highway_lstm}
\end{figure}

The highway LSTM (HLSTM) RNNs \cite{62} proposed by Yu Zhang et al. are shown in Fig.(20). It has a direct gated connection between the memory cell $c_{t}^{l}$ of the lower layer $l$ and the memory cell $c_{t}^{l+\text{1}}$ of the upper layer $l+1$ (in the red block in Fig.(20)). The carry gate controls the amount of information that flows directly from the lower unit to the upper unit. The gate function of layer $l+1$ at time step $t$ is:

\begin{equation}
d_{t}^{(l+1)}\text{=}\sigma (b_{d}^{(l+1)}+W_{xd}^{(l+1)}x_{t}^{(l+1)}+w_{cd}^{(l+1)}\odot c_{t-1}^{(l+1)}+w_{ld}^{(l+1)}\odot c_{t}^{l})
\end{equation}

where $b_{d}^{(l+1)}$ is the bias and $W_{xd}^{(l+1)}$ is the weight matrix connecting the carry gate to the input of this layer. $w_{cd}^{(l+1)}$ is the weight vector from the carry gate to the past unit state in the current layer. $w_{ld}^{(l+1)}$ is the weight vector connecting the carry gate to the lower memory unit. ${{d}^{(l+1)}}$ is the activation vector of the carry gate on layer $l+1$. Using the carry gate, HLSTM \cite{63} \cite{64} RNN calculates the state of the unit on layer $l+1$ according to the following formula:

\begin{equation}
c_{t}^{(l+1)}=d_{t}^{(l+1)}\odot c_{t}^{l}+f_{t}^{(l+1)}\odot c_{t-1}^{(l+1)}+i_{t}^{(l+1)}\odot tanh(W_{xc}^{(l+1)}x_{t}^{(l+1)}+W_{hc}^{(l+1)}m_{t-1}^{(l+1)}+{{b}_{c}})
\end{equation}

while other equations are the same as those in general LSTM.

Therefore, due to the output of the carry gate, the highway connection can smoothly change its behavior between LSTM layers, or simply pass its unit memory from the upper layer. When training a deeper LSTM, the highway connections between units of different layers make the influence between units of one layer to another layer more direct and can alleviate the problem of gradient disappearance.

\subsubsection{Bidirectional Highway LSTM RNNs}
The unidirectional LSTM RNNs described in the previous section can only use past history. However, in speech recognition, the future environment will also carry information, and it should be used to further enhance the acoustic model. Bidirectional RNNs process data by using two independent hidden layers in both directions, thereby taking advantage of past and future contexts. The results of A. Graves \cite{44} et al., N. Jaitly et al. \cite{65}, and K. Chen et al. \cite{66} showed that bidirectional LSTM RNNs can indeed improve the effect of speech recognition. Therefore, HLSTM RNNs are expanded from unidirection to bidirection. Note that the reverse layer follows the same equation used in the forward layer, except that $t-1$ is replaced by $t+1$ to take advantage of future frames, and the model runs from $t=T$ to 1. The output of the reverse layer and the forward layer are connected to form the input of the next layer.

\subsubsection{Highway Convolutional Recurrent Deep Neural Network}
The Highway Convolutional Recurrent Deep Neural Network (Highway CLDNN) structure proposed by Wei-Ning Hsu et al. \cite{67} follows the design of T. N. Sainath et al. \cite{68}. Fig.(21) illustrates the architecture of the model. Since the recurrent layer can capture the temporal relationship, at each time step, the frame input ${{x}_{t}}$ without context is passed to the network. Wei-Ning Hsu et al. used filter bank features and pitch features to represent each frame ${{x}_{t}}$.

\begin{figure}[H]
	\centering
	\includegraphics[width=0.4\textwidth]{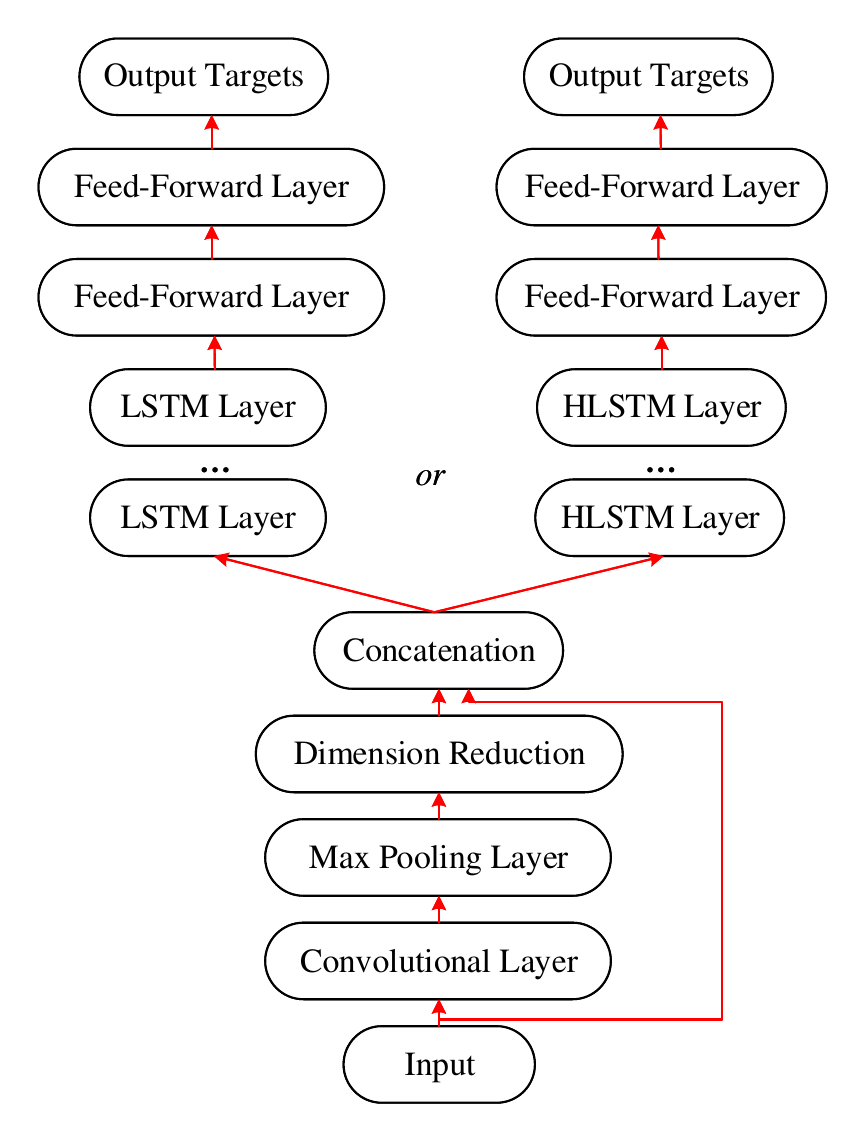}
	\caption{The left is CLDNN, and the right is Highway CLDNN.}
	\label{fig:highway_cldnn}
\end{figure}

In the field of speech recognition, in order to solve the speaker normalization problem, the input features are first passed to the convolutional layer. This layer generates 256 feature maps, and its core is composed of 1×8 receptive field. Immediately after this convolutional layer, a pooling window and a non-overlapping maximum pooling with a step size of 3 are applied. Since the output dimension is still large (i.e., $\lfloor 83/3 \rfloor \times 256$), a 256-dimensional projection layer is added on top of the maximum pooling layer. Subsequently, in order to model the temporal relationship, the output of the projection layer is fed to the recurrent layer. In addition, according to the suggestion of T. N. Sainath, et al. \cite{68}, Wei-Ning Hsu et al. also pass the original input features ${{x}_{t}}$ to the recurrent layer to provide input representations from different levels. Since the speech signal has information of different time scales \cite{69}, it is hoped that the recurrent layer can also capture the temporal relationship on different scales, and Deep RNN can capture this relationship \cite{70}. Therefore, in order to utilize the deep recurrent structure, Highway LSTM is used for the recurrent layer. Specifically, 1024 units are applied on a 5-layer LSTM, and 512 dimensions are projected on each layer. Finally, the output of the last recurrent layer is fed to the fully connected feedforward layer, which can better distinguish the output targets. In this way, two layers of 1024 hidden unit searches are applied, and each layer has a corrected linear activation function.

\subsection{Multidimensional RNNs}
Next, the various network architectures of Multidimensional RNNs will be described in detail.

\subsubsection{Two-dimensional LSTM}

\begin{figure}[H]
	\centering
	\includegraphics[width=0.7\textwidth]{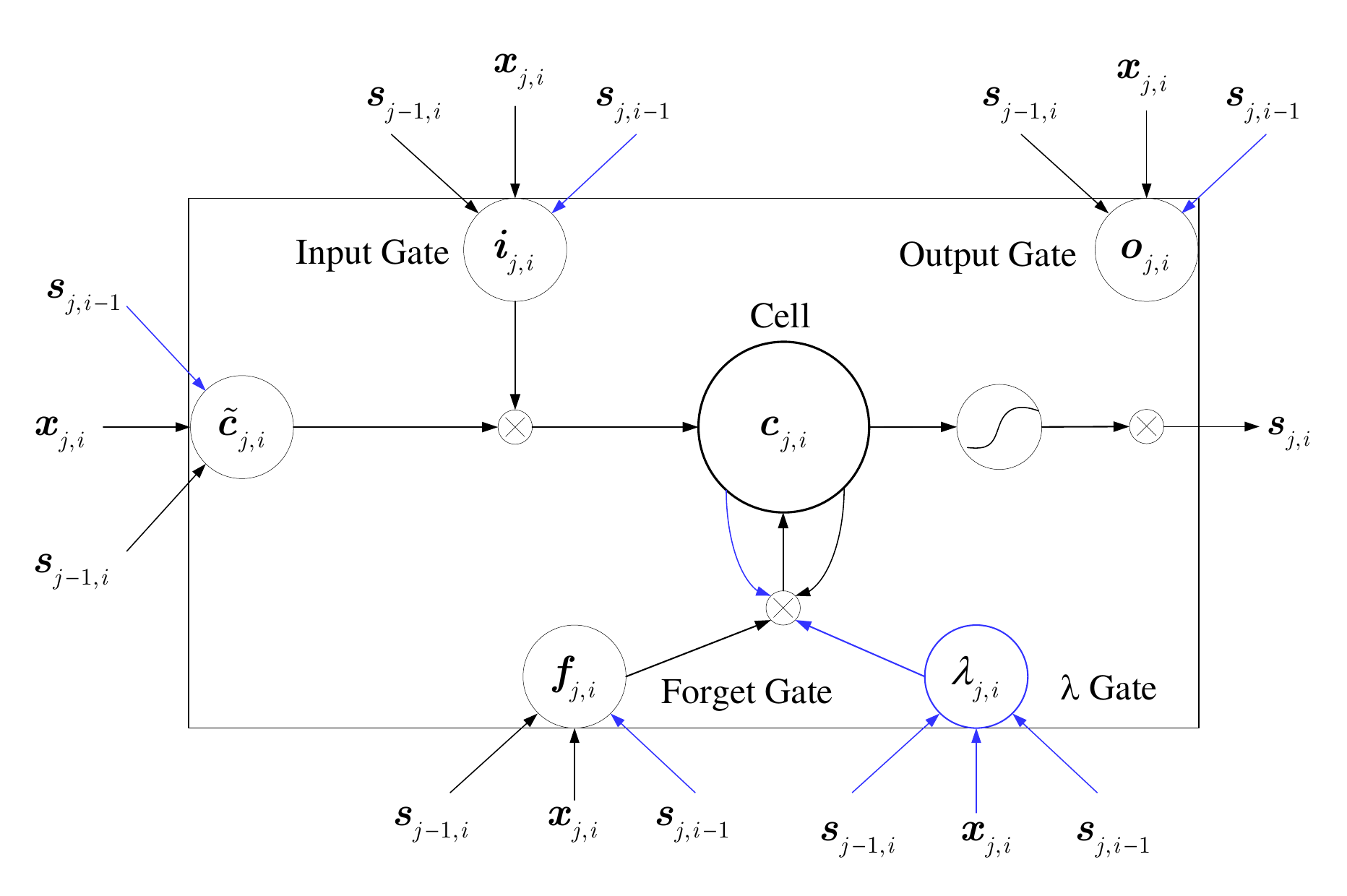}
	\caption{Two-dimensional LSTM unit. Additional connections and general LSTM are marked in blue.}
	\label{fig:2dlstm_unit}
\end{figure}

Two-dimensional LSTM (2DLSTM) \cite{71} was introduced by Graves \cite{3} as a generalization of general LSTM. Fig.(22) illustrates a stable variant proposed by Leifert \cite{72}. The 2DLSTM unit processes two-dimensional sequence data $x\in {{\mathbb{R}}^{J\times I}}$ of arbitrary length ($J$ and $I$). At time step $(j,i)$, the calculation of the unit depends on the vertical hidden state ${{s}_{j,i-1}}$ and the horizontal hidden state ${{s}_{i,j-1}}$ (see Eq.(76)--(80)). Similar to the LSTM unit, the 2DLSTM retains some state information in an internal unit state ${{c}_{j,i}}$. In addition to the input gate ${{i}_{j,i}}$, the forget gate ${{f}_{j,i}}$, and the output gate ${{o}_{j,i}}$ to control the information flows, the 2DLSTM also uses an additional $\lambda $ gate ${{\lambda }_{j,i}}$. As written in Eq.(80), the calculation of $\lambda $ gate activation is similar to other gates. Before passing the forgetting gate, use $\lambda $ gate to weight the two predecessor unit ${{c}_{j-1,i}}$ and ${{c}_{j,i-1}}$ (Eq.(81)). $g$ and $\sigma $ are the tanh and sigmoid functions. ${{V}_{S}}$, ${{W}_{S}}$and${{U}_{S}}$ are weight matrices.

To train a 2DLSTM unit, back propagation through time (BPTT) is performed in two dimensions. Therefore, the gradient passes backward from time step $(J,I)$ to $(1,1)$, which is the origin.

\begin{equation}
{{i}_{j,i}}=\sigma ({{W}_{1}}{{x}_{j,i}}+{{U}_{1}}{{s}_{j-1,i}}+{{V}_{1}}{{s}_{j,i-1}})
	\label{eq:2dlstm_gates}
\end{equation}

\begin{equation}
{{f}_{j,i}}=\sigma ({{W}_{2}}{{x}_{j,i}}+{{U}_{2}}{{s}_{j-1,i}}+{{V}_{2}}{{s}_{j,i-1}})
\end{equation}

\begin{equation}
{{o}_{j,i}}=\sigma ({{W}_{3}}{{x}_{j,i}}+{{U}_{3}}{{s}_{j-1,i}}+{{V}_{3}}{{s}_{j,i-1}})
\end{equation}

\begin{equation}
{{\tilde{c}}_{j,i}}=g({{W}_{4}}{{x}_{j,i}}+{{U}_{4}}{{s}_{j-1,i}}+{{V}_{4}}{{s}_{j,i-1}})
\end{equation}

\begin{equation}
{{\lambda }_{j,i}}=\sigma ({{W}_{5}}{{x}_{j,i}}+{{U}_{5}}{{s}_{j-1,i}}+{{V}_{5}}{{s}_{j,i-1}})
\end{equation}

\begin{equation}
{{c}_{j,i}}={{f}_{j,i}}\circ \left[ {{\lambda }_{j,i}}\circ {{c}_{j-1,i}}+(1-{{\lambda }_{j,i}})\circ {{c}_{j,i-1}} \right]+{{\tilde{c}}_{j,i}}\circ {{i}_{j,i}}
\end{equation}

\begin{equation}
	{{s}_{j,i}}=g({{c}_{j,i}})\circ {{o}_{j,i}}
\end{equation}

\subsubsection{Multidimensional RNNs}
The basic idea of multidimensional RNN (MDRNN) \cite{73} \cite{74} is to replace the single recurrent connection in RecurrentNNs with the spatio-temporal dimensions in the data. These connections allow the network to create a flexible internal representation of the surrounding context, which is robust to local distortion.

\begin{figure}[H]
	\centering
	\includegraphics[width=0.6\textwidth]{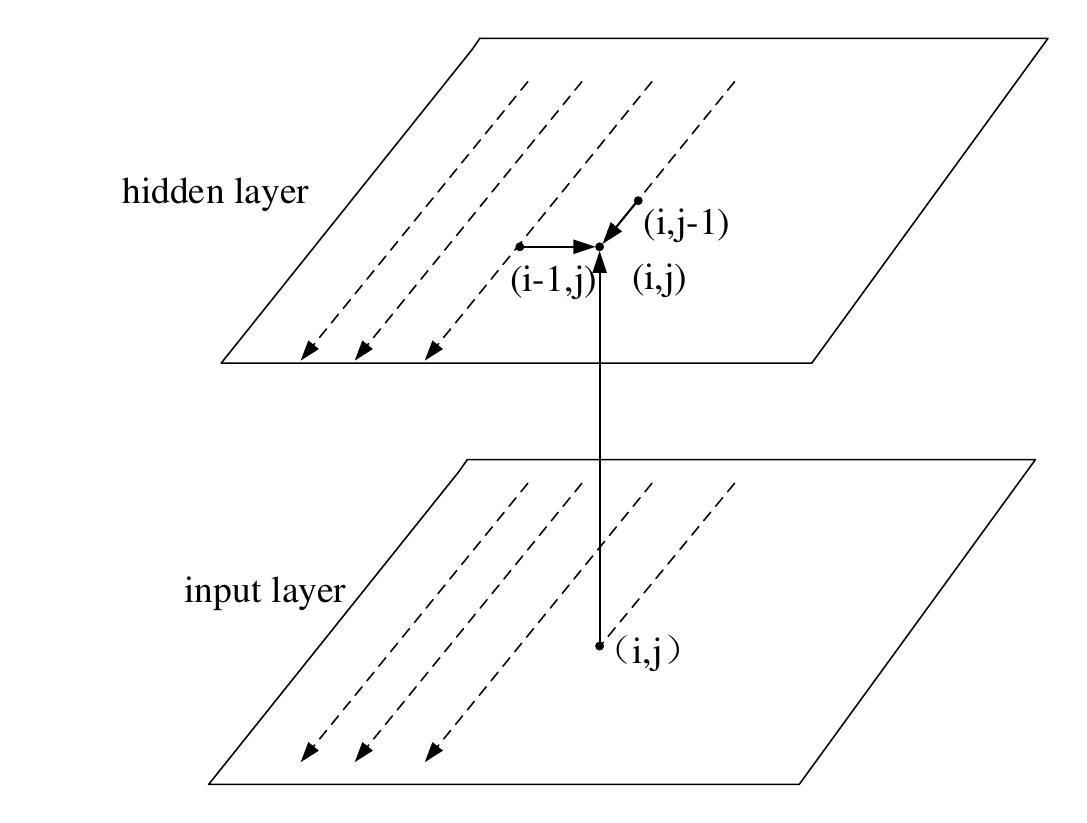}
	\caption{ Two-dimensional RNN. The thick lines indicate the connections with the current point $(i,j)$. The connections in the hidden layer plane are recurrent. The dashed lines indicate the scanning strips along which previous points were visited, starting at the top left corner.}
	\label{fig:md_rnn_structure}
\end{figure}

The hidden layer of MDRNN scans the input in the one-dimensional strips and stores the activation in the buffer. The strips are sorted in such a way that at each point, the layer has visited these points one step back along each dimension. The hidden activations of these previous points are fed to the current point through recurrent connections along with the input. The two-dimensional structure is shown in Fig.~\ref{fig:md_rnn_structure}.

One such layer is sufficient to enable the network to access all contexts from the direction of the current point scan (e.g. to the top and left of $(i,j)$ in Fig.~\ref{fig:md_rnn_structure}). However, it is usually desirable to have surrounding contexts in all directions. The problem also exists in one-dimensional networks, so it is very useful to obtain information about the future and the past. The typical one-dimensional solution is a bidirectional recurrent network, with two independent hidden layers scanning the input forward and backward. The generalization of bidirectional network to $n$-dimensional requires ${{2}^{n}}$ hidden layers, starting from each corner in the $n$-dimensional hypercube and scanning in the opposite directions. For example, a two-dimensional network has four layers, and one layer starts from the top left corner and scans down and right, one layer starts from the bottom left and scans up and right, etc. All hidden layers are connected to a single output layer, so they can receive all the surrounding information context.

Uses the $n$-dimensional expansion of backpropagation over time to calculate the error gradient of the MDRNN. As in the one-dimensional case, the order of data processing is opposite to that of the forward process. Each hidden layer receives the output derivative and its own $n$ 'future' derivatives at each time step.

Let $a_{j}^{p}$ and $b_{j}^{p}$ be respectively the input and activation of unit $j$ at point $p=({{p}_{1}},\ldots ,{{p}_{n}})$ in $n$-dimensional input sequence $x$ with dimensions $({{D}_{1}},\ldots ,{{D}_{n}})$. Let $p_{d}^{-}=({{p}_{1}},\ldots ,{{p}_{d}}-1,\ldots ,{{p}_{n}})$,$p_{d}^{+}=({{p}_{1}},\ldots ,{{p}_{d}}+1,\ldots ,{{p}_{n}})$. Let ${{w}_{ij}}$ and $w_{ij}^{d}$ be the feedforward connection weight from unit $i$ to unit $j$ and the recurrent connection weight from $i$ to $j$ along dimension $d$, respectively. Let ${{\theta }_{h}}$ be the activation function of the hidden unit $h$, for some unit $j$ and some differentiable objective function $O$, let $\delta _{j}^{p}=\frac{\partial O}{\partial a_{j}^{p}}$. Then, the forward and backward equations of an $n$-dimensional MDRNN with $I$ input unit, $K$ output units and $H$ hidden summation units are as follows:

Forward Pass:

\begin{equation}
a_{h}^{p}=\sum\limits_{i=1}^{I}{x_{i}^{p}}{{w}_{ih}}+\sum\limits_{\begin{smallmatrix} 
		d=1: \\ 
		{{p}_{d}}>0 
\end{smallmatrix}}^{n}{\sum\limits_{\hat{h}=1}^{H}{b_{{\hat{h}}}^{p_{d}^{-}}}}w_{\hat{h}h}^{d}
\end{equation}

\begin{equation}
b_{h}^{p}={{\theta }_{h}}(a_{h}^{p})
\end{equation}

Backward Pass:

\begin{equation}
\delta _{h}^{p}={{{\theta }'}_{h}}(a_{h}^{p})(\sum\limits_{k=1}^{K}{\delta _{k}^{p}{{w}_{hk}}+\sum\limits_{\begin{smallmatrix} 
			\ \ d=1: \\ 
			{{p}_{d}}<{{D}_{d}}-1 
	\end{smallmatrix}}^{n}{\sum\limits_{\hat{h}=1}^{H}{\delta _{{\hat{h}}}^{p_{d}^{+}}}}w_{h\hat{h}}^{d}})
\end{equation}

The advantages of this model include: (1) Robust to any combination of input dimensions. (2) The multidimensional context can be modeled in a flexible way.

\subsection{Bidirectional RNNs}
Next, the various network architectures of Bidirectional RNNs will be described in detail.

\subsubsection{Bidirectional Recurrent Neural Networks}
One disadvantage of RecurrentNNs is that it can only use the previous context. In speech recognition, the entire sentence is transcribed at once, so the future context must be used. Bidirectional RNNs (BRecurrentNNs) \cite{75} \cite{76} use two separate hidden layers to process data in both directions to achieve it, and then forward these hidden layers to the same output layer.

\begin{figure}[H]
	\centering
	\includegraphics[width=0.9\textwidth]{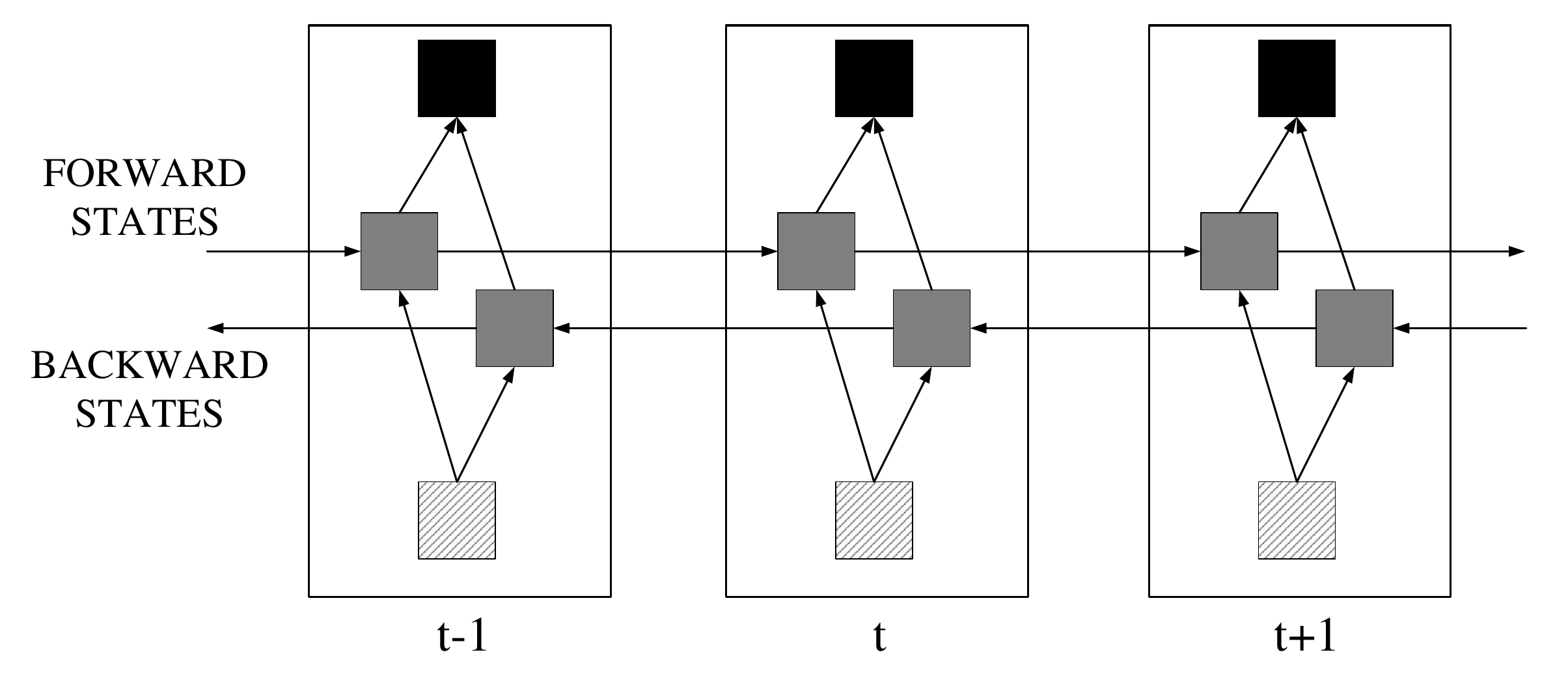}
	\caption{General structure of the BRecurrentNNs shown unfolded in time for three time steps.}
	\label{fig:bidirectional_rnn}
\end{figure}

The idea of BRecurrentNNs is to divide the state neurons of RecurrentNNs into the part responsible for the positive time direction (forward states) and the negative time direction (backward states). The outputs of the forward states are not connected to the input of the backward states, and vice versa. The above structure can be seen in Fig.~\ref{fig:bidirectional_rnn}, which unfolds along three time steps. Note that if there is no backward states, the structure is simplified to RecurrentNNs. If the forward states is removed, RecurrentNNs will be generated opposite to the time axis. When processing two time directions in the same network, the past and future input information of the current evaluation time frame can be directly used to minimize the objective function, without including the delay of future information.

The advantages of this model include: (1) It can be trained to a preset future frame without being restricted by input information. (2) Have faster development speed and better results.

\subsubsection{Bidirectional Recursive Neural Networks}
Extend the definition of the RecursiveNNs mentioned earlier to spread the information of the rest of the tree to each leaf node through the structure. In this way, decisions can be made on the leaf nodes based on the summary of the surrounding structure.

First, the upward level in the tree is represented as:

\begin{equation}
x_{\eta }^{\uparrow }=\sigma (W_{L}^{\uparrow }x_{l(\eta )}^{\uparrow }+W_{R}^{\uparrow }x_{r(\eta )}^{\uparrow }+{{b}^{\uparrow }})
\end{equation}

Note that if $\eta $ is a leaf node, $x_{\eta }^{\uparrow }$ is the initial representation ${{x}_{\eta }}$, which is similar to RecurrentNNs. Next, add a downward layer on top of this upward layer:

\begin{equation}
x_{\eta }^{\downarrow }=\left\{ \begin{matrix}
	\sigma (W_{L}^{\downarrow }x_{p(\eta )}^{\downarrow }+{{V}^{\downarrow }}x_{\eta }^{\uparrow }+{{b}^{\downarrow }}),if\,\eta \,is\,a\,left\,child  \\
	\sigma (W_{R}^{\downarrow }x_{p(\eta )}^{\downarrow }+{{V}^{\downarrow }}x_{\eta }^{\uparrow }+{{b}^{\downarrow }}),\ if\,\eta \,is\,a\,right\,child  \\
	\sigma ({{V}^{\downarrow }}x_{\eta }^{\uparrow }+{{b}^{\downarrow }}),\,\quad \quad \,if\,\eta \,is\,root(\eta =\rho )  \\
\end{matrix} \right.
\end{equation}

\begin{figure}[H]
	\centering
	\includegraphics[width=0.8\textwidth]{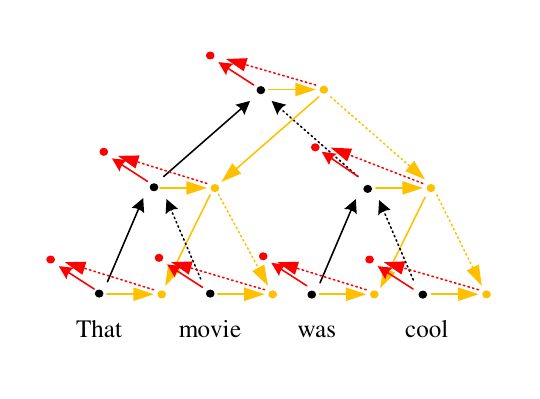}
	\caption{RecursiveNNs with bottom-up and top-down propagation. Black, orange and red indicate bottom-up, top-down, and output layers, respectively. The connection weights of the same color and style are shared.}
	\label{fig:bidirectional_recursive_nn}
\end{figure}

$p(\eta )$ is the parent node of $\eta $, $W_{L}^{\downarrow }$ and $W_{R}^{\downarrow }$ are connection weight matrices, respectively representing the downward connection of the parent node and the left and right child nodes, ${{V}^{\downarrow }}$ is the weight matrix, which connects any node upward and downward, and ${{b}^{\downarrow }}$ is the bias vector of the downward layer. Figure \ref{fig:bidirectional_recursive_nn} depicts a RecursiveNNs \cite{2} \cite{77}.

For any node, $x_{\eta }^{\uparrow }$ contains the subtree information rooted at $\eta $, and $x_{\eta }^{\downarrow }$ contains other information of the tree, so each node of the tree contributes to the calculation of $x_{\eta }^{\downarrow }$. Therefore, $x_{\eta }^{\uparrow }$ and $x_{\eta }^{\downarrow }$ can be considered as a complete information summary of the surrounding structure of $\eta $. At the leaf or any node in general, use the output layer to make the final decision:

\begin{equation}
{{y}_{n}}=\gamma ({{U}_{\downarrow }}x_{\eta }^{\downarrow }+{{U}_{\uparrow }}x_{\eta }^{\uparrow }+c)
\end{equation}

where ${{U}_{\downarrow }}$ and ${{U}_{\uparrow }}$ are the output weight matrices and $c$ is the output bias vector.

In the supervised task, supervision occurs in the output layer. Then, in the training process, the structure back propagation method \cite{78} is used to propagate the error upward, downward, downward, and upward. If necessary, the initial representation $x$ can be updated using the backpropagation error, which can be fine-tuned in the settings for the word vector representation.

Note that this definition is similar in structure to the expanded recursive autoencoder \cite{79}. However, the goals of these two structures are different. The expanded recursive autoencoder also has a downward propagation process, but the purpose is to reconstruct the initial representation. On the other hand, it is hoped that downward representation ${{x}^{\downarrow }}$ is as different as upward representation ${{x}^{\uparrow }}$ as possible, since the purpose is to capture information about the rest of the tree rather than a specific subtree under investigation. Therefore, the expanded recursive autoencoder does not use ${{x}^{\uparrow }}$ (except at the root) when calculating ${{x}^{\downarrow }}$, but the RecursiveNNs does.
.
	\section{Structured RNNs}
	Structured RNNs are divided into six categories. Grid RNNs form LSTM blocks into a multi-dimensional network. Each network contains a dimension, which not only calculates the time dimension but also calculates the depth dimension; Graph RNNs combine LSTM and CRNN to identify spatial structure learning dynamic patterns, not limited to image processing; Temporal RNNs are applied to basic time series classification tasks and label prediction problems; Lattice RNNs are mainly used for name entity recognition; Hierarchical RNNs mainly used in speech bandwidth expansion; Tree RNNs consider long and short distance interactions at different levels, and are used in language and image analysis.
	
	\subsection{Grid RNNs}
	Next, the various network architectures of Grid RNNs will be described in detail.
	
\subsubsection{Grid LSTM}
Grid LSTM \cite{80} \cite{81} was proposed by Kalchbrenner et al. \cite{82}. Different from the general LSTM model that composes LSTM blocks into a temporal chain, the grid LSTM model composes LSTM blocks into a multi-dimensional grid, so each grid contains a set of LSTM blocks in each dimension (including the depth dimension). This architecture introduces the gated linear dependence of each dimensional between the states of adjacent units, alleviating the problem of vanishing gradients in all dimensions.

\begin{figure}[H]
	\centering
	\includegraphics[width=0.8\textwidth]{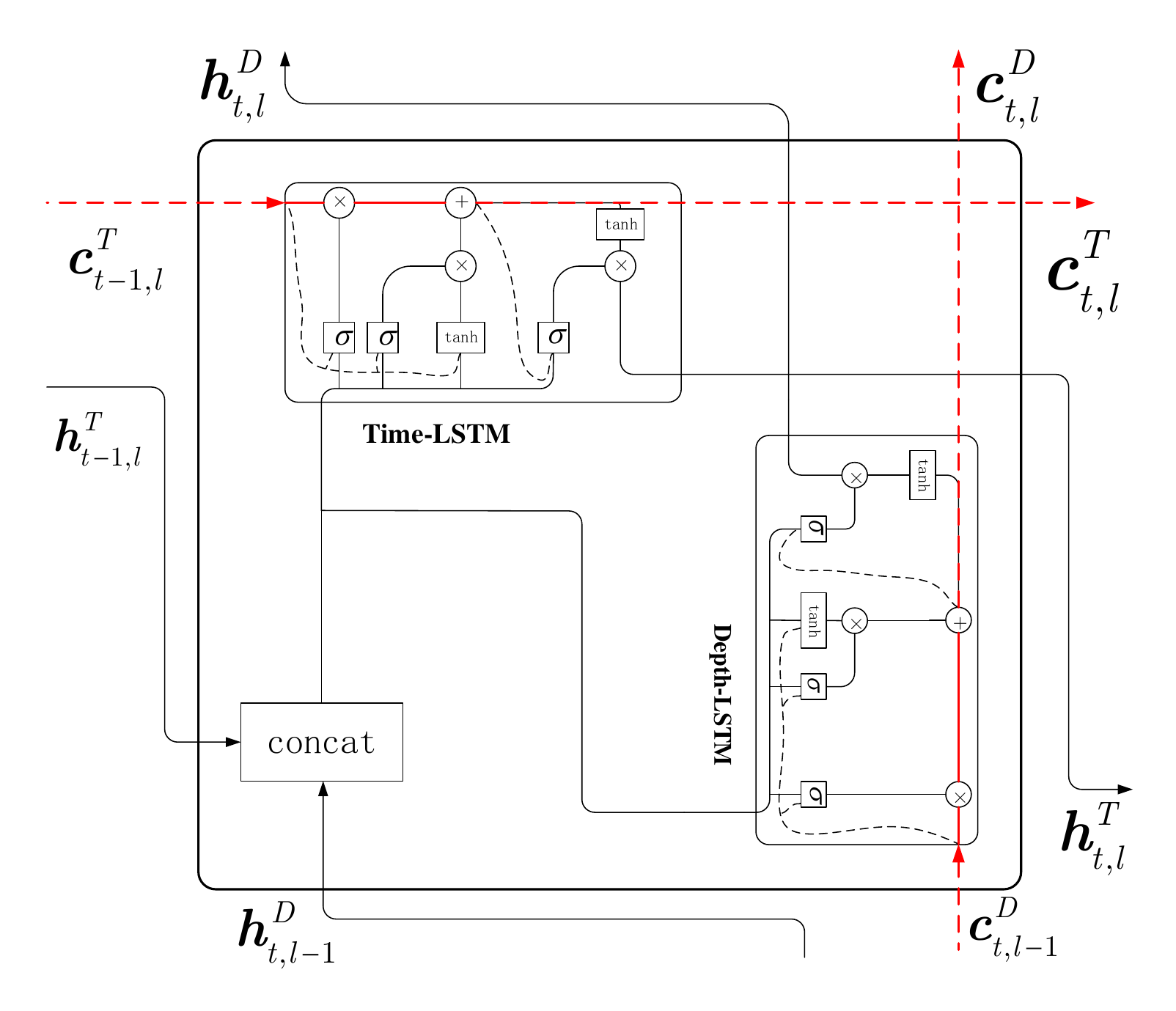}
	\caption{Grid LSTM}
	\label{fig:grid_lstm}
\end{figure}

Here, Wei-Ning Hsu et al. consider a two-dimensional grid LSTM model, which are the time and depth dimensions, as shown in Fig.~\ref{fig:grid_lstm}. The calculations in each grid are defined as follows:

\begin{equation}
{{x}_{t,l}}=\left[ h_{t,l-1}^{D};h_{t-1,l}^{T} \right]
\end{equation}

\begin{equation}
(h_{t,l}^{T},c_{t,l}^{T})=TIME-LSTM({{x}_{t,l}},c_{t-1,l}^{T},{{\Theta }^{T}})
\end{equation}

\begin{equation}
(h_{t,l}^{D},c_{t,l}^{D})=DEPTH-LSTM({{x}_{t,l}},c_{t,l-1}^{D},{{\Theta }^{D}})
\end{equation}

Note that subscripts are used to denote time and depth, and superscripts are used to denote a collection of specific LSTM blocks. $c_{t,l}^{i}$ and $h_{t,l}^{i}$ are the unit state and unit output at time $t$ and layer $l$ of the $i-LSTM$, respectively, and ${{\Theta }^{i}}$ represents all the parameters of the $i-LSTM$. The unit output $h_{t,L}^{D}$ of the last layer of $DEPTH-LSTM$ is passed to the softmax layer for classification.

The last thing that needs to be solved is $c_{t,0}^{D}$, its value is uncertain. The simplest solution is to set the value to zero, which provides flat initialization of the cell state regardless of the input value. Instead, a linear transformation can be applied:

\begin{equation}
c_{t,0}^{D}=Vh_{t,0}^{D}
\end{equation}

which has better performance in practice.

The advantages of this model include: (1) The depth can be extended in any dimension of the network. (2) The problem of vanishing gradients in all dimensions is alleviated.

\subsubsection{Prioritized Grid LSTM}
In the Eq.(39) in the previous section, the unit output of $TIME-LSTM$ is not classified at the current moment. In other words, the depth dimension should know the output of other dimensions in the current grid, making it implicitly deeper in the number of transformations before classification. Slightly modify the equation in the previous section to get:

\[x_{t,l}^{T}=\left[ h_{t,l-1}^{D};h_{t-1,l}^{T} \right]\]

\[(h_{t,l}^{T},c_{t,l}^{T})=TIME-LSTM(x_{t,l}^{T},c_{t-1,l}^{T},{{\Theta }^{T}})\]

\[x_{t,l}^{D}=\left[ h_{t,l-1}^{D};h_{t,l}^{T} \right]\]

\begin{equation}
(h_{t,l}^{D},c_{t,l}^{D})=DEPTH-LSTM(x_{t,l}^{D},c_{t,l-1}^{D},{{\Theta }^{D}})
\end{equation}

\begin{figure}[H]
	\centering
	\includegraphics[width=0.8\textwidth]{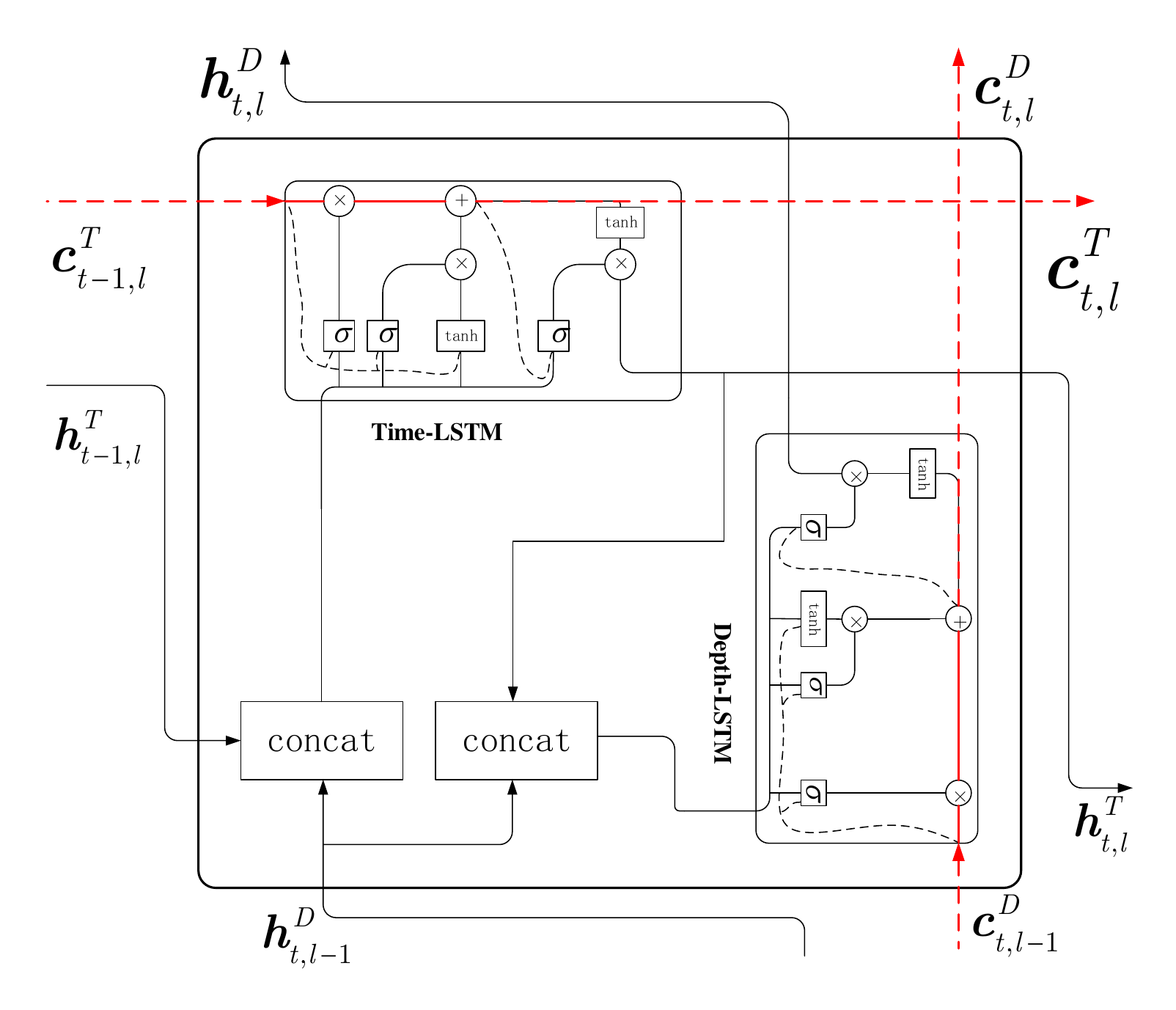}
	\caption{Prioritized Grid LSTM.}
	\label{fig:prioritized_grid_lstm}
\end{figure}

After processing the $TIME-LSTM$ of the same grid, the input of $DEPTH-LSTM$ is updated. Wei-Ning Hsu et al. call this model a prioritized grid LSTM \cite{80} and illustrate it in Fig.~\ref{fig:prioritized_grid_lstm}.

The advantages of this model include: (1) It can be calculated not only in the time dimension, but also in the depth dimension. (2) Prioritize the depth dimension in order to provide more updated information for the depth dimension. (3) Alleviate the problem of gradient disappearance.

\subsubsection{Frequency Dependent Grid-RNN}

\begin{figure}[H]
	\centering
	\includegraphics[width=0.8\textwidth]{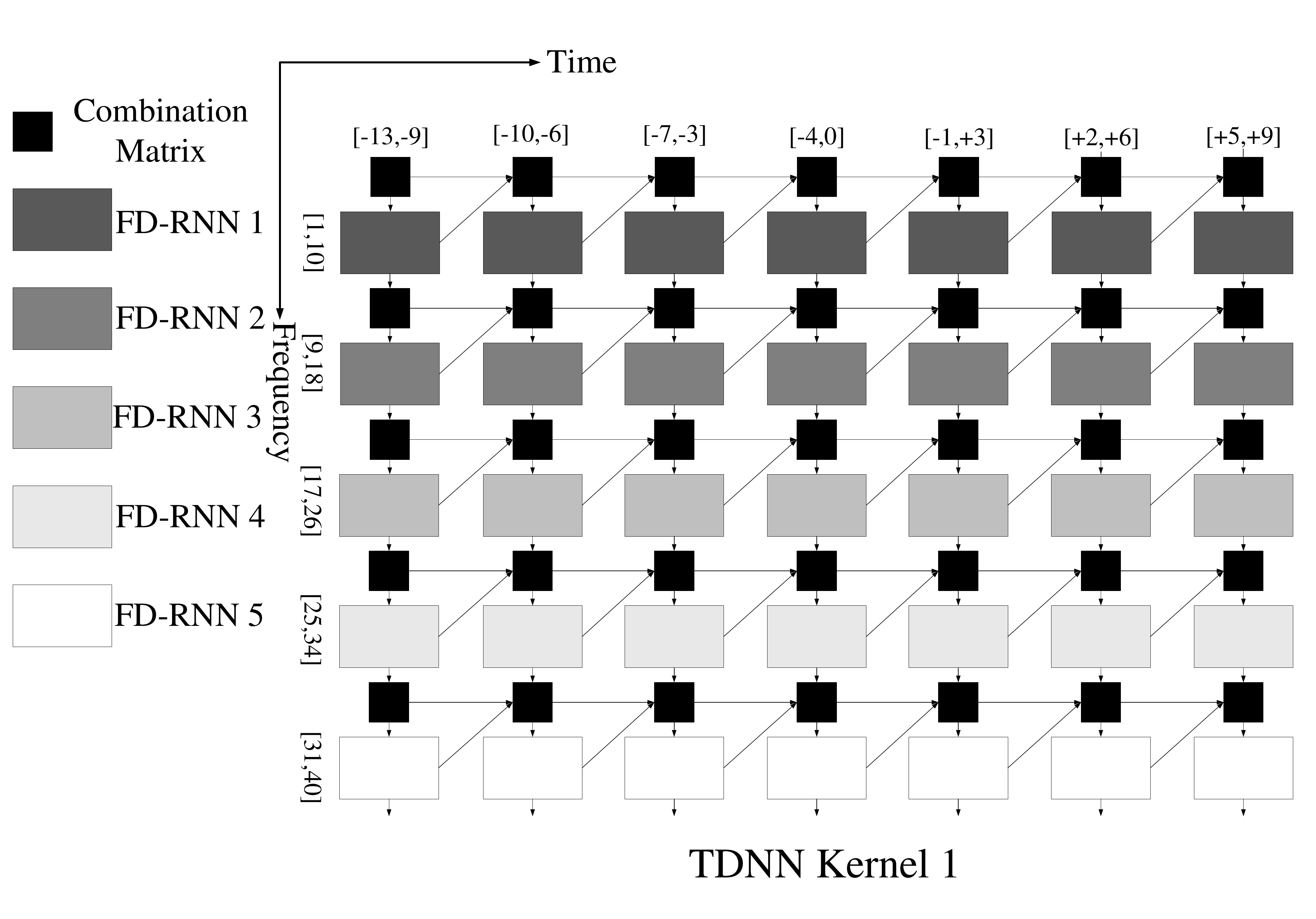}
	\caption{Frequency Dependent Grid-RNN Time Delay Neural Networks.}
	\label{fig:frequency_dependent_grid_rnn}
\end{figure}

Kreyssing et al. proposed a frequency-dependent grid RNN architecture \cite{83}, as shown in Fig.~\ref{fig:frequency_dependent_grid_rnn}. It uses RecurrentNNs and divides the input window $(t\in \left[ -13,9 \right])$ into the same seven time periods. Therefore, it can be cleverly combined with the time delay neural network structure. The frequency axis is divided into five blocks of size 10, as shown in Fig.~\ref{fig:frequency_dependent_grid_rnn}. The grid RNN is composed of two RNNs, one with $\sigma (\cdot )$ activation function and the other with linear activation function. $\sigma -RNN(h_{t,k}^{F})$ performs the feature extraction, and $C$ (Combination Matrix in Fig.~\ref{fig:frequency_dependent_grid_rnn}) models the information flow between $\tanh$ instantiations. The linear activation function is used to improve the information flow of the $\sigma (\cdot )$ activation function. The structure is trained in the expanded form shown by G. Saon et al. \cite{84}.

The advantages of this model include: (1) $Linear-RNN$ provides a linear path through the network. (2) It is helpful to untie the parameters of the $\sigma -RNN$ along the frequency axis.

\subsection{Graph RNNs}
Next, the various network architectures of Graph RNNs will be described in detail.

\subsubsection{Graph Convolutional Recurrent Network}
Seo et al. \cite{85} proposed two graph convolutional recurrent network (GCRN) \cite{86} \cite{87} \cite{88} architectures.

1)Model 1.The most straightforward definition is to stack the graph CNN for feature extraction and the LSTM for sequence learning:

\[x_{t}^{CNN}=CN{{N}_{\mathcal{G}}}({{x}_{t}})\]

\[i=\sigma ({{W}_{xi}}x_{t}^{CNN}+{{W}_{hi}}{{h}_{t-1}}+{{w}_{ci}}\odot {{c}_{t-1}}+{{b}_{i}})\]

\[f=\sigma ({{W}_{xf}}x_{t}^{CNN}+{{W}_{hf}}{{h}_{t-1}}+{{w}_{cf}}\odot {{c}_{t-1}}+{{b}_{f}})\]

\[{{c}_{t}}={{f}_{t}}\odot {{c}_{t-1}}+{{i}_{t}}\odot tanh({{W}_{xc}}x_{t}^{CNN}+{{W}_{hc}}{{h}_{t-1}}+{{b}_{c}})\]

\[o=\sigma ({{W}_{xo}}x_{t}^{CNN}+{{W}_{ho}}{{h}_{t-1}}+{{w}_{co}}\odot {{c}_{t}}+{{b}_{o}})\]

\begin{equation}
	h_{t} = o \odot \tanh(c_{t})
\end{equation}

In this model, the input matrix ${{x}_{t}}\in {{\mathbb{R}}^{n\times {{d}_{x}}}}$ represents the observed value measured by the dynamic system network ${{d}_{x}}$ in the graph $\mathcal{G}$ at time $t$. $x_{t}^{CNN}$ is the output of the graph CNN gate. In order to prove the concept, Seo et al. choose $x_{t}^{CNN}={{W}^{CNN}}{{*}_{\mathcal{G}}}{{x}_{t}}$, where ${{W}^{CNN}}\in {{\mathbb{R}}^{K\times {{d}_{x}}\times {{d}_{x}}}}$ is the Chebyshev coefficient of the support vector $K$ graph convolution kernel. The model gives the hidden variables of the spatial distribution and the unit state of size ${{d}_{h}}$ given by the matrix ${{c}_{t}}$ and ${{h}_{t}}\in {{\mathbb{R}}^{n\times {{d}_{h}}}}$. The peephole is controlled by ${{w}_{c\cdot }}\in {{\mathbb{R}}^{n\times {{d}_{h}}}}$. The weights ${{W}_{h\cdot }}\in {{\mathbb{R}}^{{{d}_{h}}\times {{d}_{h}}}}$ and ${{W}_{x\cdot }}\in {{\mathbb{R}}^{{{d}_{h}}\times {{d}_{x}}}}$ are the parameters of the fully connected layers. Architectures such as the above may be sufficient to capture the data distribution by using local stationarity, combined properties, and dynamic properties.

2)Model 2.In order to extend the convolutional LSTM model to the graph network, Seo et al. replaced Euclidean two-dimensional convolution $*$ with graph convolution ${{*}_{\mathcal{G}}}$:

\[i=\sigma ({{W}_{xi}}{{*}_{\mathcal{G}}}{{x}_{t}}+{{W}_{hi}}{{*}_{\mathcal{G}}}{{h}_{t-1}}+{{w}_{ci}}\odot {{c}_{t-1}}+{{b}_{i}})\]

\[f=\sigma ({{W}_{xf}}{{*}_{\mathcal{G}}}{{x}_{t}}+{{W}_{hf}}{{*}_{\mathcal{G}}}{{h}_{t-1}}+{{w}_{cf}}\odot {{c}_{t-1}}+{{b}_{f}})\]

\[{{c}_{t}}={{f}_{t}}\odot {{c}_{t-1}}+{{i}_{t}}\odot tanh({{W}_{xc}}{{*}_{\mathcal{G}}}{{x}_{t}}+{{W}_{hc}}{{*}_{\mathcal{G}}}{{h}_{t-1}}+{{b}_{c}})\]

\[o=\sigma ({{W}_{xo}}{{*}_{\mathcal{G}}}{{x}_{t}}+{{W}_{ho}}{{*}_{\mathcal{G}}}{{h}_{t-1}}+{{w}_{co}}\odot {{c}_{t}}+{{b}_{o}})\]

\begin{equation}
{{h}_{t}}=o\odot tanh({{c}_{t}})
\end{equation}

In this model, the graph convolution kernel support vector $K$ defined by the Chebyshev coefficient ${{W}_{h\cdot }}\in {{\mathbb{R}}^{K\times {{d}_{h}}\times {{d}_{h}}}}$ and ${{W}_{x\cdot }}\in {{\mathbb{R}}^{K\times {{d}_{h}}\times {{d}_{x}}}}$ determines the number of parameters, which is independent of the number $n$ of nodes. In order to keep the symbol simple, Seo et al. used ${{W}_{xi}}{{*}_{\mathcal{G}}}{{x}_{t}}$ to represent the graph convolution of ${{x}_{t}}$ and ${{d}_{h}}{{d}_{x}}$ filters, which is a function of Laplacian $L$ parameterized by Chebyshev coefficient $K$. In a distributed computing setting, $K$ controls the communication overhead, that is, the number of nodes that any given node should exchange in order to calculate its local state.

The advantages of this model include: (1) The isotropic filter can outperform the classic two-dimensional filter with fewer parameters. (2) The combination of graphs with CNN and RNNs is a multi-functional method to introduce and utilize secondary information by constructing a data matrix. (3) Simultaneous use of graphic spatial information and dynamic data information can improve learning accuracy and learning speed.

\subsubsection{Temporal Dynamic Graph LSTM}
\begin{figure}[H]
	\centering
	\includegraphics[width=0.9\textwidth]{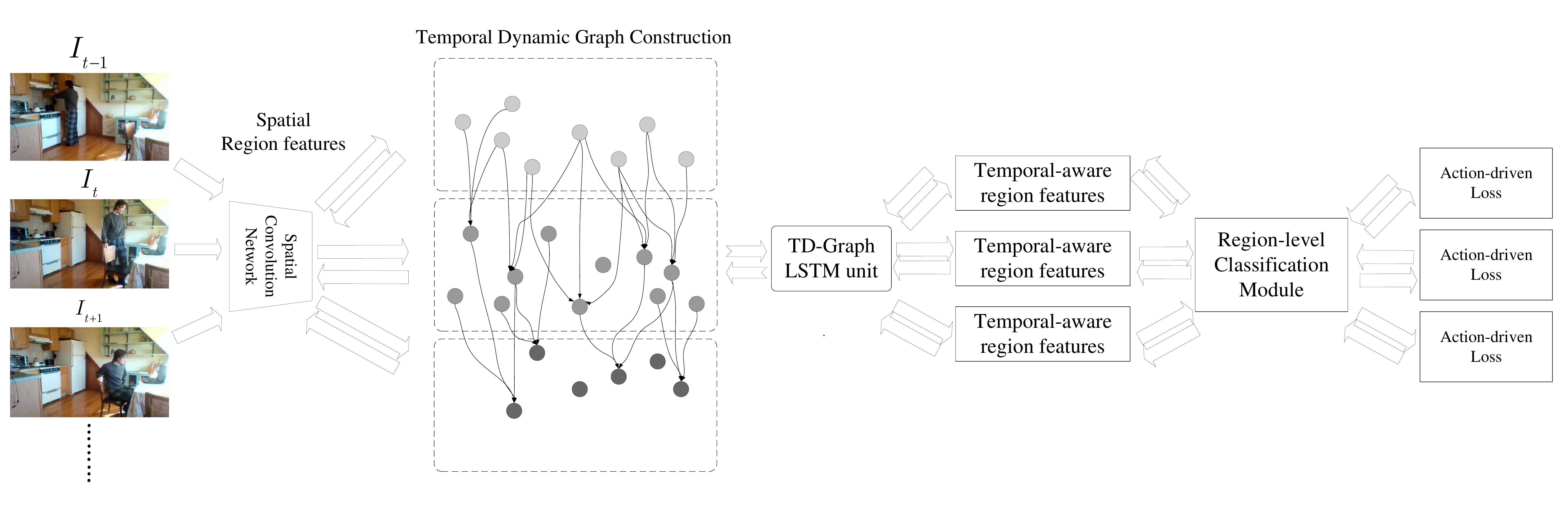}
	\caption{TD-Graph LSTM.}
	\label{fig:td_graph_lstm}
\end{figure}

Yuan Yuan et al. established a fully differentiable temporal dynamic graph LSTM (TD-Graph LSTM) \cite{89} framework for action-driven video target detection tasks. Fig.~\ref{fig:td_graph_lstm} gives an overview of TD-Graph LSTM \cite{90} \cite{91} \cite{92} \cite{93}. Each frame in the input video first passes through a spatial convolution network to obtain the spatial visual features of the region proposals. Based on visual features, similar regions in two consecutive frames are found and correlated to represent the same object across temporal domains. Connect all semantically similar regions in two consecutive frames to construct a time graph structure, where graph nodes are represented by region proposals. Then, use the TD-Graph LSTM unit to recurrently circulate the information on the entire time graph, where the LSTM unit takes the spatial visual features as the input state. Thanks to graph topology, TD-Graph LSTM can incorporate temporal motion patterns for participating objects in actions in a more effective and meaningful way. TD-Graph LSTM outputs enhanced temporal-aware features of all regions. Then the region-level classification is used to generate classification confidence. Finally, these region-level predictions can be aggregated to generate frame-level object category predictions, which are supervised by the object category in the action label. The action-driven object classification loss can therefore be back propagated to all regions in the video as a whole, where the prediction of each frame can benefit each other.

The advantages of this model include: (1) Global temporal reasoning can be realized. (2) Alleviate the label missing problem of each individual frame.

\subsection{Temporal RNNs}
Next, the various network architectures of Temporal RNNs will be described in detail.

\subsubsection{Connectionist Temporal Classification}
For tasks such as speech recognition, if the correspondence between input and label is unknown, RecurrentNNs has so far been limited to auxiliary roles. The problem is that standard training methods require that each input correspond to a separate output, which is usually impossible. To solve this problem, the traditional solution is to use the hidden Markov model to generate the target of the RNN, and then invert the output of the RNN to provide the observation probability.

Connectionist Temporal Classification (CTC) \cite{94} is specifically used for temporal series classification tasks \cite{95}, that is, for sequence data label prediction problems where the correspondence between input and target labels is unknown. The CTC output layer uses the discriminant loss function to train the RecurrentNNs network to connect the weight matrix and bias, predict the class label of the unknown sequence, and does not use the hidden Markov model.

Given a voice input sequence $x$, the speech output sequence is $y$, and the target output sequence is $z$. The length of $x$ is $T$, the length of $z$ is $U$, and the number of possible phonemes is $K$. CTC uses the softmax layer to define a separate output distribution $Pr(\left. k \right|t)$ at each time step $t$ along the input sequence. This distribution covers $K$ phonemes plus an additional blank symbol $\phi $, $\phi $ represents non-output (so the size of the softmax layer is $K+1$). Therefore, the network can decide whether to emit any tags or not to emit tags at every time step. In short, this defines the alignment distribution between the input sequence and the target sequence. Then, CTC uses a forward-backward algorithm to sum all possible alignments and determines the normalized probability $Pr(\left. z \right|x)$ of the target sequence given the input sequence. Similar procedures have been used elsewhere in speech and handwriting recognition to integrate possible segmentation; however, the difference with CTC is that it completely ignores segmentation, but adds up the single-time step label results.

RNNs trained with CTC are usually bidirectional to ensure that each $Pr(\left. k \right|t)$ depends on the entire input sequence, not just from input to $t$. According to the deep bidirectional network, $Pr(\left. k \right|t)$ is defined as follows:

\begin{equation}
{{y}_{t}}={{W}_{{{{\overrightarrow{h}}}^{N}}y}}\overrightarrow{h}_{t}^{N}+{{W}_{{{{\overleftarrow{h}}}^{N}}y}}\overleftarrow{h}_{t}^{N}+{{b}_{y}}
\end{equation}

\begin{equation}
Pr(\left. k \right|t)=\frac{exp({{y}_{t}}[k])}{\sum\nolimits_{{k}'=1}^{K}{exp({{y}_{t}}[{k}'])}}
\end{equation}

Here, ${{y}_{t}}[k]$ is the $k$-th element of the irregular output vector ${{y}_{t}}$ of length $K+1$, and $N$ is the number of bidirectional levels.

CTC uses a neural network to model all aspects of the sequence, and does not need to combine the network with a hidden Markov model, nor does it need to preprocess the training data, and it does not need to extract labels sequence from the network output through external post-processing. The practice of speech and handwriting recognition shows that a bidirectional LSTM network with a CTC output layer can effectively predict sequence labels, usually better than standard hidden Markov models.

\subsubsection{Spatio-Temporal LSTM}
RecurrentNNs have shown their advantages in modeling the complex dynamics of human activities as time series data, and have achieved good performance in the recognition of human behavior based on skeleton \cite{96} \cite{97} \cite{98}. In the existing literature, RecurrentNNs are mainly used in the temporal domain to discover the discriminative dynamic characteristics and action patterns of action recognition. However, there is also different spatial information in the joint position and posture configuration of each video frame, and the sequence nature of body joints can apply RecurrentNNs-based modeling to the spatial domain.

\begin{figure}[H]
	\centering
	\includegraphics[width=0.8\textwidth]{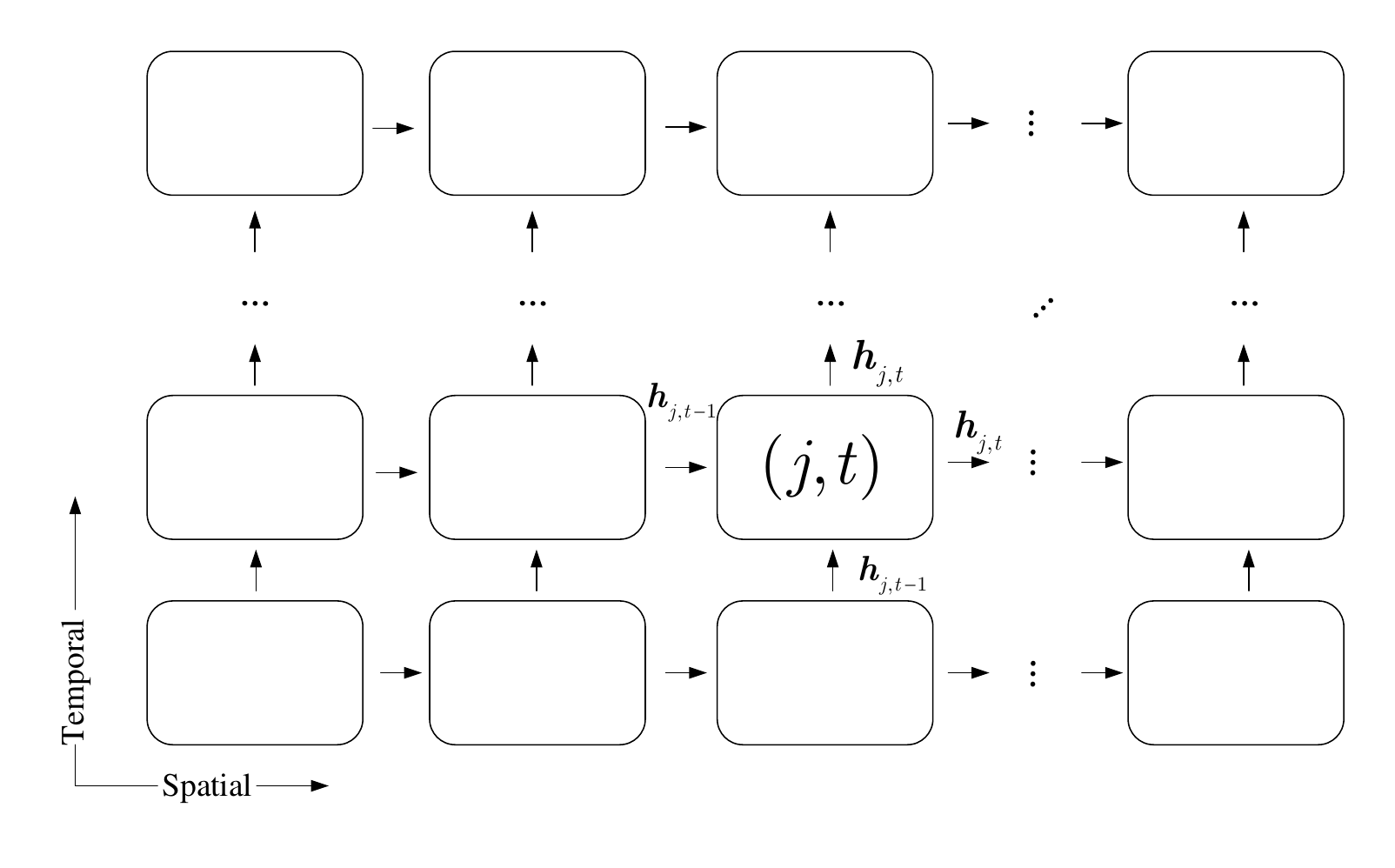}
	\caption{Illustration of the spatio-temporal LSTM network. In the temporal dimension, the corresponding body joints are fed to the frame. In the spatial dimension, the skeleton joints in each frame are fed sequentially. Each unit receives the hidden representation of the previous joint and the hidden representation of the same joint in the previous frame.}
	\label{fig:st_lstm_structure}
\end{figure}

Different from the existing method of connecting joint information as a whole, Jun Liu et al. extended the recurrent analysis to the spatial domain by discovering the spatial dependence pattern between different body joints, so they proposed a spatio-temporal LSTM (ST-LSTM) network \cite{99}, which can simultaneously simulate the temporal dependence between different frames and the spatial dependence of different joints at the same frame. Each ST-LSTM unit corresponds to a body joint, and it receives the hidden representation of its own joint at the previous time step, as well as the hidden representation of the previous joint in the current frame. The pattern of this model is shown in Fig.~\ref{fig:st_lstm_structure}.

\begin{figure}[H]
	\centering
	\includegraphics[width=0.7\textwidth]{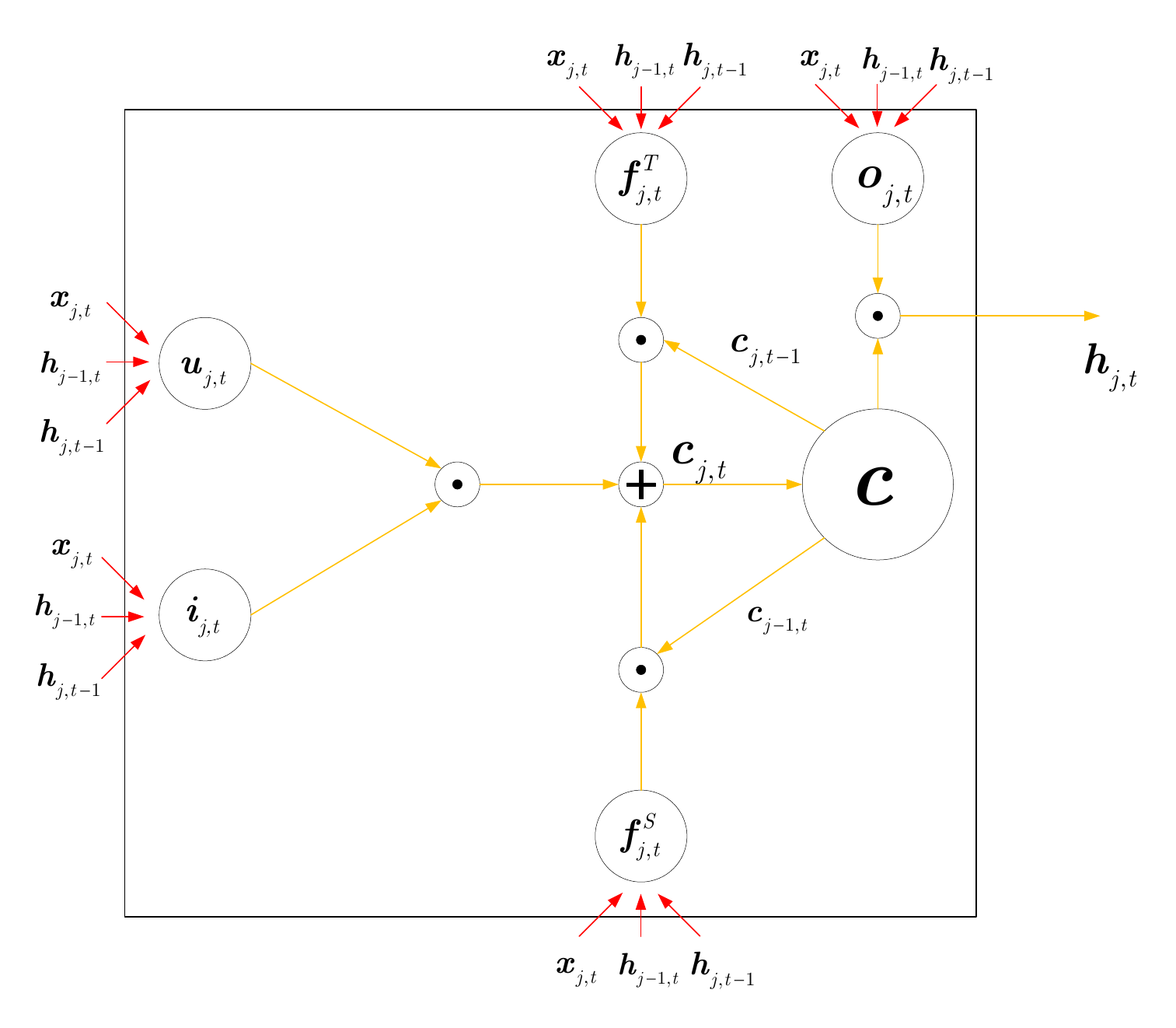}
	\caption{Illustration of the proposed ST-LSTM with one unit.}
	\label{fig:st_lstm_unit}
\end{figure}

First, suppose the joints are arranged in a simple chain sequence. Use $j$ and $t$ to represent joints and frames respectively, $j\in \left\{ 1,\ldots ,J \right\}$,$t\in \left\{ 1,\ldots ,T \right\}$. Each ST-LSTM unit is represented by the input (${{x}_{j,t}}$, the information of the corresponding joint at the current time step), the previous joint hidden $({{h}_{j-1,t}})$ at the current time step and the same joint hidden $({{h}_{j,t-1}})$ at the previous time step. As shown in Fig.~\ref{fig:st_lstm_unit}, each unit also has two forget gates, $f_{j,t}^{T}$ and $f_{j,t}^{S}$, which deal with contextual information sources in the temporal and spatial dimensions, respectively. ${{u}_{j,t}}$ is the modulated input. The ST-LSTM transition equation is as follows:

\begin{equation}
\left( \begin{matrix}
	{{i}_{j,t}}  \\
	f_{j,t}^{S}  \\
	f_{j,t}^{T}  \\
	{{o}_{j,t}}  \\
	{{u}_{j,t}}  \\
\end{matrix} \right)=\left( \begin{matrix}
	\sigma   \\
	\sigma   \\
	\sigma   \\
	\sigma   \\
	\tanh   \\
\end{matrix} \right)\left( M\left( \begin{matrix}
	{{x}_{j,t}}  \\
	{{h}_{j-1,t}}  \\
	{{h}_{j,t-1}}  \\
\end{matrix} \right) \right)
\end{equation}

\begin{equation}
{{c}_{j,t}}={{i}_{j,t}}\odot {{u}_{j,t}}+f_{j,t}^{S}\odot {{c}_{j-1,t}}+f_{j,t}^{T}\odot {{c}_{j,t-1}}
\end{equation}

\begin{equation}
{{h}_{j,t}}={{o}_{j,t}}\odot tanh({{c}_{j,t}})
\end{equation}

The advantages of this model include: (1) Better simultaneous analysis of information hidden sources related to actions in human skeleton sequences. (2) The reliability of the sequence data can be learned, and the influence of the input data on the long-term context representation update process stored in the unit memory unit can be adjusted accordingly.

\subsubsection{Temporal Attention Model (TAM) for Feature Learning}
\begin{figure}[H]
	\centering
	\includegraphics[width=0.6\textwidth]{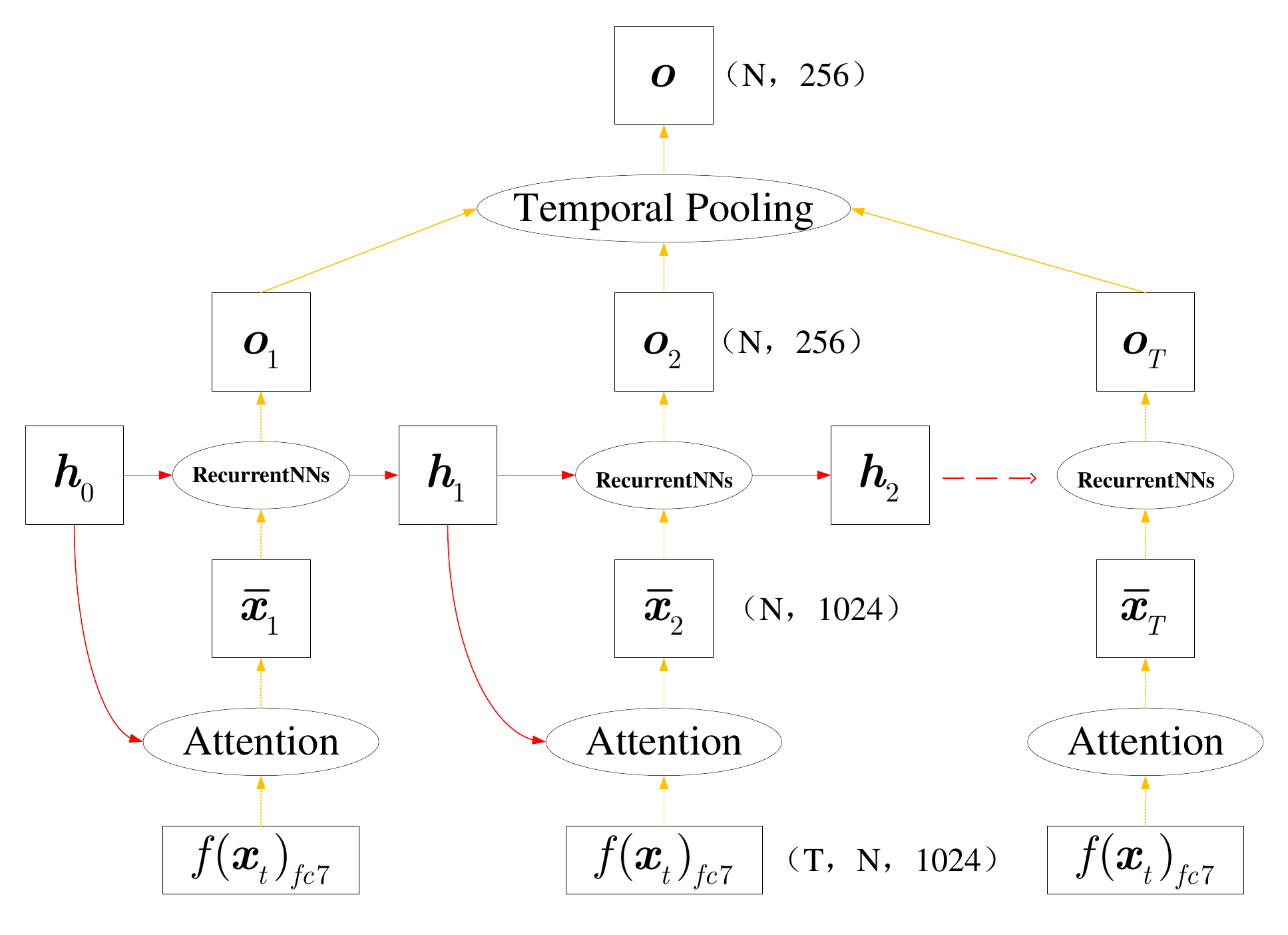}
	\caption{The structure of TAM. The input is $T$ feature maps of the $fc7$-th layer of the image sequence $x$. $N$ is the data size of one batch. After the attention subnet, the weighted average ${{\bar{x}}_{t}}$ of the feature $T$ is obtained. Then ${{\bar{x}}_{t}}$ is fed into the RecurrentNNs, and the RecurrentNNs outputs the feature ${{o}_{t}}$ at each time step, and the final representation of $x$ is the temporal average pool of $\left\{ {{o}_{t}} \right\}_{t=1}^{T}$.}
	\label{fig:tam_structure}
\end{figure}

In order to selectively focus on the most relevant images, the attention mechanism is applied to explore the temporal structure of a given image sequence. The entire process of TAM \cite{100} is shown in Fig.~\ref{fig:tam_structure}.

\begin{figure}[H]
	\centering
	\includegraphics[width=0.8\textwidth]{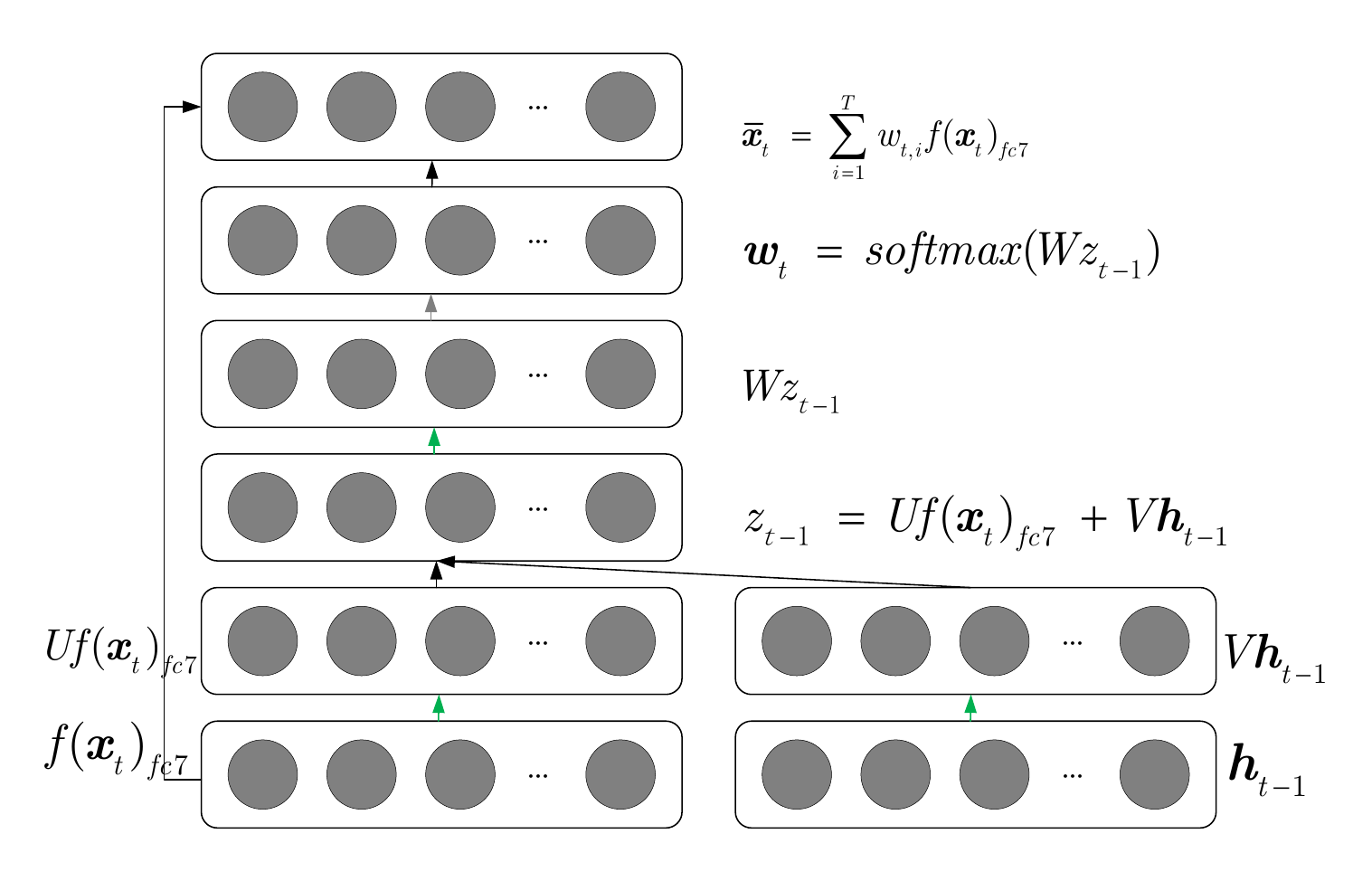}
	\caption{The subnet used to learn the correlation of each image in the image sequence. The subnet is represented by ${{w}_{t}}=\left\{ {{w}_{t,i}} \right\}_{i=1}^{T}$. The green line indicates that they are fully connected, the black line indicates the sum or inner product of the elements, and the gray line indicates the softmax operation.}
	\label{fig:tam_subnet}
\end{figure}

Assuming that each image sequence is represented as $x=\left\{ {{x}_{t}}\left| {{x}_{t}}\in {{\mathbb{R}}^{D}} \right. \right\}_{t=1}^{T}$, $T$ is the length of the image sequence, and $D$ is the dimension of the image. Let $f(x)$ denote the CNN. This model consists of two parts, namely attention unit and RecurrentNNs unit. At each time step $t$, the attention unit accepts $\left\{ f{{({{x}_{t}})}_{fc7}} \right\}_{t=1}^{T}$ as input and generates a weighted average of these features, namely:

\begin{equation}
{{\bar{x}}_{t}}=\sum\limits_{i=1}^{T}{{{w}_{t,i}}f{{({{x}_{t}})}_{fc7}}}
\end{equation}

where $\left\{ {{w}_{t,i}} \right\}$ is learned by the subnet, as shown in Fig.~\ref{fig:tam_subnet}. ${{h}_{t-1}}$ is the hidden state of RecurrentNNs at time step $t-1$. $Uf{{({{x}_{t}})}_{fc7}}$, $V{{h}_{t-1}}$ and $W{{z}_{t-1}}$ are obtained through the fully connected layer. The softmax operation is used to guarantee $\sum\nolimits_{i}{{{w}_{t,i}}=1}$.

Then ${{\bar{x}}_{t}}$ is fed to RecurrentNNs, which adopts the LSTM structure \cite{101}, which can summarize useful information in long-distance sequences. The final representation of the image sequence is the temporal average pool of each output \cite{102}:

The advantages of this model include: (1) Through a temporal attention model, the most discriminative frame in a given video is automatically extracted. (2) The temporal attention model is conducive to feature learning, and the spatial recurrent model is conducive to metric learning.

\subsection{Lattice RNNs}
Next, the various network architectures of Lattice RNNs will be described in detail.

\subsubsection{Character Lattice Model}
\begin{figure}[H]
	\centering
	\includegraphics[width=0.9\textwidth]{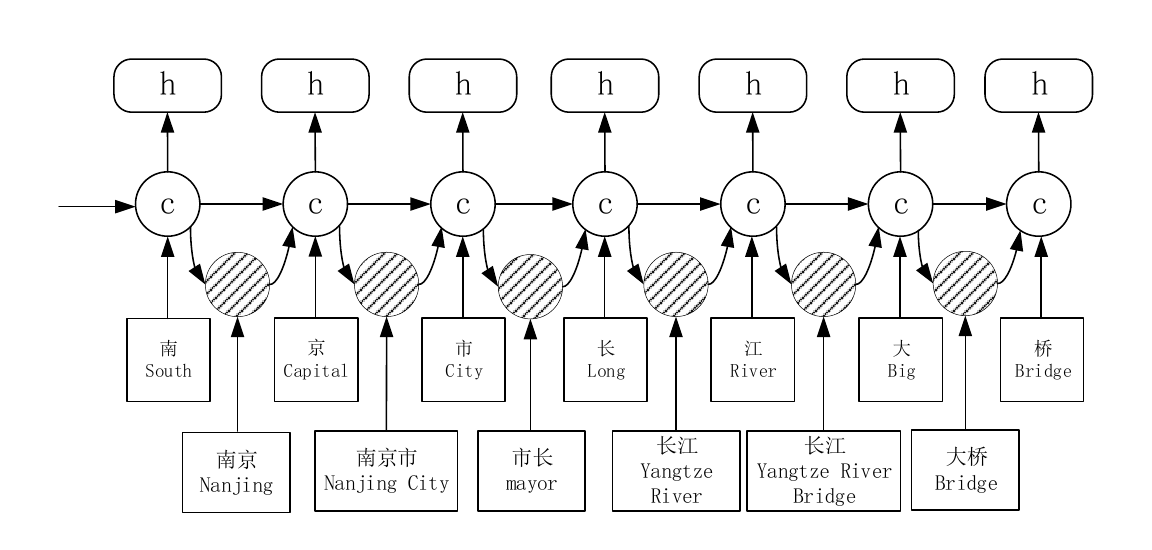}
	\caption{Lattice LSTM structure.}
	\label{fig:lattice_lstm}
\end{figure}

The overall structure of the character lattice model is shown in Fig.~\ref{fig:lattice_lstm} \cite{103} \cite{104} \cite{105}. It can be seen as an extension of the character model, integrating character units and additional gates that control the flow of information.

The input of the model is the character sequence ${{c}_{1}},{{c}_{2}},\ldots ,{{c}_{m}}$, and all character subsequences matching words in dictionary $\mathbb{D}$, where ${{c}_{j}}$ represents the $j$-th character. The model involves four types of vectors: input vector, output hidden vector, unit vector and gate vector. In the character-based model, the character input vector is used to represent each character ${{c}_{j}}$ in the character-based model:

\begin{equation}
x_{j}^{c}={{e}^{c}}({{c}_{j}})
\end{equation}

${{e}^{c}}$ represents the same character embedding lookup table. Use character unit vector $c_{j}^{c}$ and hidden vector $h_{j}^{c}$ on each ${{c}_{j}}$ to construct the basic recurrent structure of the model, where $c_{j}^{c}$ is used to record the recurrent information flow from the beginning of the sentence to ${{c}_{j}}$, and $h_{j}^{c}$ is used for conditional random field sequence marking.

However, unlike character-based models, the calculation of $c_{j}^{c}$ now considers the dictionary subsequence $w_{b,e}^{d}$ in the sentence. Each subsequence $w_{b,e}^{d}$ is expressed as follows:

\begin{equation}
x_{b,e}^{w}={{e}^{w}}(w_{b,e}^{d})
\end{equation}

where ${{e}^{w}}$ represents the same character embedding lookup table.

In addition, the word unit $c_{b,e}^{w}$ is used to represent the recurrent state of $x_{b,e}^{w}$ from the beginning of the sentence. The calculation formula of $c_{b,e}^{w}$ is as follows:

\begin{equation}
\left[ \begin{aligned}
	& i_{b,e}^{w} \\ 
	& f_{b,e}^{w} \\ 
	& \tilde{c}_{b,e}^{w} \\ 
\end{aligned} \right]=\left[ \begin{aligned}
	& \ \sigma  \\ 
	& \ \sigma  \\ 
	& tanh \\ 
\end{aligned} \right]({{W}^{w\top }}\left[ \begin{aligned}
	& x_{b,e}^{w} \\ 
	& h_{b}^{c} \\ 
\end{aligned} \right]+{{b}^{w}})
\end{equation}

\begin{equation}
c_{b,e}^{w}=f_{b,e}^{w}\odot c_{b}^{c}+i_{b,e}^{w}\odot \tilde{c}_{b,e}^{w}
\end{equation}

where $i_{b,e}^{w}$ and $f_{b,e}^{w}$ are the set of input and forget gates. There are no output gates for word units because tags are only executed at the character level.

With $c_{b,e}^{w}$, there are more recurrent paths to flow information into each $c_{j}^{c}$. For example, in Fig.~\ref{fig:lattice_lstm}, the input sources of $c_{7}^{c}$ include $x_{7}^{c}$ (桥 Bridge),$c_{6,7}^{w}$ (大桥 Bridge) and $c_{4,7}^{w}$ (长江大桥 Yangtze River Bridge). Connect all $c_{b,e}^{w}$ ($b\in \left\{ {b}'\left| w_{{b}',e}^{d}\in \mathbb{D} \right. \right\}$) to unit $c_{e}^{c}$, and then use an additional gate $i_{b,e}^{c}$ for each sub-sequence unit $c_{b,e}^{w}$ to control its contribution to $c_{b,e}^{c}$:

\begin{equation}
i_{b,e}^{c}=\sigma ({{W}^{l\top }}\left[ \begin{aligned}
	& x_{e}^{c} \\ 
	& c_{b,e}^{w} \\ 
\end{aligned} \right]+{{b}^{l}})
\end{equation}

Therefore, the word unit value $c_{j}^{c}$ is calculated as follows:

\begin{equation}
c_{j}^{c}=\sum\limits_{b\in \left\{ {b}'\left| w_{{b}',j}^{d}\in \mathbb{D} \right. \right\}}{\alpha _{b,j}^{c}}\odot c_{b,j}^{w}+\alpha _{j}^{c}\odot \tilde{c}_{j}^{c}
\end{equation}

In the above formula, by setting the sum to 1, the gate value $i_{b,j}^{c}$ and $i_{j}^{c}$ are normalized to $\alpha _{b,j}^{c}$ and $\alpha _{j}^{c}$.

\begin{equation}
\alpha _{b,j}^{c}=\frac{exp(i_{b,j}^{c})}{exp(i_{j}^{c})+\sum\nolimits_{{b}'\in \left\{ {b}''\left| w_{{b}'',j}^{d}\in \mathbb{D} \right. \right\}}{exp(i_{{b}',j}^{c})}}
\end{equation}

\begin{equation}
\alpha _{j}^{c}=\frac{exp(i_{j}^{c})}{exp(i_{j}^{c})+\sum\nolimits_{{b}'\in \left\{ {b}''\left| w_{{b}'',j}^{d}\in \mathbb{D} \right. \right\}}{exp(i_{{b}',j}^{c})}}
\end{equation}

The final hidden vector $h_{j}^{c}$ is calculated in the same way as the general LSTM. During the named entity recognition (NER) \cite{106} training period, the loss value is back propagated to the parameters ${{W}^{c}}$, ${{b}^{c}}$, ${{W}^{w}}$, ${{b}^{w}}$, ${{W}^{l}}$ and ${{b}^{l}}$ making the model dynamically focus on more relevant words during NER marking.

The advantages of this model include: (1) Free choice of words in the context for disambiguation. (2) Use word information more effectively.

\subsubsection{Lattice Language Models}
Jacob Buckman et al. \cite{107} defined some terms, using the term "token" represented by ${{x}_{i}}$ to describe any indivisible items in the vocabulary that do not have other vocabulary items as components. The term "component" represented by ${{k}_{i}}$ or $x_{i}^{j}$ is used to describe a sequence of one or more tokens. This component represents a part of the complete string $X$ and contains unit tokens from ${{x}_{i}}$ to ${{x}_{j}}$: $x_{i}^{j}=[{{x}_{i}},{{x}_{i+1}},\ldots ,{{x}_{j}}]$, which also defines a "token vocabulary" (contains only a subset of the token vocabulary) and "component vocabulary" (similarly includes all components).

\begin{figure}[H]
	\centering
	\includegraphics[width=0.8\textwidth]{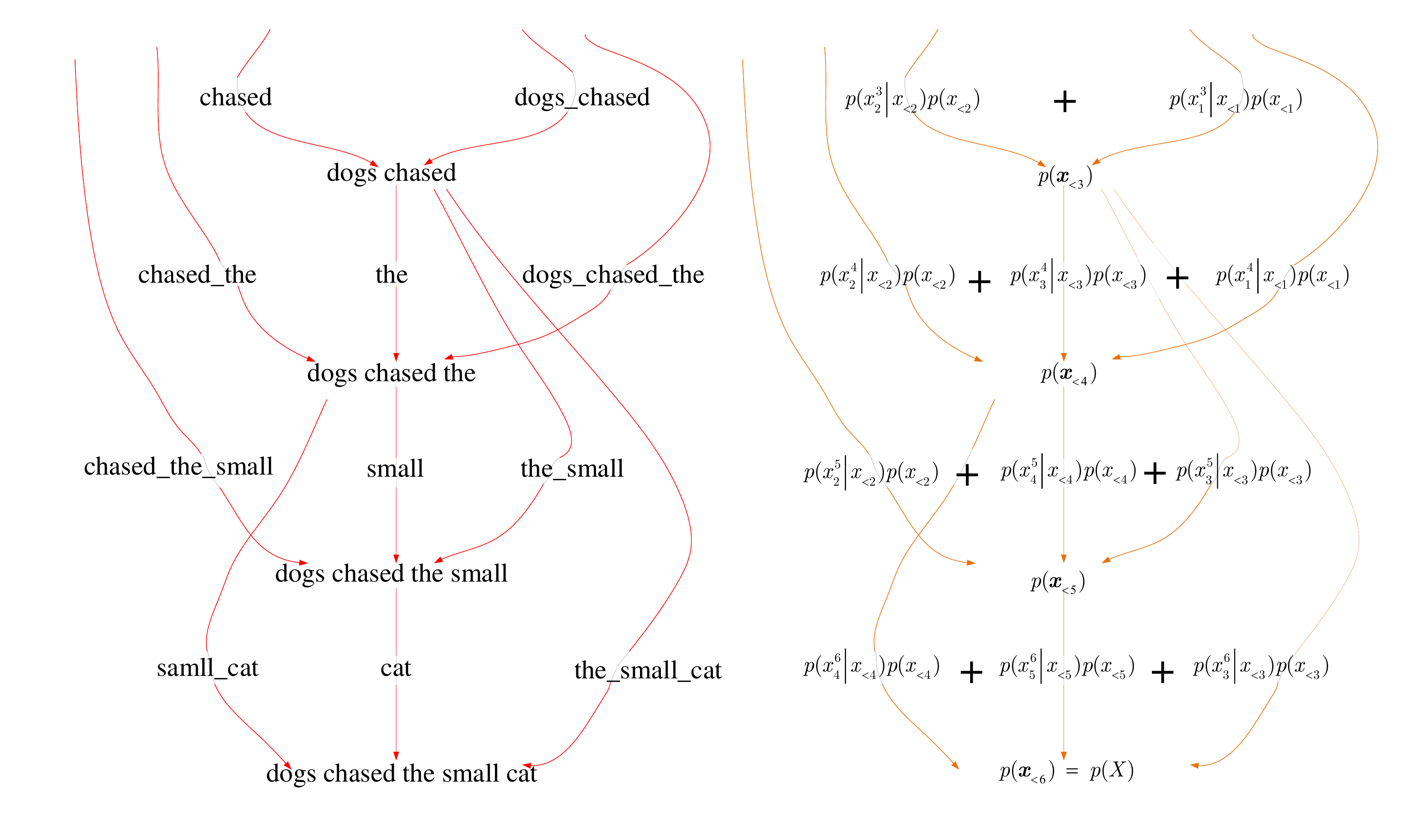}
	\caption{The lattice decomposition of a sentence and its corresponding lattice language model probability calculation.}
	\label{fig:lattice_decomposition}
\end{figure}

All segments of $X$ can be expressed as the edge of the lattice path on the label prefix of $X:{{x}_{<1}},{{x}_{<2}},\ldots ,X$. The infimum is the empty prefix ${{x}_{<1}}$; the supremum is $X$; if and only if $x_{i}^{j}$ exists in the component vocabulary, there is an edge from prefix ${{x}_{<i}}$ to the prefix ${{x}_{<j}}$, make $[{{x}_{<i}};x_{i}^{j}]={{x}_{<j}}$. As shown in Fig.~\ref{fig:lattice_decomposition}, all paths through the lattice from ${{x}_{<i}}$ to $X$ are segments from $X$ to the token list on the traversed edge.

The probability of a particular prefix $p({{x}_{<j}})$ is calculated by marginalizing all segments before ${{x}_{j-1}}$:

\begin{equation}
p({{x}_{<j}})=\sum\limits_{S\in S({{x}_{<j}})}{\prod\limits_{t=1}^{\left| S \right|}{p(x_{{{s}_{t-1}}}^{{{s}_{t}}-1}\left| {{x}_{<{{s}_{t-1}}}} \right.)}}
\end{equation}

which defines ${{s}_{\left| S \right|}}=j$. The key to effectively calculating this probability is that this is a recursive formula, so it can be marginalized on the previous edge in lattice ${{A}_{j}}$ instead of marginalizing all segments. Each item in ${{A}_{j}}$ is a position $i\,(={{s}_{t-1}})$, which indicates that there is an edge between the prefix ${{x}_{<i}}$ and the prefix ${{x}_{<j}}$ corresponding to the token $x_{i}^{j}$ in the lattice. Therefore, $p({{x}_{<j}})$ can be calculated as

\begin{equation}
p({{x}_{<j}})=\sum\limits_{i\in {{A}_{j}}}{p({{x}_{<i}})}p(x_{i}^{j}\left| {{x}_{<i}} \right.)
\end{equation}

Since $X$ is the supremum prefix node, $p(X)$ can be calculated by using the formula $j=\left| X \right|$. To do this, it is necessary to calculate the probability of each of its $\left| X \right|$ predecessors. Each one needs to calculate $\left| X \right|$, which means that the calculation of $p(X)$ can be completed in $O({{\left| X \right|}^{2}})$. If it can be guaranteed that each node has the maximum number of input edges $D$, then there is $\left| {{A}_{j}} \right|\le D$ for all $j$, and the boundary can be reduced to $O(D\left| X \right.)$ time.

\begin{figure}[H]
	\centering
	\includegraphics[width=0.6\textwidth]{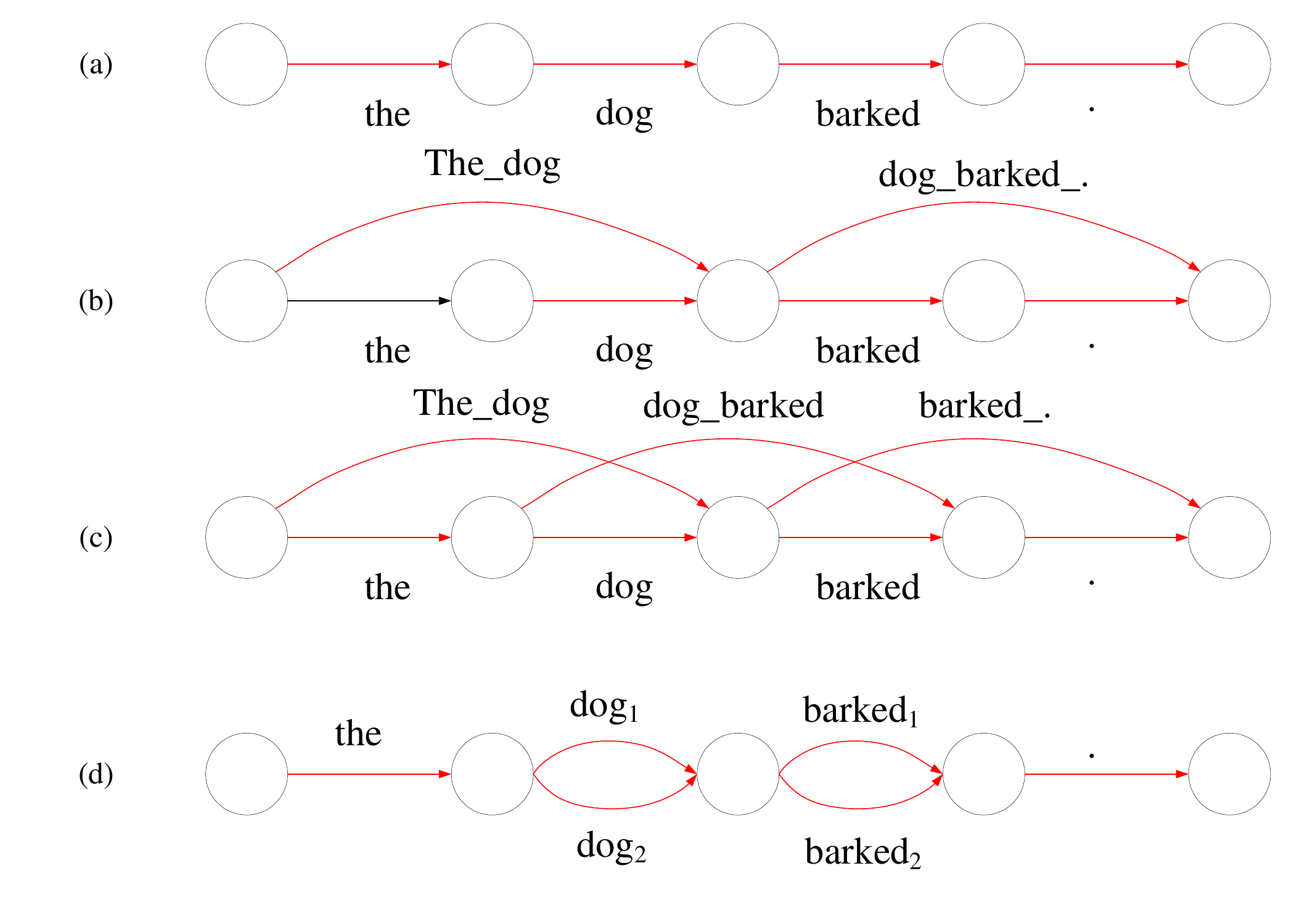}
	\caption{Examples of (a) single-path lattice, (b) sparse lattice, (c) dense lattice with D = 2 and (d) multiple lattices with D = 2 in the sentence "dog barking".}
	\label{fig:lattice_variants}
\end{figure}

The model proposed \cite{108} above is agnostic to the shape of the lattice. Fig.~\ref{fig:lattice_variants} illustrates several potential lattice variants. Depending on how the lattice is constructed, this method is useful in many different contexts.

\subsubsection{Neural Lattice Language Models}
There is still a lack of part in the process of trying to apply the neural language model to the lattice. In the overall probability of equation(111), the probability $P(t | p)$ of the next segment must be calculated based on historical records. However, considering that there may be multiple paths in the lattice leading to $p$, this is not as simple as the case where only one segment is possible. Previous work on lattice-based language models \cite{109} \cite{110} \cite{111} used a count-based $n$-gram model that only relied on each step in a limited historical context, so that the marginal probability is calculated in an accurate and effective way through dynamic programming. On the other hand, recurrent neural models rely on the entire context, causing them to lack this ability. Therefore, the main technical contribution of Jacob Buckman et al. is to describe the merging of lattices into a neural framework with infinite context by providing a method to approximate the hidden state of a RecurrentNNs. Technical methods currently include: direct approximation, Monte Carlo approximation, marginal approximation, Gumbel-Softmax interpolation.

The advantages of this model include: (1) All segments of the sentence are marginalized in an end-to-end manner. (2) Whether it is used to merge multiple phrases or embed multiple words, the neural lattice language model is better than the LSTM-based baseline in language modeling tasks.

\subsubsection{Lattice Recurrent Unit}
The Lattice Recurrent Unit (LRU) \cite{112} model aims to have different information flows in time and depth dimensions. The LRU model can be seen as an extension of the GRU model.

The first LRU model is called Projected State LRU (PS-LRU), which will only decouple the projected state of each dimension. The second model is called Reset Gate LRU (RG-LRU), and it will further decouple the reset gate. The last one is called Update Gate LRU (UG-LRU or LRU), which can decouple all three components, including the update gate.

(1) PS-LRU

As the first model PS-LRU, the projection state $\hat{h}$ is decoupled to create two new projection states ${{\hat{h}}_{1}}$ and ${{\hat{h}}_{2}}$. Each is used to calculate a separate output state: ${h}'$ and ${{{h}'}_{2}}$. Formally, use the following update function to define PS-LRU:

\begin{equation}
{{\hat{h}}_{1}}=tanh\left( [W_{1}^{h}\ \ U_{1}^{h}]\left[ \begin{aligned}
	& \ \ {{h}_{1}} \\ 
	& r\odot {{h}_{2}} \\ 
\end{aligned} \right] \right)
\end{equation}

\begin{equation}
{{\hat{h}}_{2}}=tanh\left( [W_{2}^{h}\ \ U_{2}^{h}]\left[ \begin{aligned}
	& r\odot {{h}_{1}} \\ 
	& \ \ {{h}_{2}} \\ 
\end{aligned} \right] \right)
\end{equation}

\begin{equation}
{{{h}'}_{1}}=z\odot {{h}_{1}}+(1-z)\odot {{\hat{h}}_{2}}
\end{equation}

\begin{equation}
{{{h}'}_{2}}=z\odot {{h}_{2}}+(1-z)\odot {{\hat{h}}_{1}}
\end{equation}

The PS-LRU model uses the same update and reset gates for the two output states. Note that since there are two different outputs in this model, the hidden state of GRU output is decomposed into equation(114) and (115).

(2)RG-LRU

The reset gate is further decoupled, and two new gates ${{r}_{1}}$ and ${{r}_{2}}$ are obtained. This model is called RG-LRU and is defined as follows:

\begin{equation}
{{r}_{1}}=\sigma \left( \left[ W_{1}^{r}\ \ U_{1}^{r} \right]\left[ \begin{aligned}
	& {{h}_{1}} \\ 
	& {{h}_{2}} \\ 
\end{aligned} \right] \right)
\end{equation}

\begin{equation}
{{r}_{2}}=\sigma \left( \left[ W_{2}^{r}\ \ U_{2}^{r} \right]\left[ \begin{aligned}
	& {{h}_{1}} \\ 
	& {{h}_{2}} \\ 
\end{aligned} \right] \right)
\end{equation}

\begin{equation}
{{\hat{h}}_{1}}=\tanh \left( [W_{1}^{h}\quad U_{1}^{h}]\left[ \begin{aligned}
	& \ \ \,{{h}_{1}} \\ 
	& {{r}_{2}}\odot {{h}_{2}} \\ 
\end{aligned} \right] \right)
\end{equation}

\begin{equation}
{{\hat{h}}_{2}}=\tanh \left( [W_{2}^{h}\quad U_{2}^{h}]\left[ \begin{aligned}
	& {{r}_{1}}\odot {{h}_{1}} \\ 
	& \ \ \,\,{{h}_{2}} \\ 
\end{aligned} \right] \right)
\end{equation}

Note that since $r$ is now decoupled, the hidden state is decomposed into equation (118) and (119).

(3)UG-LRU
In the final model UG-LRU, all three main components of the GRU are decoupled, including the update gate, and the ${{z}_{1}}$ and ${{z}_{2}}$ gate are defined as:

\begin{equation}
{{z}_{1}}=\sigma \left( \left[ W_{1}^{z}\ \ U_{1}^{z} \right]\left[ \begin{aligned}
	& {{h}_{1}} \\ 
	& {{h}_{2}} \\ 
\end{aligned} \right] \right)
\end{equation}

\begin{equation}
{{z}_{2}}=\sigma \left( \left[ W_{2}^{z}\ \ U_{2}^{z} \right]\left[ \begin{aligned}
	& {{h}_{1}} \\ 
	& {{h}_{2}} \\ 
\end{aligned} \right] \right)
\end{equation}

\begin{equation}
{{{h}'}_{1}}={{z}_{1}}\odot {{\hat{h}}_{2}}+(1-{{z}_{1}})\odot {{h}_{1}}
\end{equation}

\begin{equation}
{{{h}'}_{2}}={{z}_{2}}\odot {{\hat{h}}_{1}}+(1-{{z}_{2}})\odot {{h}_{2}}
\end{equation}

Note that equation(114) and (115) need to be modified slightly because $z_t^T$ and $z_t^D$ are now decoupled.

The advantages of this model include: (1) When dealing with small data sets, it has the best accuracy, convergence speed and statistical efficiency among all baseline models. (2) Realize the goal of multi-layer network with limited learning resources by creating information flow along time and depth.

\subsection{Hierarchical RNNs}
Next, the various network architectures of Hierarchical RNNs will be described in detail.

\subsubsection{Hierarchical RNNs}
\begin{figure}[H]
	\centering
	\includegraphics[width=1.0\textwidth]{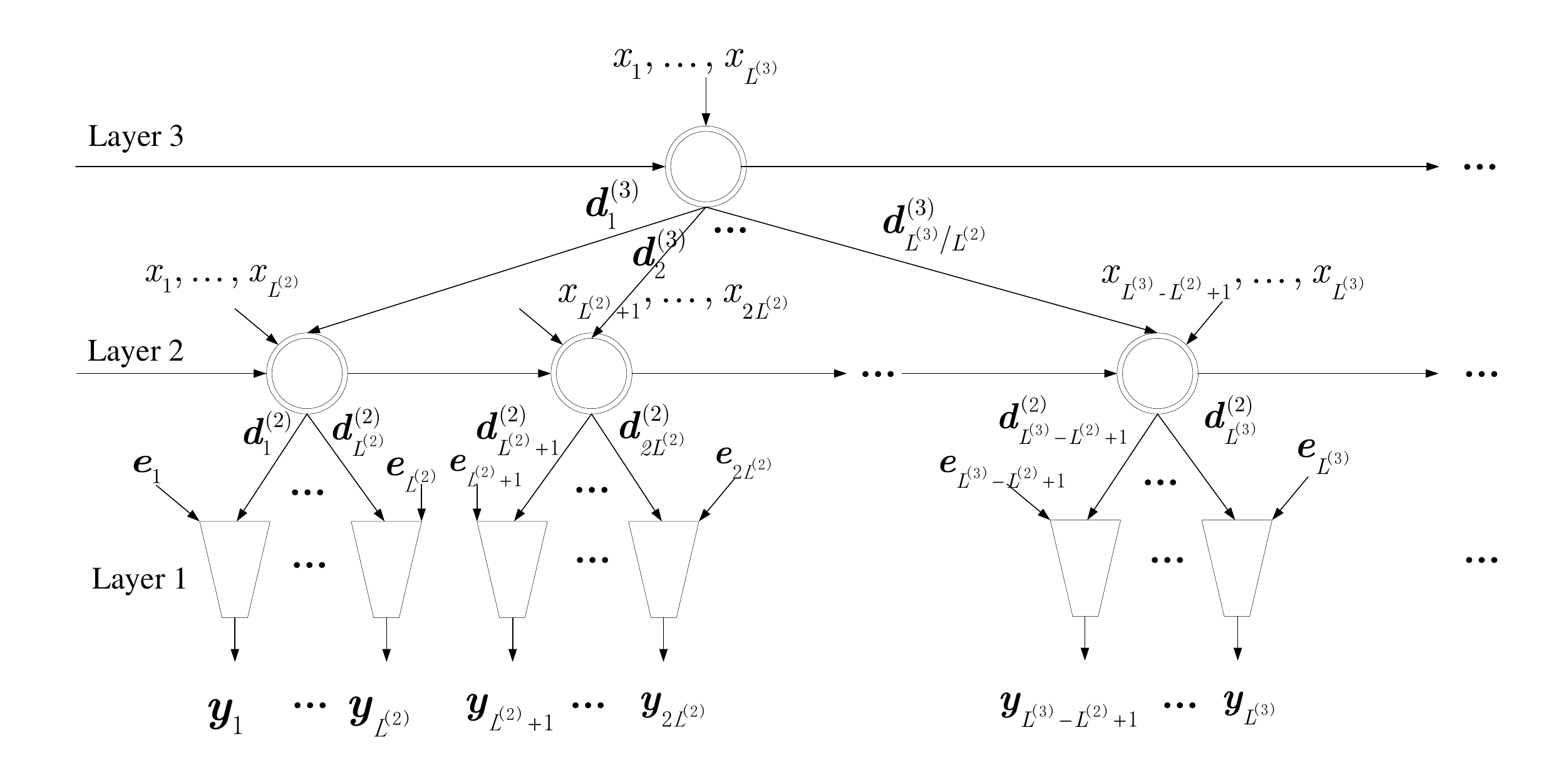}
	\caption{The structure of HRNNs for BWE..}
	\label{fig:hrnn_structure}
\end{figure}

The hierarchical RNN (HRNN) \cite{113} \cite{114} \cite{115} structure of speech bandwidth extension (BWE) is shown in Fig.~\ref{fig:hrnn_structure}. HRNN is composed of LSTM layer and feed-forward (FF) layer. These LSTM and FF layers in HRNN form a multi-layered hierarchical structure, and each layer works with a specific time resolution. The bottom layer (i.e., layer 1 in Figure \ref{fig:hrnn_structure}) processes a single sample and outputs sample-level predictions. Each higher layer operates at a lower temporal resolution (i.e., processes more samples each time step). Except for the top layer, each layer sets conditions on the layer above it. The model structure is similar to the sampleRNN \cite{116}, the main difference is that the original sampleRNN model is an unconditional audio generator, which uses the history of the output waveform as the network input and generates the output waveform in an autoregressive way. However, the HRNN model shown in Fig.~\ref{fig:hrnn_structure} directly describes the mapping relationship between the two waveform sequences without considering the autoregressive characteristics of the output waveform. This HRNN structure is specifically designed for BWE, because the narrowband waveform is used as input in this task. Removal of autoregressive connections helps reduce computational complexity and facilitate parallel calculations during generation.

The concentric circles in the figure above represent the LSTM layer, and the inverted trapezoid represents the FF layer.

Assume that the HRNN has $K$ layers in total (for example, $K=3$ in Fig.~\ref{fig:hrnn_structure}). The first layer is at the sample level, and the other $K-1$ layers are at the frame level, and their working time resolution is lower than the samples.

\subsubsection{Conditional Hierarchical RNNs}
\begin{figure}[H]
	\centering
	\includegraphics[width=0.8\textwidth]{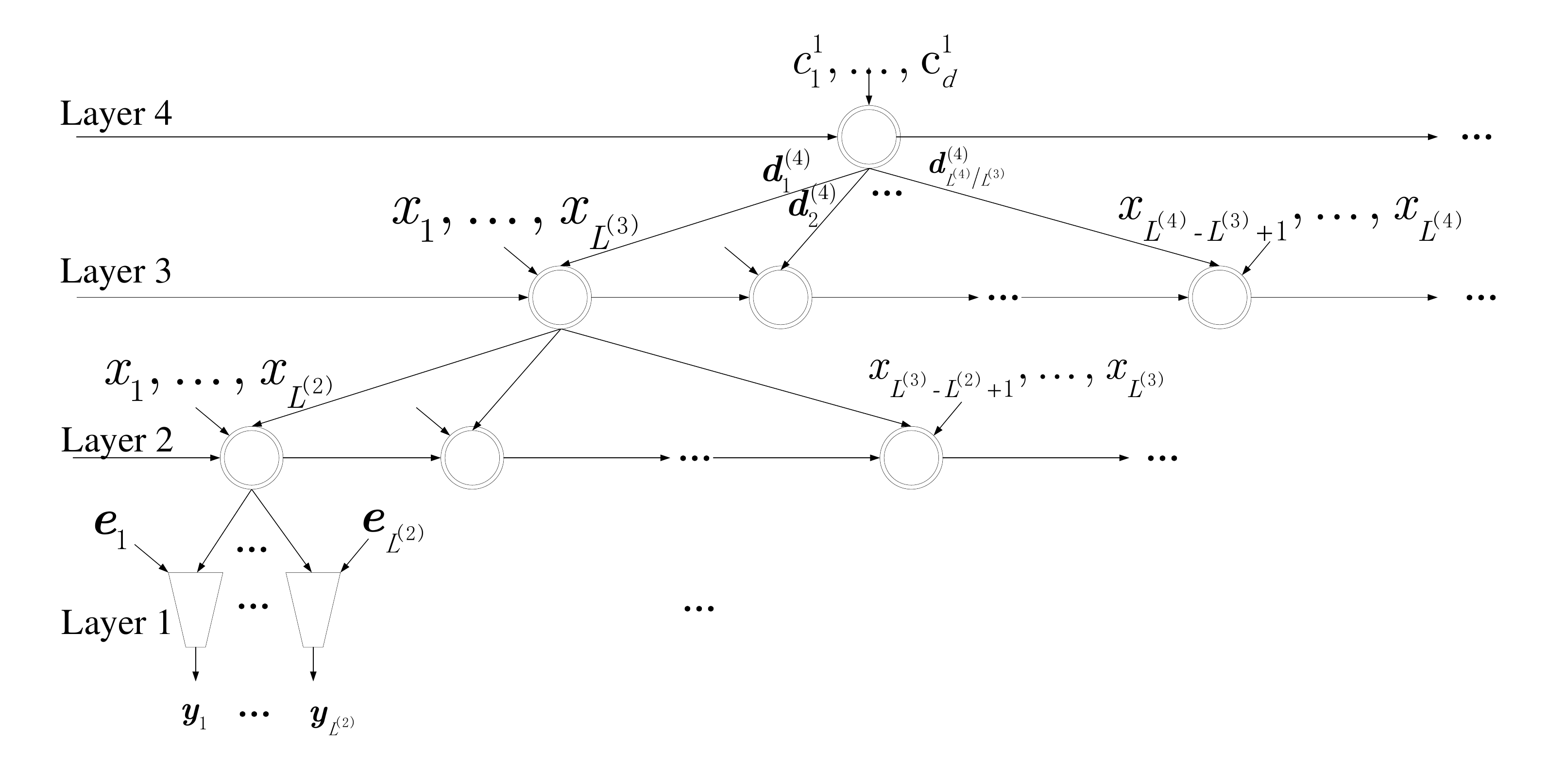}
	\caption{The structure of conditional HRNNs for BWE.}
	\label{fig:conditional_hrnn}
\end{figure}

Some frame-level auxiliary features extracted from input narrowband waveforms, such as bottleneck (BN) features \cite{117}, have shown that they are very effective in improving the performance of vocoder-based BWE \cite{118}. In order to combine these auxiliary inputs with the HRNN model, a conditional HRNN structure is designed, as shown in Fig.~\ref{fig:conditional_hrnn}.

Compared with HRNNs, conditional HRNNs \cite{113} \cite{119} \cite{120} add a conditional layer at the top. The input feature of the conditional layer is the frame-level auxiliary feature vector extracted from the input waveform, not the waveform sample. Assume that the total number of layers in the condition HRNN is $K$ (for example, $K=4$ in Fig.~\ref{fig:conditional_hrnn}), and let ${{L}^{(K)}}$ represent the frame shift of the auxiliary input feature.

\subsubsection{Hierarchical LSTM}
Hierarchical LSTM (H-LSTM) \cite{121} \cite{122} \cite{123} contains two coupled sub-networks: Pixel LSTM (P-LSTM) and Multi-scale Super-pixel LSTM (MS-LSTM) for processing surface labeling and relationship prediction. These two subnets provide complementary information to develop hierarchical scene contexts, and they are jointly optimized to improve performance.

\begin{figure}[H]
	\centering
	\includegraphics[width=0.9\textwidth]{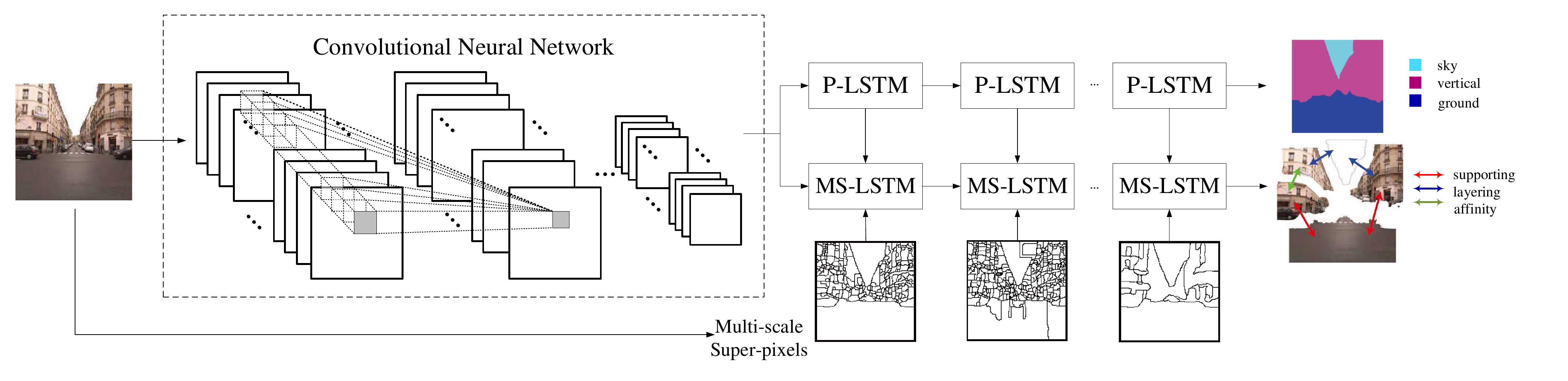}
	\caption{The proposed recurrent framework for geometric scene analysis. Each still image is first fed into several convolutional layers. Then these feature maps are respectively transferred to the stacked P-LSTM layer and MS-LSTM to generate the geometric surface labeling of each pixel and the interaction relationship between the regions.}
	\label{fig:hierarchical_lstm}
\end{figure}

As shown in Fig.~\ref{fig:hierarchical_lstm}, the input image first passes through a stack of convolutional layers and pooling layers to generate a set of convolutional feature maps. Then P-LSTM and MS-LSTM take these feature maps as input in the sharing mode, and output the pixel-wise geometric surface labeling and the interaction relationship between adjacent regions.

Each LSTM unit in the $i$-th layer receives input ${{x}_{i}}$ from the previous state and determines the current state composed of the hidden unit ${{h}_{i+1}}\in {{\mathbb{R}}^{d}}$ and the memory unit ${{c}_{i+1}}\in {{\mathbb{R}}^{d}}$ where $d$ is the dimension of the network output. Similar to the work in Graves et al. \cite{44}, Zhanglin Peng et al. use ${{g}^{u}}$, ${{g}^{f}}$, ${{g}^{c}}$, and ${{g}^{o}}$ to denote input, forget, memory and output gates respectively. Define ${{W}^{u}}$, ${{W}^{f}}$, ${{W}^{c}}$, ${{W}^{o}}$ as the corresponding recurrent gate weight. Therefore, the hidden unit and memory unit of the current state can be calculated in the following way:

\begin{equation}
{{g}^{u}}=\phi ({{W}^{u}}*{{H}_{i}})
\end{equation}

\begin{equation}
{{g}^{f}}=\phi ({{W}^{f}}*{{H}_{i}})
\end{equation}

\begin{equation}
{{g}^{o}}=\phi ({{W}^{o}}*{{H}_{i}})
\end{equation}

\begin{equation}
{{g}^{c}}=tanh({{W}^{c}}*{{H}_{i}})
\end{equation}

\begin{equation}
{{c}_{i+1}}={{g}^{f}}\odot {{c}_{i}}+{{g}^{u}}\odot {{g}^{c}}
\end{equation}

\begin{equation}
{{h}_{i+1}}=tanh({{g}^{o}}\odot {{c}_{i}})
\end{equation}

where ${{H}_{i}}$ represents the concatenation of input ${{x}_{i}}$ and previous state ${{h}_{i}}$. $\phi $ is the sigmoid function in the form of $\phi (t)={1}/{(1+{{e}^{-t}})}\;$ represents the element-wise product. According to Kalchbrenner et al. \cite{82}, the expressions (124)-(129) can be simplified to:

\begin{equation}
({{c}_{i+1}},{{h}_{i+1}})=LSTM({{H}_{i}},{{c}_{i}},W)\
\end{equation}

where $W$ is the concatenation of four different types of recurrent gate weights.

The advantages of this model include: (1) Capturing different levels of context and better performing local reasoning. (2) A new recurrent neural network model for geometric scene analysis is proposed, which combines geometric surface labeling and relationship prediction.

\subsubsection{Hierarchical Recurrent Attention Network}
\begin{figure}[H]
	\centering
	\includegraphics[width=0.9\textwidth]{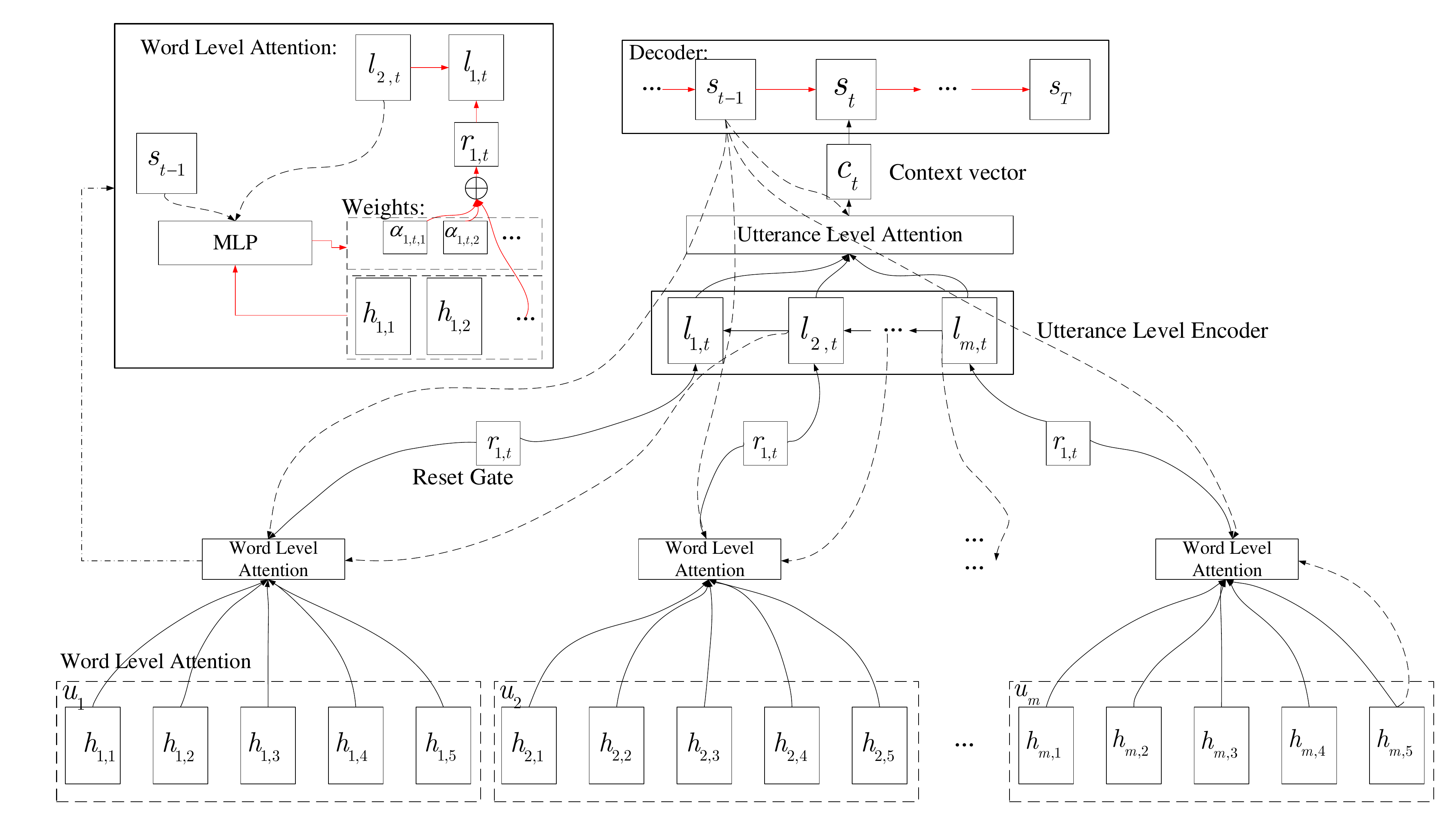}
	\caption{HRAN.}
	\label{fig:hran_structure}
\end{figure}

Chen Xing et al. proposed Hierarchical Recurrent Attention Network (HRAN) \cite{124} \cite{125} \cite{126} to model the generation probability $p({{y}_{1}},\ldots ,{{y}_{T}}\left| U \right.)$. $U$ represents the context of the conversation and $R$ represents the response. Fig.~\ref{fig:hran_structure} shows the architecture of HRAN. Roughly speaking, before generating probabilities, HRAN uses a word-level encoder to encode the information of each utterance in the context into a hidden vector. Then, when generating each word, the hierarchical attention mechanism will focus on the important parts within and between the utterances with word-level attention and utterance-level attention, respectively. With these two attention levels, HRAN works in a bottom-up way: word-level attention processes the hidden vector of the utterance and uploads it to the utterance-level encoder to form the hidden vector of the context. Utterance-level attention further processes the hidden vectors of the context into context vectors and uploads them to the decoder to generate words.

\subsubsection{Hierarchical multiscale LSTM}
The key element of the model is the introduction of a parameterized boundary detector, which outputs a binary value in each layer of the stacked RNN and learns when segmentation should end in a way that optimizes the overall goal. Whenever the boundary detector is turned on at layer $l$ (i.e., when the boundary state is 1), the model regards it as the end of the segment corresponding to the potential abstraction level (such as a word or phrase) of the layer and send the summary representation of the detected segment to the upper layer $(l+1)$. Using boundary states, at each time step, each layer selects one of the following operations: UPDATE, COPY or FLUSH. The choice depends on (1) the boundary states at the current time step in the layer below $z_{t}^{l-1}$ and (2) the boundary state at the previous time step in the same layer $z_{t-1}^{l}$.

Later, Junyoung Chung et al. described Hierarchical Multiscale Recurrent Neural Network (HM-RNN) based on LSTM update rules, and called this model Hierarchical Multiscale LSTM (HM-LSTM) \cite{127} \cite{128}. Consider the HM-LSTM model of $L$ layer $(l=1,\ldots ,L)$, which performs the following updates at time step $t$ at each layer $l$:

\begin{equation}
h_{t}^{l},c_{t}^{l},z_{t}^{l}=f_{HM-LSTM}^{l}\left( c_{t-1}^{l},h_{t-1}^{l},h_{t}^{l-1},h_{t-1}^{l+1},z_{t-1}^{l},z_{t}^{l-1} \right)
\end{equation}

Here, $h$ and $c$ represent the hidden state and the cell state, respectively. The implementation of function $f_{HM-LSTM}^{l}$ is as follows. First, using the two boundary states $z_{t-1}^{l}$ and $z_{t}^{l-1}$, the unit state is updated in the following way:

\begin{equation}
c_{t}^{l}=\left\{ \begin{aligned}
	& f_{t}^{l}\odot c_{t-1}^{l}+i_{t}^{l}\odot g_{t}^{l}\ if\,z_{t-1}^{l}=0\,and\,z_{t}^{l-1}=1(UPDATE) \\ 
	& c_{t-1}^{l}\quad \quad \quad \ \ \ if\,z_{t-1}^{l}=0\,and\,z_{t}^{l-1}=0(COPY) \\ 
	& i_{t}^{l}\odot g_{t}^{l}\quad \quad \quad if\,z_{t-1}^{l}=1(FLUSH) \\ 
\end{aligned} \right.\,
\end{equation}

Then get the hidden state in the following way:

\begin{equation}
h_{t}^{l}=\left\{ \begin{aligned}
	& h_{t-1}^{l}\quad \quad \ \,if\,COPY \\ 
	& o_{t}^{l}\odot tanh(c_{t}^{l})\,\,otherwise \\ 
\end{aligned} \right.
\end{equation}

Among them, $(f,i,o)$ are forget, input, and output gates, and $g$ is the unit proposal vector. Note that, unlike LSTM, it is not necessary to calculate these gates and cell proposal values at every time step. For example, in the case of a COPY operation, these values do not need to be calculated, so the amount of calculation can be saved.

The advantages of this model include: (1) Capture the potential hierarchical structure in the sequence by encoding the temporal dependence of different time scales. (2) Effective hierarchical multi-scale representation can be learned to obtain better generalization performance.

\subsection{Tree RNNs}
Next, the various network architectures of Tree RNNs will be described in detail.

\subsubsection{Tree LSTM}
Tai KS et al. proposed a tree LSTM \cite{129} \cite{130} \cite{131}, which is mainly used in language analysis \cite{132}. It can fully capture the characteristics of the vessel tree structure to solve the problem of anatomical labeling. In the tree LSTM, each LSTM unit can merge information from all subunit. The tree LSTM unit at node $t$ includes input gate ${{i}_{t}}$, output gate ${{o}_{t}}$, memory gate ${{m}_{t}}$, memory unit ${{c}_{t}}$ and hidden state ${{h}_{t}}$. The update of the gate vector and memory unit vector depends on the state of all subunits. The tree LSTM unit has a forgetting gate ${{f}_{l}}$ for each child node $l$ of node $t$. This allows the tree LSTM unit to selectively retain information from each child node. Therefore, the hidden state and memory unit of the node in the tree LSTM can be updated as:

\begin{equation}
{{H}_{t}}=\sum\limits_{l\in {{C}_{t}}}{{{h}_{l}}}
\end{equation}

\begin{equation}
{{i}_{t}}=\sigma ({{W}^{(i)}}{{x}_{t}}+{{U}^{(i)}}{{H}_{t}}+{{b}^{(i)}})
\end{equation}

\begin{equation}
{{f}_{tl}}=\sigma ({{W}^{(f)}}{{x}_{t}}+{{U}^{(f)}}{{H}_{l}}+{{b}^{(f)}})
\end{equation}

\begin{equation}
{{o}_{t}}=\sigma ({{W}^{(o)}}{{x}_{t}}+{{U}^{(o)}}{{H}_{t}}+{{b}^{(o)}})
\end{equation}

\begin{equation}
{{m}_{t}}=tanh({{W}^{(m)}}{{x}_{t}}+{{U}^{(m)}}{{H}_{t}}+{{b}^{(m)}})
\end{equation}

\begin{equation}
{{c}_{t}}={{i}_{t}}\otimes {{m}_{t}}+\sum\limits_{l\in {{C}_{t}}}{{{f}_{tl}}}\otimes {{c}_{l}}
\end{equation}

\begin{equation}
{{h}_{t}}={{o}_{t}}\otimes tanh({{c}_{t}})
\end{equation}

where ${{x}_{t}}$ is the input at node $t$, and ${{W}^{(i)}}$, ${{W}^{(f)}}$,${{W}^{(o)}}$and ${{W}^{(m)}}$are the weight matrices of input, forget, output and memory gate respectively. ${{c}_{t}}$ is a subset of node $t$, $\sigma $ represents the sigmoid function, and $\otimes $ represents element wise multiplication.

\subsubsection{Bidirectional Tree LSTM}
Dan Wu extended the tree LSTM in both directions by adding an additional set of hidden state vectors in the Up-to-Down (UTD) direction \cite{133}. Different from the Down-to-Up (DTU) direction, each hidden state in UTD-LSTM has only one parent node. In fact, the path from the root node to any node forms a sequential LSTM. At each node $t$, ${{h}_{t}}$ calculation follows the bidirectional sequential LSTM model \cite{134}. Denote the DTU hidden state vector as $h_{t}^{(\uparrow )}$, the UTD hidden state vector as $h_{t}^{(\downarrow )}$, and concatenate the final representation ${{\tilde{h}}_{t}}$:

\begin{equation}
{{\tilde{h}}_{t}}=h_{t}^{(\uparrow )}\oplus h_{t}^{(\downarrow )}
\end{equation}

The softmax function is used to predict the probability ${{\hat{p}}_{{{y}_{j}}}}$ of the segment label $j$ from $\tilde{h}$. In the training stage, given the training set $D$, the training classifier maximizes the conditional log likelihood of the true class label ${{y}_{j}}$ of the training sample in each vessel segment. The training objectives are defined as follows:

\begin{equation}
{{L}_{\Theta }}=-\frac{1}{N}\sum\limits_{k=1}^{N}{log}{{\hat{p}}_{{{y}_{j}}}}+\frac{\lambda }{2}\left( {{\left\| Z \right\|}^{2}}+{{\left\| V \right\|}^{2}} \right)
\end{equation}

where $N$ is the number of vessel segments labeled in the training set, $\Theta =(Z,V)$ is the network weight, and $\lambda $ is the regularization parameter. $Z$ and $V$ are the parameters of MLP and bidirectional tree LSTM respectively.

The advantages of this model include: (1) It can effectively learn the spatial and topological dependence of blood vessels, thereby capturing the characteristics of the tree structure. (2) Competitive performance can be produced on large data sets.

\subsubsection{Attention with Tree Memory Network}
\begin{figure}[H]
	\centering
	\includegraphics[width=0.6\textwidth]{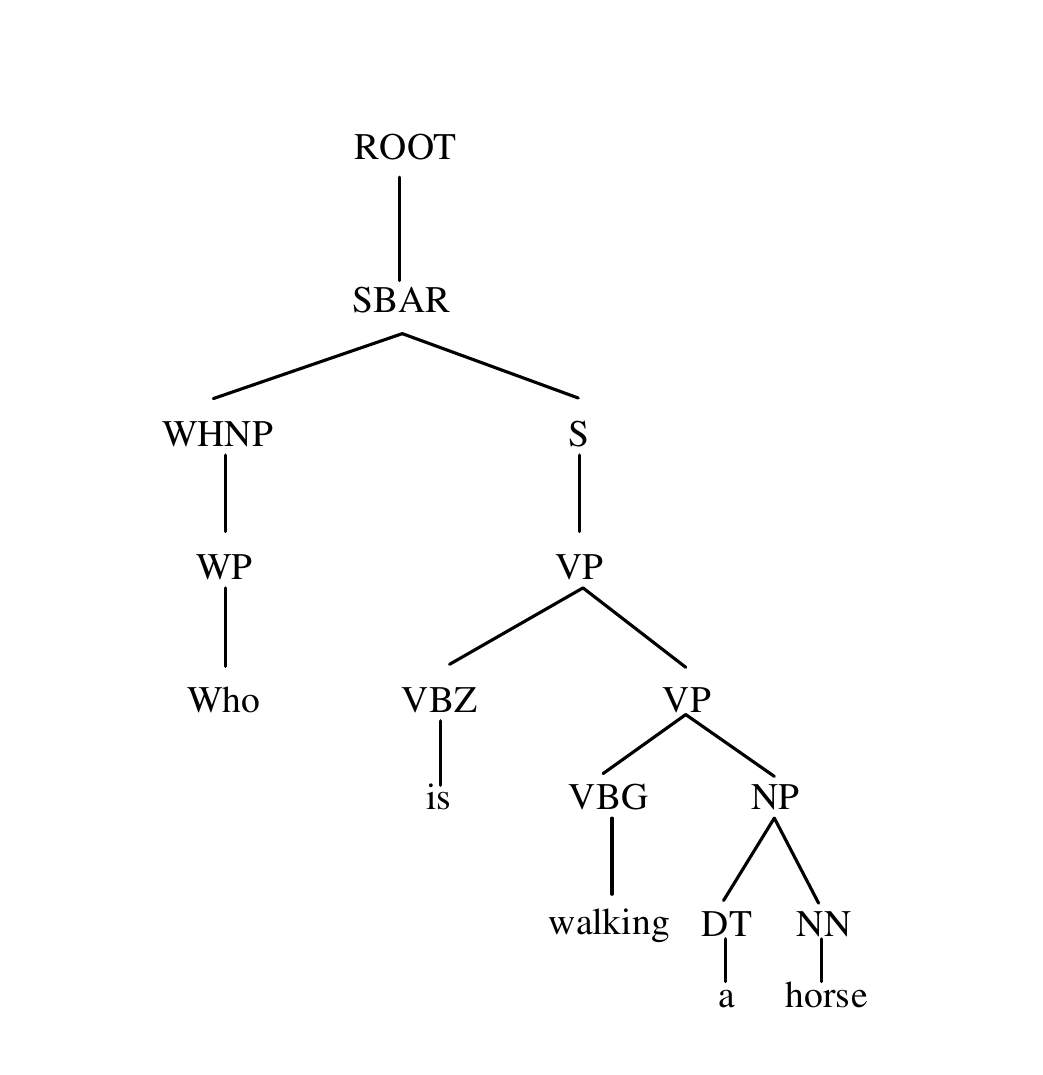}
	\caption{Stanford StanfordParser question-the original parse tree of who is walking a horse}
	\label{fig:parse_tree}
\end{figure}

\begin{figure}[H]
	\centering
	\includegraphics[width=0.4\textwidth]{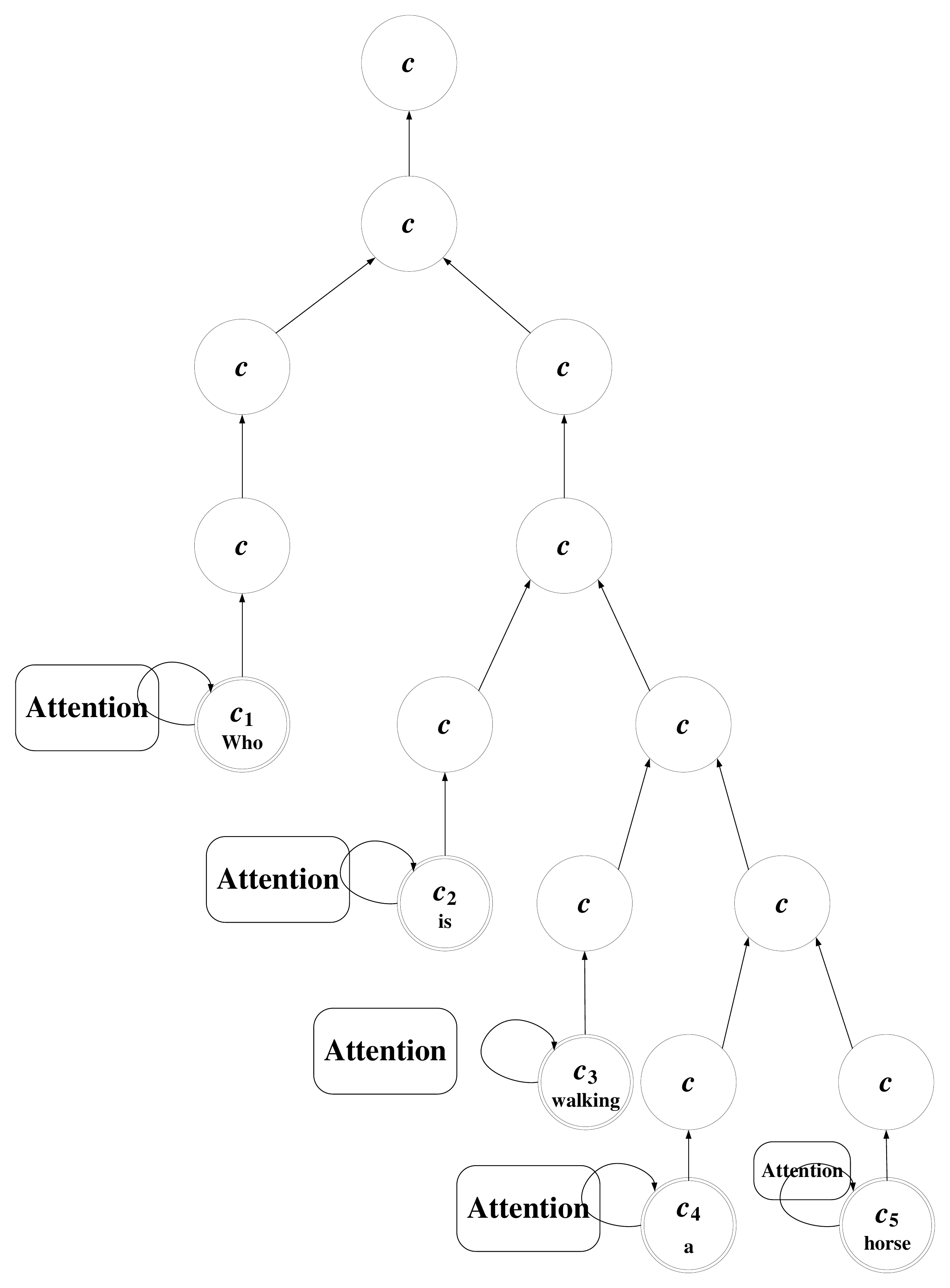}
	\caption{The Tree Memory Network (TreeMN) corresponds to the question in Fig.~\ref{fig:parse_tree}. Double-circle nodes are leaf nodes, and they are the words in the question. Each leaf node needs to calculate the attention to the video frame. The encoded features propagate from the leaf to the root.}
	\label{fig:tree_memory_network}
\end{figure}

For video question and answer questions, parse trees for different questions can have different topological structures. Hongyang Xue \cite{135} and others used the StanfordParser \cite{136} tool to construct grammar trees for sentences. Given the problem $q$, a parse tree can be constructed for it (for example, see Fig.~\ref{fig:parse_tree}). The corresponding tree structure memory network is shown in Fig.~\ref{fig:tree_memory_network}. The nodes of the memory network can be divided into three types: leaves, intermediate nodes and roots. Denote the $n$-dimensional state vector of node ${{c}_{i}}$ as ${{h}_{i}}\in {{\mathbb{R}}^{n}}$, ${{h}_{i}}$ is calculated from all its child nodes, and the child nodes are denoted as $C(i)$. Now describe how to calculate the states of each node in a bottom-up manner.

Given a video $v=\left\{ {{v}_{1}},\ldots ,{{v}_{T}} \right\}$ and a question $q=\left\{ {{q}_{1}},\ldots ,{{q}_{N}} \right\}$, where ${{v}_{i}}$ is the $m$-dimensional visual feature of the frame and ${{q}_{i}}$ is the $n$-dimensional embedding of the mark, the purpose of the task is to generate the answer from the answer list $A$. In order to avoid symbol confusion, $V$ is used to represent the matrix in ${{\mathbb{R}}^{T\times m}}$, where the $i$-th row of $V$ is the eigenvector of the $i$-th frame.

1)Attention module:Attention module is similar to E-SA model \cite{137} (a video question answering model). First, the video frame features are processed through bidirectional LSTM. The processed feature is $\left\{ {{v}_{1}},\ldots ,{{v}_{T}} \right\}$, where $T$ is the number of frames. Given state ${{h}_{i}}$, the attention of the video frame is calculated as

\begin{equation}
\begin{aligned}
	& {{h}_{A,j}}=tanh\left( {{W}_{Q}}{{h}_{i}}+{{W}_{V}}{{v}_{j}}+{{b}_{V}} \right) \\ 
	& p=softmax\left( {{W}_{p}}{{h}_{A}} \right) \\ 
	& at{{t}_{i}}=Attention\left( {{h}_{i}},V \right)=\sum\limits_{k=1}^{T}{{{p}_{k}}{{v}_{k}}} \\ 
\end{aligned}
\end{equation}

Among them, ${{h}_{A,j}}$ is the $j$-th component of ${{h}_{A}}$, ${{p}_{k}}$ is the $k$-th component of $p$, and ${{W}_{Q}}\in {{\mathbb{R}}^{n\times n}}$,${{W}_{V}}\in {{\mathbb{R}}^{n\times m}}$ and ${{b}_{V}}\in {{\mathbb{R}}^{n}}$ are parameters that convert visual and text features into joint space. ${{W}_{p}}\in {{\mathbb{R}}^{T\times n}}$ is linear transformation.

2)Leaf nodes:Leaf nodes correspond to words in interrogative sentences. The state of a leaf node does not come from its child nodes (it has no child nodes), but is calculated based on the word embedding and the attention of the video frame. Initialize the state ${{h}_{j}}$ of leaf node ${{c}_{j}}$ as ${{h}_{j}}={{q}_{j}}$, and its new state is calculated by the following formula:

\begin{equation}
{{h}_{j}}={{W}_{A}}Attention\left( {{h}_{j}},V \right)+{{b}_{A}}+{{h}_{j}}
\end{equation}

where $Attention\left( {{h}_{j}},V \right)$ is the attention of the video frame guided by ${{h}_{j}}$, and ${{W}_{A}}\in {{\mathbb{R}}^{n\times n}}$, ${{b}_{A}}\in {{\mathbb{R}}^{n}}$ are the linear transformations shared by all leaf nodes. Denote the output of node ${{c}_{j}}$ as ${{o}_{j}}$, and the output of leaf nodes as ${{o}_{j}}=tanh\left( {{h}_{j}} \right)$.

3)Intermediate node:For an intermediate node corresponding to the intermediate representation of the sentence (such as $NP$, $VP$ and $NN$), calculate the state ${{h}_{i}}$ of the intermediate node ${{c}_{i}}$ from its child node $C(i)$:

\begin{equation}
{{h}_{i}}=\sum\limits_{j\in C\left( i \right)}{{{W}_{B}}{{o}_{j}}+{{b}_{B}}}
\end{equation}

where ${{W}_{B}}\in {{\mathbb{R}}^{n\times n}}$ and ${{b}_{B}}\in {{\mathbb{R}}^{n}}$ are linear layers shared by all intermediate nodes. Its output is ${{o}_{i}}=tanh\left( {{h}_{i}} \right)$.

Perform calculations from bottom to top in a recursive manner. Information is propagated from the leaves to the parent node, and finally to the root. At the root ${{c}_{0}}$, the state ${{h}_{0}}$ can be calculated after calculating the intermediate node. Then place the linear layer, and then place the softmax layer for answer classification:

\begin{equation}
y=sofatmax\left( W{{h}_{0}}+b \right)
\end{equation}

where $W\in {{\mathbb{R}}^{z\times n}}$, $b\in {{\mathbb{R}}^{z}}$, $z$ is the number of answers.

The advantages of this model include: (1) Improve word-level attention. (2) It can alleviate the shortcomings of E-SA model's poor performance on long problems.

\subsubsection{Parse-tree-guided Reasoning Network}
\begin{figure}[H]
	\centering
	\includegraphics[width=1.0\textwidth]{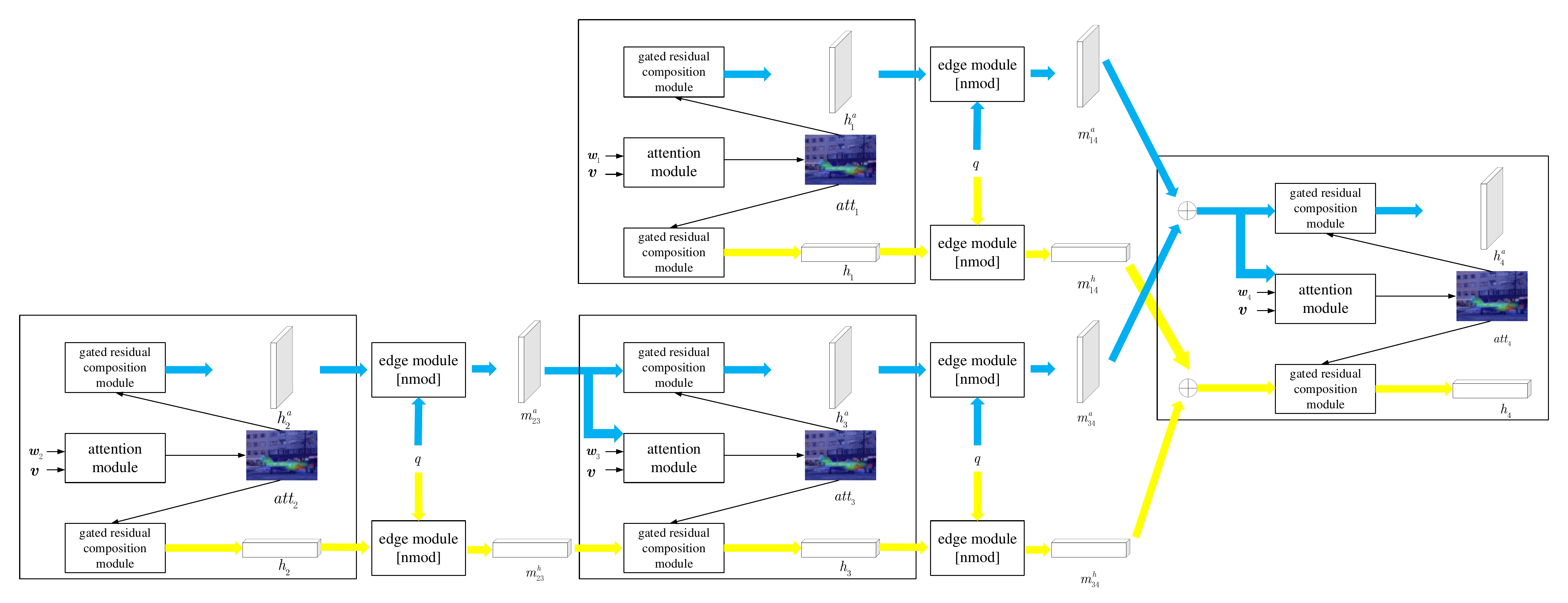}
	\caption{PTGRN method delivery path. Each PTGRN module consists of an attention module and two gated residual composition modules. Each node receives the encoded attention map and hidden feature vector of the child node, as well as image feature and word encoding. The attention module generates a new attention map based on image features, word encoding and previous attention regions given by child nodes. The gated residual composition module trains to evolve higher-level representations by dropping the features of the sub-modules and combining them with local visual characteristics. The edge module transform the outputs attention and hidden feature vectors according to the problem code and relationship type (nmod: noun modifier, dobj: direct object, nsubj: noun subject).  The blue arrow indicates the propagation process of attention mapping, and the yellow arrow indicates the process of visual hidden representation.}
	\label{fig:ptgrn_structure}
\end{figure}

Give the tree structure layout of the dependency tree, the Parsed-Tree-Guided Reasoning Network (PTGRN) \cite{138} module proposed by Qingxing Cao et al. is used in order on each word node to excavate visual evidence and integrate the features of its children from bottom to top, and then predict the final answer at the root of the tree, as shown in Fig.~\ref{fig:ptgrn_structure}. Formally, each PTGRN module can be written as

\begin{equation}
	\begin{aligned}
	& at{{t}_{j}}={{f}_{a}}\left( {{\left\{ m_{ij}^{a} \right\}}_{{{e}_{ij}}\in E}},v,w \right) \\ 
	& \ \ {{v}_{j}}=at{{t}_{j}}*v \\ 
	& \ \ {{h}_{j}}={{f}_{h}}\left( {{\left\{ m_{ij}^{h} \right\}}_{{{e}_{ij}}\in E}},{{v}_{j}} \right) \\ 
	& \ \ h_{j}^{att}=f_{h}^{att}\left( {{\left\{ m_{ij}^{a} \right\}}_{{{e}_{ij}}\in E}},at{{t}_{j}} \right) \\ 
	& \ \ m_{jk}^{h}=f_{{{e}_{jk}}}^{h}\left( {{h}_{j}},q \right) \\ 
	& \ \ m_{jk}^{a}=f_{{{e}_{jk}}}^{a}\left( h_{j}^{att},q \right) \\ 
\end{aligned}
\end{equation}

where, $v$ is the image feature vector, $w$ is the word embedding vector, $q$ is the problem code, ${{f}_{a}}$ is the attention module, $at{{t}_{j}}$ is the attention map, ${{h}_{j}}$ is the hidden representation, $E$ is a set of edges of the parse tree, and $h_{j}^{att}$ is the attention map encoding, ${{f}_{h}}$ and $f_{h}^{att}$ are gated residual composition modules, and $m_{ij}^{a}$ is the attention feature vector of child nodes.

Qingxing Cao et al. process each node by performing post-order traversal on the dependency tree. The edge type indicates whether the node acts as a modifier (a modifier can modify its parent node by referencing a more specific object) or as the subject/object of its predicate parent node. Therefore, based on the edge, the attention map and hidden representation of the node are passed to the parent node, so that the parent node can generate a more accurate attention map $at{{t}_{j}}$ or integrate the features of the child node to enhance the representation of a given predicate.

After propagating through all word nodes, the output message $\left[ m_{root}^{h},m_{root}^{a} \right]$ of the root node is used to predict the answer. Qingxing Cao et al. performed global maximum pooling on the encoded attention map $m_{root}^{a}$ and connected it with $m_{root}^{h}$. This concatenated feature predicts the final answer $y$ through a multilayer perceptron with three layers.

The model is formed by stacking a list of node modules following a tree structure layout. The weight is shared among all node modules. The entire model can be trained in an end-to-end manner using only the supervision signal $y$.

The advantages of this model include: (1) Starting from the problem, an explainable reasoning process is automatically executed on a general dependent parse tree, which greatly broadens the application field. (2) The workload becomes smaller and the model performance becomes better.

\subsubsection{General TreeNet}
The general TreeNet \cite{139} consists of two parts: a token encoder and a semantic compositor. In the constituency tree, tokens (such as words and punctuation) are basic elements and leaf nodes. Since the dimension of the word vector is generally different from the dimension of the sentence vector, the token encoder is the part of semantic conversion between words and sentences. Then, the semantic compositor learns the representation of internal nodes on the constituency tree in a partial order.

1)Token encoder:Convert leaf nodes (word embedding) into semantic space of sentences.

2)Compositor:Learn the representation of the current node by merging the left sibling and right child of the node.

\begin{figure}[H]
	\centering
	\includegraphics[width=0.6\textwidth]{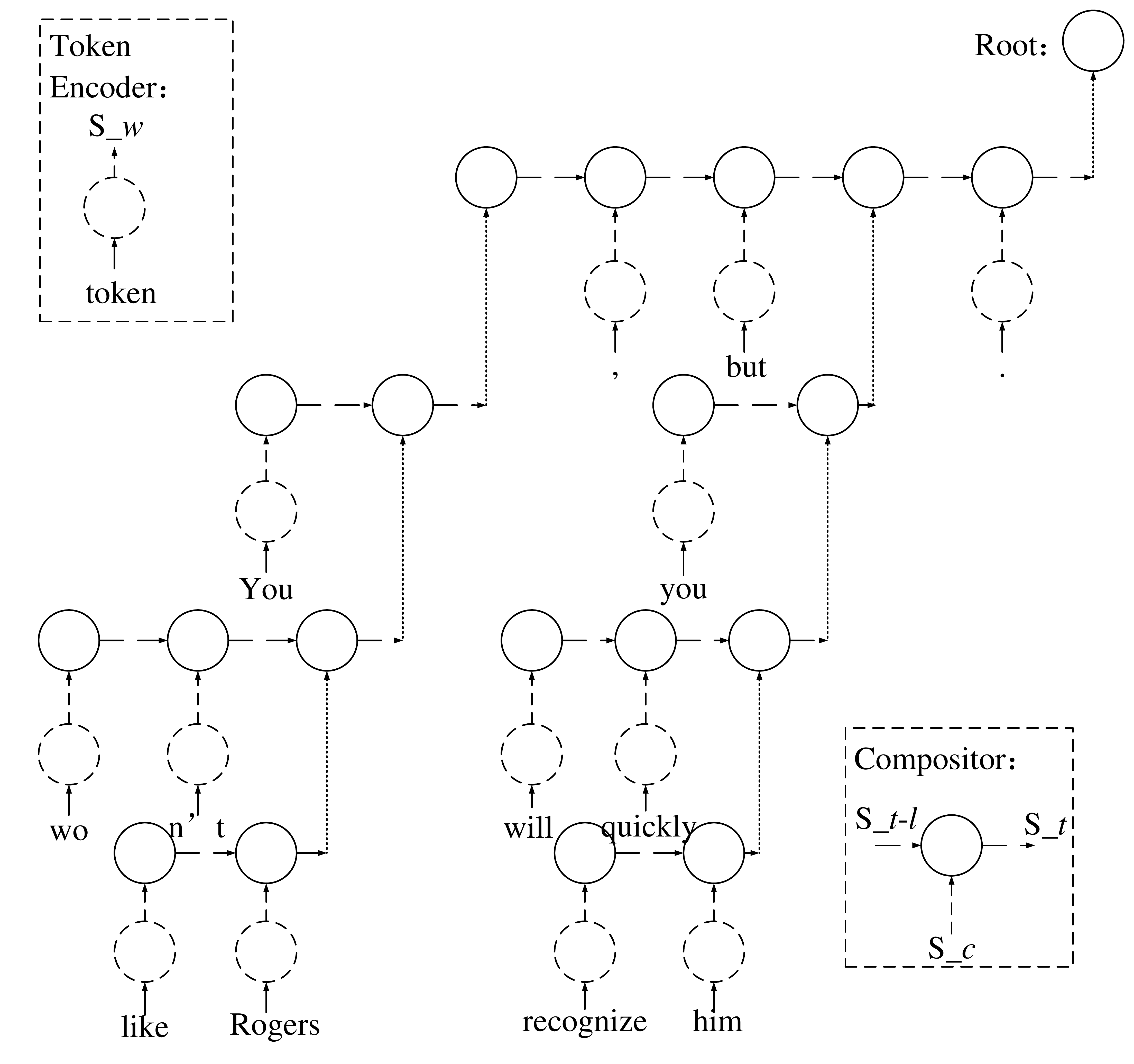}
	\caption{The calculation process of the constituency tree. First, encode the tokens from word space to sentence space. Second, calculate the internal nodes with the same parent node from left to right. At the same time, they incorporate the representation of the right child node that represents all descendants into themselves.}
	\label{fig:treenet_structure}
\end{figure}

Fig.~\ref{fig:treenet_structure} shows the dependency graph calculated on the constituency tree. The dependency graph becomes a partial order. Zhou Cheng et al. believe that this order conforms to the cognitive rules.

For each sentence, the $d$-dimensional distribution vector is used as its semantic representation, and the token can be expressed as a one-hot vector, a random distribution vector of any dimension, or a pre-training vector from other works such as GloVe \cite{140}. Therefore, general TreeNet must use a token encoder to learn a transformation function that encodes the distributed representation of tokens into sentence representations.

\begin{equation}
{{s}_{w}}=Encoder\left( {{w}_{emb}} \right)
\end{equation}

where ${{w}_{emb}}$ represents the word vector of the $w$-th word, and ${{s}_{w}}$ is its corresponding vector in the sentence semantic space. $Encoder$ stands for any neural network.

Unlike leaf nodes, each internal node has one or more sibling nodes and child nodes. Since the left internal node has no left sibling node, a zero initialization vector is used during calculation. For an internal node with multiple child nodes, the output of the right child node will be used as a representation of its child nodes. All child nodes of the same parent node are processed from left to right, so that one person can understand a sentence. The formula is as follows:

\begin{equation}
{{s}_{t}}=Compositor\left( {{s}_{t-1}},{{s}_{c}} \right)
\end{equation}

where ${{s}_{c}}$ represents the only child node of the current node $t$ (a label represents ${{s}_{w}}$) or the right child node of all child nodes, ${{s}_{t-1}}$ represents the output of its left sibling node, and the $Compositor$ function can be any network that learns how to merge them.

In TreeNet, the compositor function with calculation process plays a vital role in understanding sentences or parts of sentences. From the point of view of sibling nodes, the compositor processes them in sequence like a recurrent network because it assumes that the child node is its input and the left sibling node is its previous state. In terms of the relationship between the parent node and the child node, when the parent node applies the compositor to its left sibling node and right child node, they are updated in a recursive way. This feature of TreeNet is due to the simple calculation process and the intrinsic structure of the language.

The advantages of this model include: (1) Sentences can be captured structurally on the original unconstrained selection tree, where the number of child nodes can be arbitrary. (2) The network can learn on any tree. (3) Reduce the model parameters.

\subsubsection{Multiplicative Tree LSTM}
Encoding rich linguistic analysis introduces many different edge types or relationships between nodes, such as syntactic dependencies and semantic roles. This provides many possibilities for parameterization, but these methods are not considered in most existing syntax-aware LSTM methods, which only use the input node information.

Nam Khanh Tran et al. fill this gap by proposing a multiplicative tree LSTM (mTreeLSTM) \cite{141} \cite{142} (an extension of the tree LSTM model), and injecting relationship information into each node in the tree. mTreeLSTM introduces more fine-grained parameters based on the edge type. Inspired by multiplicative RNN \cite{18}, the hidden-hidden propagation in mTreeLSTM includes a separately learned transition matrix ${{W}_{hh}}$, and each possible edge type is learned by the following formula:

\begin{equation}
	{{\tilde{h}}_{j}}=\sum\limits_{k\in C\left( j \right)}{W_{hh}^{r\left( j,k \right)}}{{h}_{k}}
\end{equation}

where $r\left( j,k \right)$ represents the connection type between node $k$ and its parent node $j$. This parameterization is straightforward, but when there are many types of edges, a large number of parameters are required. For example, there are dozens of syntactic edge types, and each type corresponds to a Stanford dependent label.

In order to reduce the number of parameters and take advantage of the potential correlation between fine-grained edge types, Nam Khanh Tran et al. learned the embedding of edge types, and used the product of two dense matrices shared across edge types to factorize transition matrix $W_{hh}^{r\left( j,k \right)}$, which is a intermediate diagonal matrix with dependent edge type:

\begin{equation}
W_{hh}^{r\left( j,k \right)}={{W}_{hm}}diag\left( {{W}_{mr}}{{e}_{jk}} \right){{W}_{mh}}
\end{equation}

where ${{e}_{jk}}$ is the embedding of the edge type and is trained together with other parameters. The mapping from ${{h}_{k}}$ to ${{\tilde{h}}_{j}}$ is as follows:

\begin{equation}
\begin{aligned}
	& {{m}_{jk}}=\left( {{W}_{mr}}{{e}_{jk}} \right)\odot \left( {{W}_{mh}}{{h}_{k}} \right) \\ 
	& \ {{{\tilde{h}}}_{j}}=\sum\limits_{k\in C\left( j \right)}{{{W}_{hm}}{{m}_{jk}}} \\ 
\end{aligned}
\end{equation}

The calculation method of the gating unit (input gate $i$, output gate $o$ and forget gate $f$) is the same as that of TreeLSTM.

mTreeLSTM can be applied to any tree of connection type between given nodes. For example, in the dependency tree, the semantic relationship $r\left( j,k \right)$ between nodes is provided by the dependency parser.

The advantages of this model include: (1) Allow child nodes to use different combination functions, which makes the model more expressive. (2) It can effectively handle long-term dependencies.

\section{Other RNNs}
Other RNNs are divided into three categories. Array structure RNNs is not to build a single hierarchical structure, but to build a more complex memory structure inside RNNs, creating a bottleneck function by sharing internal state, forcing the learning process to use memory cells to pool the similar or interchangeable content belonging to a hidden unit; Nested and Stacked RNNs are applied to sequence prediction problems, by establishing a temporal hierarchical structure to selectively access long-term information that is only contextually relevant; Memory RNNs use keys and values to convert visual space context into language space, and effectively capture the relationship between visual features and text descriptions. It can be applied to various document reading and question answering applications.

\subsection{Array LSTM}
\begin{figure}[H]
	\centering
	\includegraphics[width=0.5\textwidth]{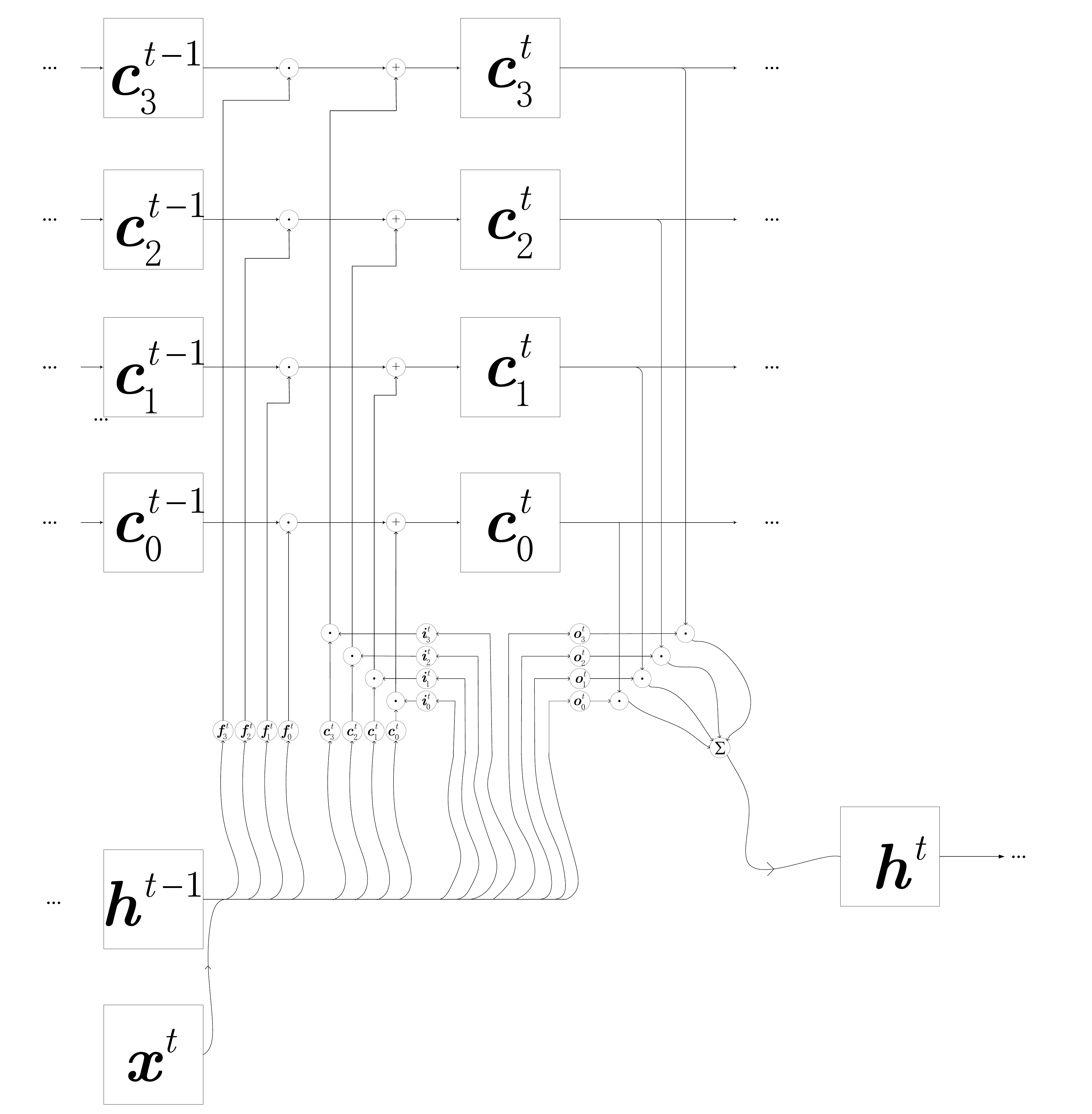}
	\caption{Array LSTM. Each hidden control unit has 4 memory units; the connections that have been modulated and being modulated, non-linearity and bias are omitted for brevity; when each hidden unit only has 1 memory unit, it becomes a standard LSTM (see 2.2.2).}
	\label{fig:array_lstm}
\end{figure}

The main idea of array LSTM is not to build a single layer hierarchical structure, (such as stacked LSTM, gated feedback RNNs), but to build a more complex memory structure in the RecurrentNNs unit \cite{143} (similar ideas by constructing more complex transition functions within the layer). Rocki et al. want to create a $bottleneck$ function by sharing internal state, forcing the learning process to use a memory unit belonging to a hidden unit to pool similar or interchangeable content \cite{144}. A similar concept already exists in convolutional networks, i.e. spatial pooling. The hidden state should be treated as a complex unit pool to handle multiple possible sub-states. Fig.~\ref{fig:array_lstm} and Eq.(12)-(17) describe the array LSTM architecture.

\begin{equation}
f_{k}^{t}=\sigma ({{W}_{fk}}{{x}^{t}}+{{U}_{fk}}{{h}^{t-1}}+{{b}_{fk}})
\end{equation}

\begin{equation}
i_{k}^{t}=\sigma ({{W}_{ik}}{{x}^{t}}+{{U}_{ik}}{{h}^{t-1}}+{{b}_{ik}})
\end{equation}

\begin{equation}
o_{k}^{t}=\sigma ({{W}_{ok}}{{x}^{t}}+{{U}_{ok}}{{h}^{t-1}}+{{b}_{ok}})
\end{equation}

\begin{equation}
c_{k}^{t}=f_{k}^{t}\odot c_{k}^{t-1}+i_{k}^{t}\odot \tilde{c}_{k}^{t}
\end{equation}

\begin{equation}
{{h}^{t}}=\sum\limits_{k}{o_{k}^{t}\odot tanh}(c_{k}^{t})
\end{equation}

The advantages of this model include: (1) More memory units. Given the same amount of memory allocation, because the matrix between hidden units and gates is smaller, the array LSTM effectively uses more memory units. (2) More parallelism. The greater the number of units, the more independent elementwise calculations are produced. (3) More data locality. Units belonging to the same hidden unit will be related to each other in some way, which can improve the cache hit ratio. In addition, for more memory units in each unit, there may be connections from sparse hidden layers to hidden layers. (4) Approximate. The stochastic array memory is resilient to noise input.

\subsection{Nested and Stacked RNNs}
Next, the various network architectures of Nested and Stacked RNNs will be described in detail.

\subsubsection{Stacked LSTM}
Stacked LSTM is now a stable technology that challenges sequence prediction problems. The stacked LSTM \cite{145} \cite{146} \cite{147} structure can be defined as an LSTM model composed of multiple LSTM layers. The upper LSTM layer provides sequence output rather than a single value output to the lower LSTM layer. Specifically, each input time step corresponds to one output, rather than all input time steps correspond to one output time step. The stacked LSTM formula is as follows:

\[i_{t}^{l}=\sigma (W_{i}^{l}h_{t}^{l-1}+U_{i}^{l}h_{t-1}^{l}+b_{i}^{l})\]

\[f_{t}^{l}=\sigma (W_{f}^{l}h_{t}^{l-1}+U_{f}^{l}h_{t-1}^{l}+b_{f}^{l})\]

\[\tilde{c}_{t}^{l}=tanh(W_{c}^{l}h_{t}^{l-1}+U_{c}^{l}h_{t-1}^{l}+b_{c}^{l})\]

\[o_{t}^{l}=\sigma (W_{o}^{l}h_{t}^{l-1}+U_{o}^{l}h_{t-1}^{l}+b_{o}^{l})\]

\[c_{t}^{l}=i_{t}^{l}\odot \tilde{c}_{t}^{l}+f_{t}^{l}\odot c_{t-1}^{l}\]

\begin{equation}
h_{t}^{l}=o_{t}^{l}\odot tanh(c_{t}^{l})
\end{equation}

where $t\in \left\{ 1,\ldots ,T \right\}$, $l\in \left\{ 1,\ldots ,T \right\}$, $W_{\cdot }^{l}\in {{\mathbb{R}}^{{{d}_{l}}\times {{d}_{l-1}}}}$, $U_{\cdot }^{l}\in {{\mathbb{R}}^{{{d}_{l}}\times {{d}_{l}}}}$ and $b_{\cdot }^{l}\in {{\mathbb{R}}^{{{d}_{l}}}}$ are trainable parameters. $\sigma (\cdot )$ and $tanh(\cdot )$ are sigmoid activation and hyperbolic tangent activation functions, respectively. Choi et al. assume that $h_{t}^{0}={{x}_{t}}\in {{\mathbb{R}}^{{{d}_{0}}}}$, where ${{x}_{t}}$ is the $t$-th input of the network.

The input gate $i_{t}^{l}$ and the forget gate $f_{t}^{l}$ control the amount of information sent from $\tilde{c}_{t}^{l}$ and $c_{t-1}^{l}$ (pre-cell state and previous cell state) to the new cell state $c_{t}^{l}$. Similarly, the output gate $o_t^l$ soft selects which part of the cell state $o_{t}^{l}$ is used in the final hidden state $c_{t}^{l}$.

\begin{figure}[H]
	\centering
	\includegraphics[width=0.7\textwidth]{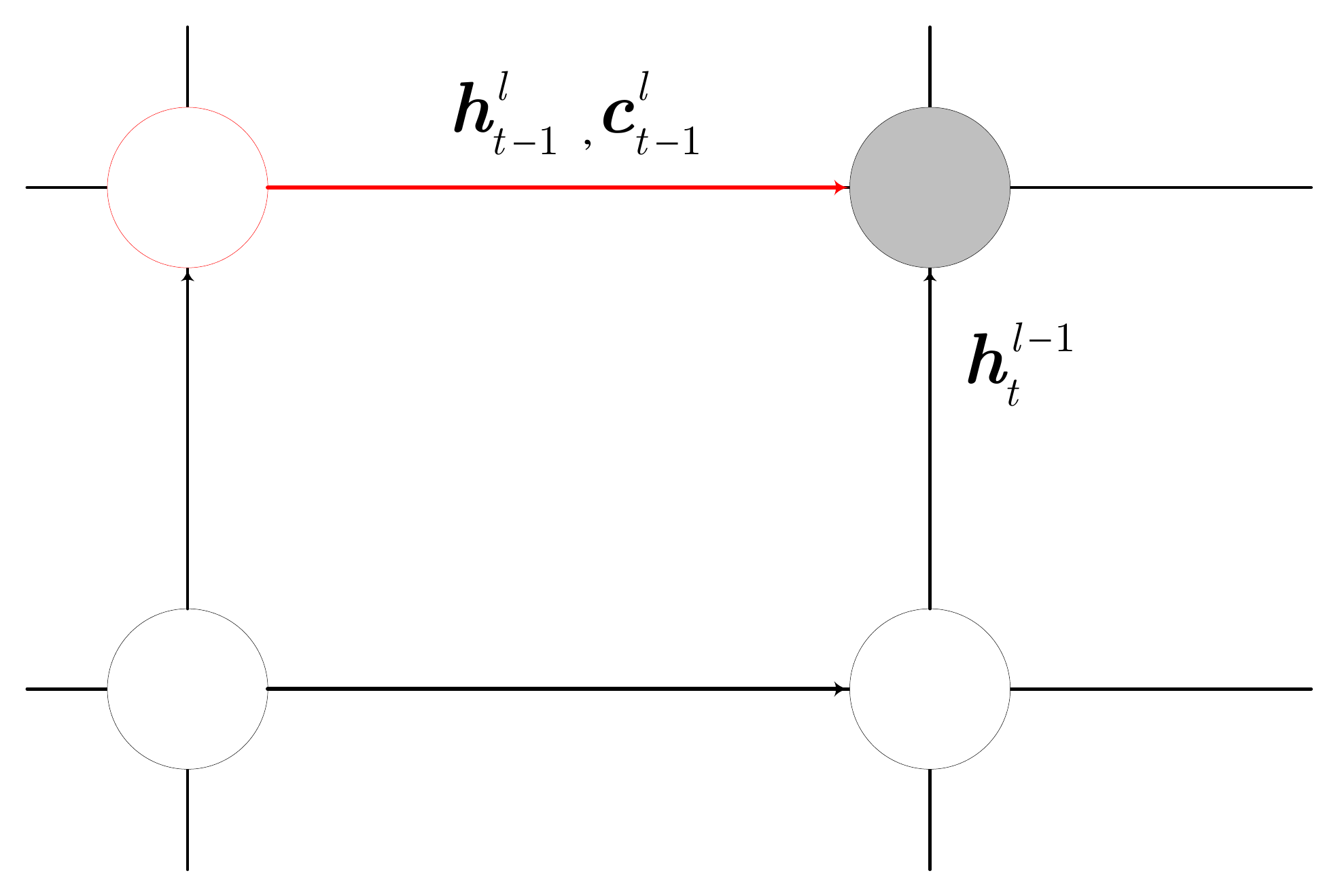}
	\caption{Stacked LSTM.}
	\label{fig:stacked_lstm}
\end{figure}

The advantages of this model include: (1) The amount of information to be transmitted can be adjusted not only in the horizontal recurrence, but also in the vertical connection. (2) The useful features extracted from the lower layer are effectively transferred to the upper layer.

\subsubsection{Nested LSTM}
The output gate in LSTM can encode an intuition that irrelevant memories at the current time step are still worth remembering. Nested LSTM (NLSTM) \cite{148} \cite{149} uses this intuition to create a temporal hierarchy of memories. Access to internal memory is gated in the same way, so long-term information that is only situationally relevant can be selectively accessed.

\begin{figure}[H]
	\centering
	\includegraphics[width=0.8\textwidth]{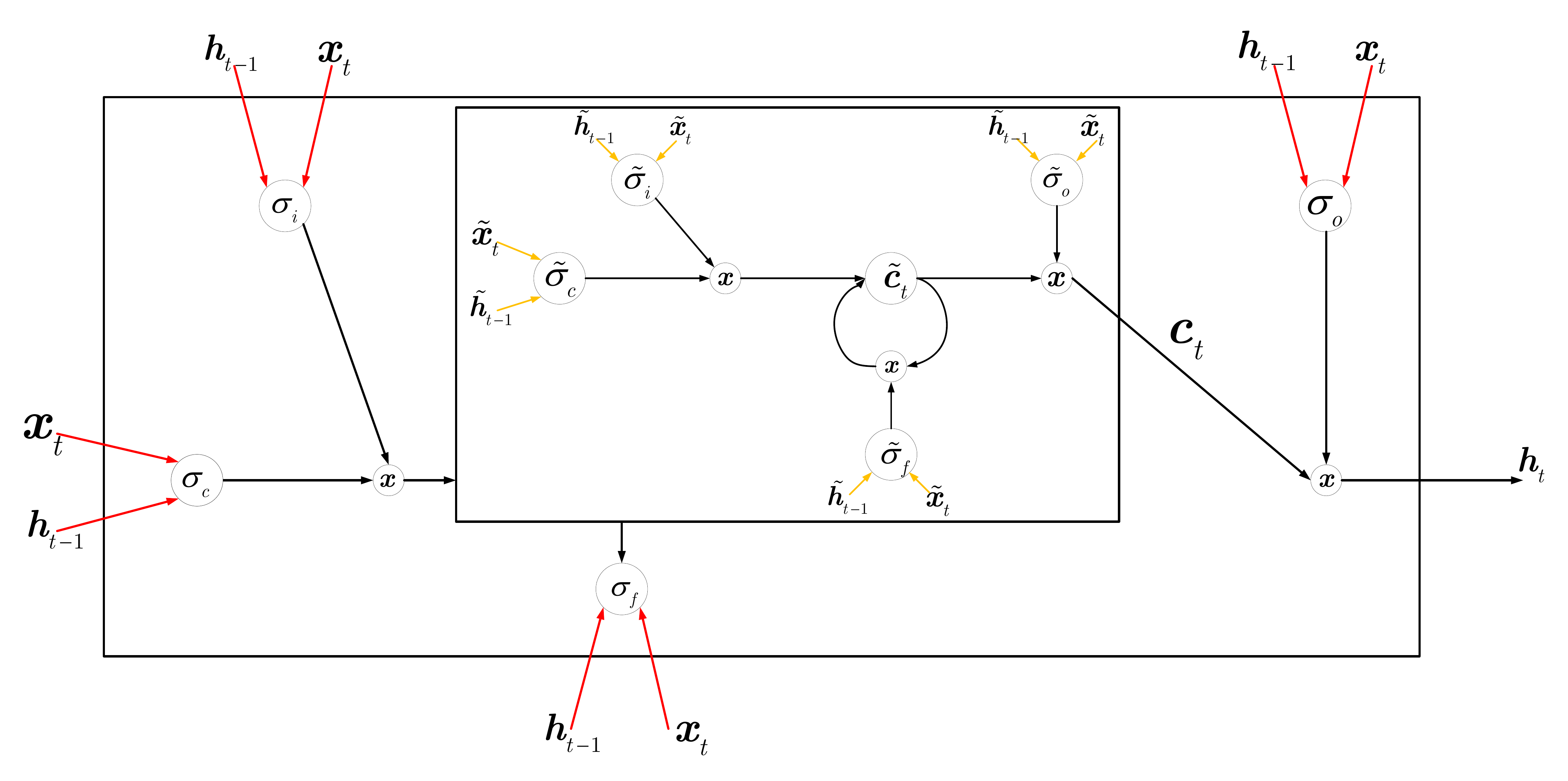}
	\caption{The NLSTM architecture.}
	\label{fig:nlstm_architecture}
\end{figure}

NLSTM replaces the addition operation used to calculate ${{c}_{t}}$ in the general LSTM with the learned state function, ${{c}_{t}}={{m}_{t}}({{f}_{t}}\odot {{c}_{t-1}},{{i}_{t}}\odot {{g}_{t}})$. Krueger et al. regard the state $m$ of the function at time $t$ as internal memory, and call the function that counts ${{c}_{t}}$ and count ${{m}_{t+1}}$ at the same time. Use the memory function as another LSTM memory unit, thus generating an NLSTM (see Fig.~\ref{fig:nlstm_architecture}). The memory function can be another nested LSTM unit, which can be nested to any depth.

Given these structures, the input and hidden state of the memory function in NLSTM become:

\[{{\tilde{h}}_{t-1}}={{f}_{t}}\odot {{c}_{t-1}}\]

\begin{equation}
{{\tilde{x}}_{t}}={{i}_{t}}\odot {{\sigma }_{c}}({{x}_{t}}{{W}_{xc}}+{{h}_{t-1}}{{W}_{hc}}+{{b}_{c}})
\end{equation}

It should be particularly noted that if the memory function is addition, the system will become a general LSTM, because the unit update will become:

\begin{equation}
{{c}_{t}}={{\tilde{h}}_{t-1}}+{{\tilde{x}}_{t}}
\end{equation}

In the NLSTM structural variant, LSTM is used as a memory function, and the internal LSTM operates as follows:

\[{{\tilde{i}}_{t}}={{\tilde{\sigma }}_{i}}({{\tilde{x}}_{t}}{{\tilde{W}}_{xi}}+{{\tilde{h}}_{t-1}}{{\tilde{W}}_{hi}}+{{\tilde{b}}_{i}})\]

\[{{\tilde{f}}_{t}}={{\tilde{\sigma }}_{f}}({{\tilde{x}}_{t}}{{\tilde{W}}_{xf}}+{{\tilde{h}}_{t-1}}{{\tilde{W}}_{hf}}+{{\tilde{b}}_{f}})\]

\[{{\tilde{c}}_{t}}={{\tilde{f}}_{t}}\odot {{\tilde{c}}_{t-1}}+{{\tilde{i}}_{t}}\odot {{\tilde{\sigma }}_{c}}({{\tilde{x}}_{t}}{{\tilde{W}}_{xc}}+{{\tilde{h}}_{t-1}}{{\tilde{W}}_{hc}}+{{\tilde{b}}_{c}})\]

\[{{\tilde{o}}_{t}}={{\tilde{\sigma }}_{o}}({{\tilde{x}}_{t}}{{\tilde{W}}_{xo}}+{{\tilde{h}}_{t-1}}{{\tilde{W}}_{ho}}+{{\tilde{b}}_{o}})\]

\begin{equation}
{{\tilde{h}}_{t}}={{\tilde{o}}_{t}}\odot {{\tilde{\sigma }}_{h}}({{\tilde{c}}_{t}})
\end{equation}

The unit states of the external LSTM are updated to:

\begin{equation}
{{\tilde{c}}_{t}}={{\tilde{h}}_{t}}
\end{equation}

The advantages of this model include: (1) A clearer temporal hierarchies in the activation of its memory units. (2) The internal memory unit of the NLSTM forms an internal memory, which can only be accessed by other computing units through the external memory unit, realizing the form of a temporal hierarchical structure.

\subsection{Memory RNNs}
Next, the various network architectures of Memory RNNs will be described in detail.

\subsubsection{Key-Value Memory Networks}
Key-Value Memory Networks (KVMN) \cite{150} \cite{151} \cite{152}, the memory slots in the network are key-value pairs $({{k}_{1}},{{v}_{1}}),\ldots ,({{k}_{T}},{{v}_{T}})$. Keys and values are used to convert visual space context into language space and effectively capture the relationship between visual features and text descriptions. The definition of memory, addressing and reading schema are summarized as follows \cite{153}:

Keys $(K)$:Use a CNN encoder to generate a visual context key ${{k}_{i}}$ for each frame ${{I}_{i}}$ of the video. These appearance feature vectors are combined with the sequential structure (video is an image sequence) through the RNN encoder. It can be seen from ${{k}_{i}}=h_{i}^{e}$ that the hidden state $h_{i}^{e}$ at each time step is extracted as the key ${{k}_{i}}$.

Value $(V)$:For each image frame ${{I}_{i}}$, a semantic embedding $v_t^{\text{sem}}$ representing the meaning of a specific frame/key text is generated. It is very difficult to jointly learn visual semantic embedding in the Encoder-Decoder model, and the supervision signal only comes from the annotation description \cite{154}. In order to alleviate this situation, it is necessary to pre-calculate the semantic embedding corresponding to each frame in the video. For example, suppose that ${{v}_{i}}$ is obtained from the pre-trained model $\psi $, which simulates the visual and semantic embedding of the image given by ${{v}_{i}}=\psi ({{I}_{i}})$ \cite{155}\cite{156}. Now, for each frame ${{I}_{i}}$ in the video, there is a key-value memory slot $({{k}_{i}},{{v}_{i}})$.

Key Addressing:This corresponds to the soft attention mechanism used to assign a relevance probability $\alpha_t$ to each memory slot. These relevance probabilities are used for value reading.

Value Reading:The value reading of the memory slot is the weighted sum of the key value feature vectors ${{\phi }_{t}}(K)$ and ${{\phi }_{t}}(V)$ at each time step. ${{\phi }_{t}}(K)$ is used for key addressing at the next time step, and ${{\phi }_{t}}(V)$ is passed as input to the decoder RNN to generate the next word.

\begin{equation}
{{\phi }_{t}}(K)=\sum\limits_{i=1}^{T}{\alpha _{i}^{(t)}{{k}_{i,{{\phi }_{t}}(V)}}=\sum\limits_{i=1}^{T}{\alpha _{i}^{(t)}{{v}_{i}}}}
\end{equation}

The advantages of this model include: (1) The key-value memory network is extended to a multi-modal environment to generate a natural description of the video. (2) Solve the problem of maintaining long-term temporal dependencies in the videos.

\subsubsection{Recurrent Memory Network}
It has been proven that RNNs can retain input information for a long period. However, the existing RNNs architecture is difficult to analyze at each time step which information is accurately retained in its hidden state, especially when the data has a complex underlying structure (this is common in natural languages). Based on this problem, Ke Tran et al. proposed a novel RNN architecture called Recurrent Memory Network (RMN) \cite{157}. For linguistic data, RMN can not only determine which linguistic information will be retained over time and why this is the case, but also find dependencies in the data. RMN consists of two components: LSTM and Memory Block (MB). MB takes the hidden state of LSTM and uses the attention mechanism to compare it with the latest input \cite{158}\cite{159}. Memory block is actually a variant of memory network \cite{160}. Therefore, analyzing the attention weight of a trained model can provide a deep understanding of the information retained over time in the LSTM.

\begin{figure}[H]
	\centering
	\includegraphics[width=0.8\textwidth]{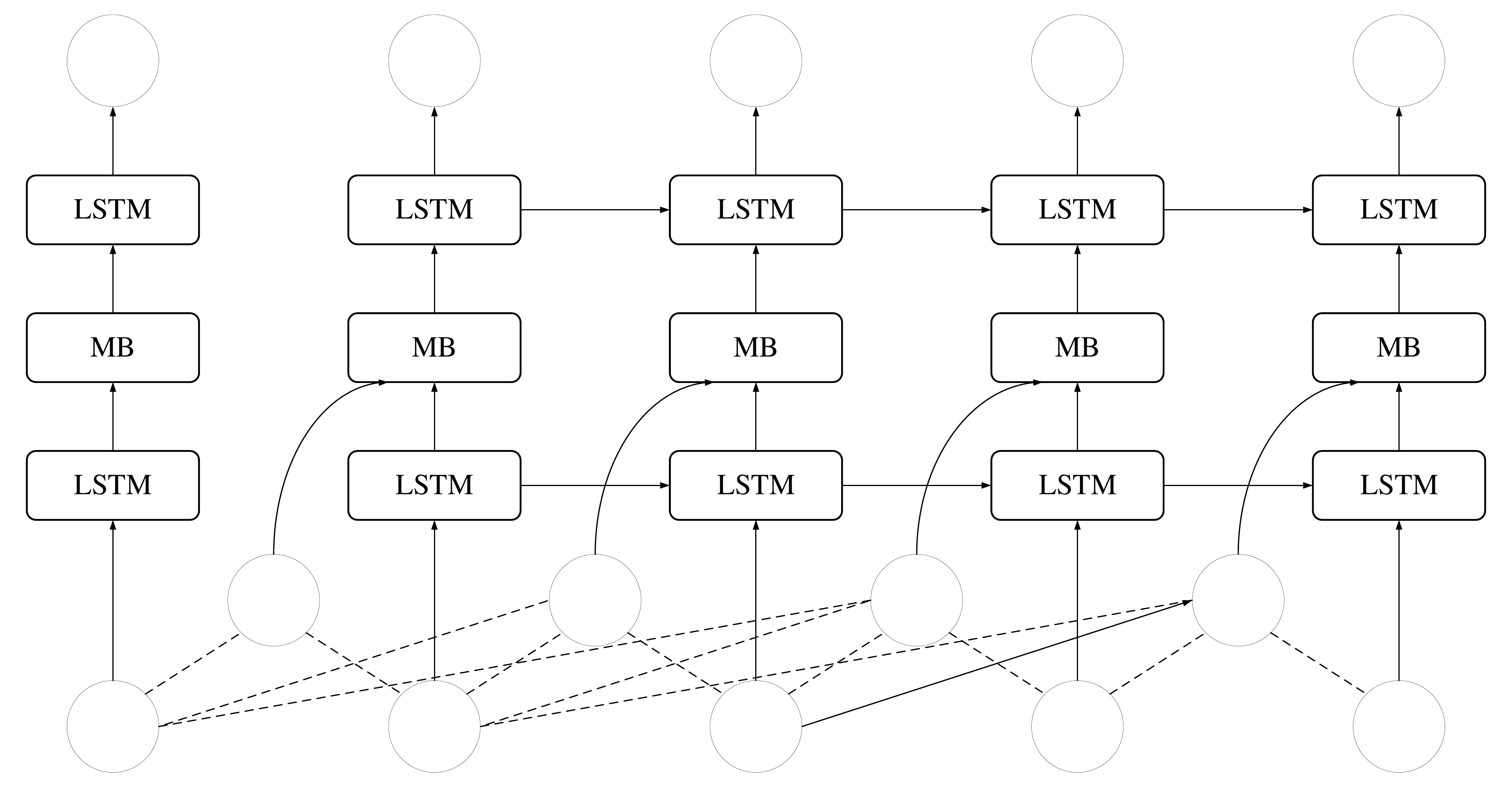}
	\caption{RMN structure.}
	\label{fig:rmn_structure}
\end{figure}

The advantages of this model include: (1) Implicitly capture certain types of dependencies that are important for word prediction. (2) The function of RecurrentNNs is magnified, and it is helpful to understand its internal function, and can discover potential patterns in the data.

\section{The Problems and Future Directions}
RNNs try to combine with other networks to achieve more complex functions and achieve better results. They have made significant progress in recent years and are the current mainstream artificial intelligence learning direction. This article sorts out the main types of RNNs, and gives the model's construction process, advantages and disadvantages, and problems in the model. Although RNNs have great potential, there are also many challenges:

\textbf{The disappearance and explosion of gradient:} This is not only the problem of RNNs. Due to the chain rule and nonlinear activation function, almost all networks will have the problem of gradient disappearance and explosion. Many model variants that have emerged to solve this problem can only slightly alleviate the problem of gradient disappearance and explosion. Therefore, there is still a lot of space to completely solve the problem of gradient disappearance and explosion.

\textbf{Dedicated to multi-task learning and general natural language understanding:} all models applied throughout the paper are task-specific. But people are becoming more and more interested in multi-task learning in NLP. Since common concepts in the language will be applied to a single task, sharing information across tasks is intuitive. This can be seen as the first step in a universal, task-independent natural language understanding model. Although people have always been interested in using neural networks for multi-task learning, the improvement is still relatively small, and the best mechanism for sharing knowledge across tasks is unclear, so multi-task learning is an area worthy of in-depth research.

\textbf{Generalization ability:} Machine learning theory believes that a good model must have better generalization ability. From the viewpoints of model complexity and bias-variance trade-offs, theoretically discuss the learning mechanisms of various RNNs, enrich the theoretical foundation of the model, and improve the generalization ability of recursive and recurrent neural networks.

\textbf{Application field expansion:} The application range of RNNs is relatively small. How to apply the ideas and results of RNNs in common scenarios and how to accelerate the integration with these fields is the key direction for the further development of RNNs in the future. At present, RNNs are mainly used in the field of artificial intelligence, and applications in other fields need to be further developed.

\textbf{Expansion of evaluation indicators:} Although some results have been achieved in measuring the pros and cons of the model, there are still some problems. For some complex tasks such as video tracking, language and image analysis, it is difficult to use existing indicators to evaluate the model. Therefore, expanding the evaluation index can make the model more accurate.

\textbf{Processing and expansion of available data sets:} Although RNNs show good performance on the existing prescribed data sets, in practical applications, the instability and complexity of the data will greatly reduce the accuracy of the model, so the effective preprocessing of the data is a problem worth studying.

\textbf{Parameters and calculations:} Although RNNs are efficient, the corresponding variants combine different structures to form a complex hybrid structure, resulting in an increase in model parameters and a substantial increase in calculations. Therefore, how to reduce model parameters and improve calculation efficiency under the condition of ensuring model accuracy is a problem worthy of in-depth study.

\textbf{Model versatility:} RNNs and their variants also have many limitations in the fields that can be applied, including the data set that can be processed is limited, so whether the model can be made relatively universal in a certain field is a problem worth studying.

\section{Conclusion}
In this paper, the branches of Recursive and Recurrent Neural Networks are classified in detail according to the network structure, training objective function and learning algorithm implementation. They are roughly divided into three categories: The first category is General RNNs, including RecursiveNNs and RecurrentNNs, LSTM, Convolutional RNNs, Differential RNNs, One-layer RNNs, High-Order RNNs, Highway Networks, Multi-Dimensional RNNs, Bidirectional RNNs, their basic subjects are RecurrentNNs and very classic variants LSTM and ordinary variants of LSTM. They use various methods to solve the problem that ordinary Recursive and Recurrent Neural Networks cannot learn large-scale sequence dependencies; The second category is Structured RNNs, including Grid RNNs, Graph RNNs, Temporal RNNs, Lattice RNNs, Hierarchical RNNs, Tree RNNs, which realize their functions in different situations through the combination of different structures; the third category is Other RNNs, including Array LSTM, Nested and Stacked RNNs, Memory RNNs. With the development and integration of various networks, many complex sequence, voice and image problems have been solved. From the summary of the above various models, it can be seen that the types of RNNs are quite rich and develop rapidly. Although all types of models have certain problems and limitations, it cannot be denied that with the further deepening of theoretical research and the further expansion of application fields, RNNs will surely become mainstream technologies in the field of artificial intelligence in the future.
	
	% References section
	
	\end{CJK*}
\end{document}